\pdfoutput=1                       % pdfLaTeX, for the PDF figures
\documentclass[12pt,a4paper]{article}

\usepackage[margin=2cm]{geometry}
\usepackage{amsmath,amssymb,bm}
\usepackage{graphicx}
\usepackage{booktabs}
\usepackage{siunitx}
\usepackage[numbers,sort&compress]{natbib}
\usepackage{xcolor}
\usepackage{tikz}
\usetikzlibrary{positioning,arrows.meta,calc,patterns}
\usepackage{pgfplots}
\pgfplotsset{compat=1.18}
\usepackage{algorithm}
\usepackage{algpseudocode}
\usepackage{subcaption}
\usepackage[section]{placeins}
\usepackage[colorlinks=true,linkcolor=blue,citecolor=blue,urlcolor=blue]{hyperref}

\graphicspath{{figures/}}

\definecolor{cblue}{HTML}{1F77B4}
\definecolor{cred}{HTML}{D62728}
\definecolor{cgreen}{HTML}{2CA02C}
\definecolor{corange}{HTML}{FF7F0E}
\definecolor{cpurple}{HTML}{9467BD}
\pgfplotsset{
  paperaxis/.style={
    tick align=inside,
    tick label style={font=\footnotesize},
    label style={font=\small},
    legend style={font=\footnotesize, draw=none, fill=none},
    legend cell align=left,
  }
}

\tikzset{
  spec/.style={fill=cblue!6, draw=black!70, line width=0.7pt},
  crackline/.style={cred, line width=1.2pt, line cap=round},
  clamphatch/.style={pattern=north east lines, pattern color=black!55},
  loadarrow/.style={-{Stealth[length=2.0mm]}, cred, line width=0.8pt},
  dimline/.style={{Stealth[length=1.5mm]}-{Stealth[length=1.5mm]},
                  black!55, line width=0.5pt},
  dimfont/.style={font=\footnotesize, text=black!75},
}
\newcommand{\ub}{\bm{u}}                          % displacement vector
\newcommand{\xb}{\bm{x}}                          % physical coordinates
\newcommand{\xib}{\bm{\xi}}                       % parametric coordinates
\newcommand{\pf}{\phi}                            % phase field
\newcommand{\straintensor}{\bm{\varepsilon}}
\newcommand{\stresstensor}{\bm{\sigma}}
\newcommand{\Jgeo}{\bm{J}}                        % geometry-map Jacobian
\newcommand{\detJ}{\lvert \det \Jgeo \rvert}

\newcommand{\Gc}{G_{\mathrm{c}}}
\newcommand{\lreg}{l}                             % regularization length (our convention)
\newcommand{\kres}{\kappa}                        % residual stiffness (kappa: plain k is the layer index in the MLP recursion)

\newcommand{\grad}{\nabla}
\newcommand{\lap}{\Delta}
\newcommand{\dive}{\operatorname{div}}
\newcommand{\trace}{\operatorname{tr}}
\newcommand{\dV}{\,\mathrm{d}\Omega}
\newcommand{\dA}{\,\mathrm{d}\Gamma}

\newcommand{\Ene}{\Psi}                           % total energy functional
\newcommand{\edens}{\psi_{\mathrm{e}}}            % elastic strain energy density
\newcommand{\psiir}{\psi_{\mathrm{ir}}}           % irreversibility penalty density
\newcommand{\psihyb}{\psi_{\mathrm{hyb}}}         % hybrid elastic energy density
\newcommand{\psitot}{\psi_{\mathrm{tot}}}         % total energy density (integrand)
\newcommand{\dmg}{g}                              % degradation function

\newcommand{\R}{\mathbb{R}}
\newcommand{\Ex}{\mathbb{E}}                      % expectation (MC estimator)
\newcommand{\sg}[1]{\mathrm{sg}\!\left[#1\right]} % stop-gradient (frozen coupling)
\newcommand{\pospart}[1]{\left\langle #1 \right\rangle_{+}}
\newcommand{\negpart}[1]{\left\langle #1 \right\rangle_{-}}
\newcommand{\devop}{\operatorname{dev}}
\newcommand{\Fenc}{\bm{F}}                        % multiresolution feature vector

\title{A mesh-free multiresolution deep energy method with phase-field
modeling of brittle fracture}
\author{%
  Han Zhang$^{\mathrm{a}}$\thanks{E-mail: han.zhang7@unsw.edu.au}~,
  Mehrisadat Makki Alamdari$^{\mathrm{a}}$,
  Babak Shahbodagh$^{\mathrm{a}}$,\\
  Mohammad Vahab$^{\mathrm{b}}$,
  Cosmin Anitescu$^{\mathrm{c}}$,
  Timon Rabczuk$^{\mathrm{c}}$,
  Elena Atroshchenko$^{\mathrm{a}}$\thanks{Corresponding author.
  E-mail: e.atroshchenko@unsw.edu.au}\\[1.2ex]
  {\normalsize $^{\mathrm{a}}$University of New South Wales, Australia}\\
  {\normalsize $^{\mathrm{b}}$Central Queensland University,
  Melbourne, Australia}\\
  {\normalsize $^{\mathrm{c}}$Bauhaus-Universit\"at Weimar, Germany}}
\date{}

\begin{document}
\maketitle

% =====================================================================
\begin{abstract}
Phase-field modeling of brittle fracture removes the need to track
cracks explicitly by recasting their evolution as the minimization of
an energy functional. In return it requires a discretization dense
enough to resolve a localization band whose width is set by a
regularization length and whose path is not known in advance. In this work we
propose a mesh-free discretization in which a single neural network
represents the displacement and phase fields and is trained by
minimizing the incremental energy directly. The coordinates enter the
network through a multiresolution feature encoding built from
$C^{1}$ quadratic B-spline grids, so the finest scale the
representation can express is set by choice rather than reached
through slow training. The energy is estimated by stratified Monte
Carlo integration on points redrawn at every optimizer iteration,
which concentrates resolution on the damaged region and leaves no
fixed point set for the optimizer to exploit. This pairing proves
critical, since the crack fails to advance both when the integration
points are held fixed and when the encoding is too coarse to represent
the band, while each ingredient tolerates a wide range of settings
once the other is in place. Essential boundary conditions are
imposed exactly through lifts, and curved domains are treated through
an exact map built from non-uniform rational B-splines together with a
domain mask. Since the field representation
is globally $C^{1}$, the second- and the fourth-order
fracture energy densities run on the identical discretization. The
method is assessed on single-edge-notched tension and shear, crack
branching, the coalescence of en-echelon cracks, crack nucleation at
a circular hole and a thick-walled ring computed on a single spline
patch. The computed load--displacement curves follow staggered
finite element references at matched regularization length,
with peak loads within about $1\%$ on the single-edge-notched
tests and within $8\%$ where the crack pattern changes
topology. On a public benchmark dataset of
random multi-crack configurations the method classifies the active
or dormant state of $90\%$ of the seeded cracks in twenty
zero-shot runs, where the deep Ritz baseline of the dataset authors
fails. Robustness with
respect to the network initialization and the sampling sequence is
quantified over an ensemble of random seeds.
\end{abstract}

% =====================================================================
\section{Introduction}
\label{sec:intro}

The prediction of crack nucleation and growth remains one of the
central difficulties of computational solid mechanics. Classical
computational fracture builds on the sharp-crack description of linear
elastic fracture mechanics~\cite{griffith1921,irwin1957} and
represents the crack explicitly, whether by remeshing around an
advancing front~\cite{bittencourt1996}, by cohesive interface
elements along candidate paths~\cite{xu1994}, or by enriching the
finite element space with discontinuous modes so that the crack can
cross elements~\cite{moes1999}. All of these treatments require the
crack geometry to be tracked as it evolves, and they must be
supplemented with separate criteria for nucleation, for the kinking
angle and for branching, which becomes increasingly delicate once
several cracks interact or merge.

The wish to free the crack from the mesh is older than machine
learning. Element-free Galerkin methods build the fields from moving
least-squares approximants on scattered nodes~\cite{belytschko1994},
and meshfree crack methods in this tradition, such as the cracking
particles of Rabczuk and Belytschko~\cite{rabczuk2004}, have treated
arbitrarily evolving cracks without any remeshing. Two burdens remain
in these methods. The crack itself stays an explicit geometric
object, carried by visibility or enrichment constructions around its
faces, and the integrals of the weak form are evaluated on background
cells, so the quadrature keeps a mesh-like structure that must track
the solution even though the approximation does not. Removing the
mesh does not eliminate the burden; it moves it into the integration.

The variational view introduced by Francfort and
Marigo~\cite{francfort1998} removes the crack bookkeeping altogether
by recasting
Griffith's energetic criterion as a minimization problem for the sum
of elastic and surface energy, so that nucleation, propagation,
kinking, branching and merging arise as outcomes of energy
minimization rather than through criteria attached to each event. The
formulation became computable through the regularization of Bourdin
et al.~\cite{bourdin2000,bourdin2008}, in which the unknown crack set
is replaced by a scalar phase field that localizes in a diffuse band
of width proportional to a regularization length $\lreg$, with
convergence to the free-discontinuity problem as $\lreg$
vanishes~\cite{ambrosio1990}. The model matured through
thermodynamically consistent formulations and operator
splits~\cite{miehe2010,miehe2010cmame,kuhn2010}, energy decompositions
that restrict damage to tension-like
states~\cite{amor2009,freddi2010,ambati2015,vicentini2024}, a
gradient-damage reading that gives the regularized model its own
mechanical content~\cite{pham2011,wu2017}, solution strategies from
alternating minimization to monolithic and active-set
schemes~\cite{gerasimov2016,heister2015}, and irreversibility enforced
either through history fields~\cite{miehe2010cmame} or through
penalization~\cite{gerasimov2019}.

One energetic setting has since carried dynamic
branching~\cite{borden2012}, ductile
fracture~\cite{ambati2015ductile}, fatigue~\cite{carrara2020},
hydrogen-assisted cracking~\cite{martinez2018}, thermal-shock crack
patterns~\cite{sicsic2014,bourdin2014} and coupled multi-physics
fracture~\cite{miehe2015,tangella2022}, and the field has been
reviewed comprehensively~\cite{ambati2015,wu2020,li2023review}. Two
developments matter here. The first is nucleation from a smooth
boundary, which discrete-crack descriptions must postulate and which
arises naturally in the regularized setting, connecting the
regularization length to a material
strength~\cite{tanne2018,kumar2020,vicentini2024}. The second is the
fourth-order fracture energy density
with improved convergence properties in $\lreg$ proposed by
Borden et al.~\cite{borden2014}, following the isogeometric treatment
of other higher-order phase-field equations~\cite{gomez2008}.

These capabilities carry a resolution requirement. The energy
competition is meaningful only where the discretization resolves the
band of width proportional to $\lreg$, and since the crack path is
not known in advance, a mesh-based solver must either refine
everywhere the crack might pass or follow the evolving pattern with
adaptive refinement machinery, for which dedicated predictor-corrector
and anisotropic strategies have been
developed~\cite{heister2015,artina2015,si2023,goswami2020fourth}.
The requirement is quantitative rather than nominal, since reported
computations place several elements across the band, with element
sizes of one half of $\lreg$ and below near the expected
path~\cite{miehe2010cmame}, and the band is the one region whose
location the computation is supposed to discover. The
fourth-order model raises the requirement further, since its density
contains second derivatives and a Galerkin treatment needs trial
spaces of global $C^{1}$ continuity, a principal motivation for
isogeometric discretizations~\cite{borden2014,hughes2005,cottrell2009}.
Curved domains add a further layer through boundary-fitted meshing
or, in the isogeometric setting, through multi-patch couplings with
interface terms~\cite{ruess2014,si2023}. None of these ingredients is
prohibitive, yet each ties the discretization more tightly to a mesh
whose main task is to follow the crack.

Neural networks offer a discretization of a different kind. Used as
global trial functions~\cite{lagaris1998,berg2018,raissi2019,
karniadakis2021}, they are smooth, mesh-free and differentiable to
any order through automatic differentiation, and mature software
exists for their training~\cite{lu2021deepxde,haghighat2021}. For
problems governed by a minimum principle, the deep Ritz
method~\cite{e2018deepritz} and the closely related deep Galerkin
method~\cite{sirignano2018} minimize the governing functional
directly over the network parameters, an approach carried into solid
mechanics as the deep energy
method~\cite{samaniego2020,nguyenthanh2020} and since extended to
three-dimensional, mixed and structural
formulations~\cite{abueidda2021,fuhg2022,zhuang2021}. Variational
forms with test functions have been explored as
well~\cite{kharazmi2021}, essential boundary conditions can be
imposed exactly through distance-based lifts~\cite{sukumar2022}, and
the optimization behavior of such solvers has received its own
scrutiny~\cite{wang2021,krishnapriyan2021,jagtap2020,wang2022ntk}.
One point returns throughout this paper. The optimizer follows the
estimated functional rather than the exact one, so the reliability of
an energy-based solver rests on the estimate of its energy integral,
and integration and sampling for network-based losses have become
subjects of their own~\cite{wu2023,nabian2021}.

Since the phase-field description of brittle fracture is itself an
incremental minimization problem, this line of work extends to it
naturally. Goswami et al.\ minimized the phase-field energy with
transfer learning between load steps~\cite{goswami2020tl}, treated
the fourth-order model with a quadrature grid refined adaptively
along the crack~\cite{goswami2020fourth}, and later learned solution
operators for families of fracture
problems~\cite{goswami2022deeponet}. Manav et
al.~\cite{manav2024} gave a systematic account of the deep Ritz
treatment, demonstrating nucleation, propagation, kinking, branching
and coalescence, together with a careful analysis of the optimization
difficulties involved. Chakraborty et al.~\cite{chakraborty2022}
carried the energy formulation onto a decomposed domain, assigning a
separate network to each subdomain, coupling the subdomains through
interface penalties on the displacement and the phase field, and
treating the fourth-order model on an adaptively refined element
structure. Ghaffari Motlagh et al.~\cite{motlagh2023} compared the
residual, the weak and the energy formulations of the same
phase-field model and found the energy formulation the most accurate
of the three. Wang et al.~\cite{wang2026xdem} brought the discrete and
the diffuse descriptions into one energy-based framework, enriching the
trial displacement with a discontinuity and with the crack-tip
asymptotics of linear elastic fracture mechanics where the crack is
represented explicitly, and retaining the phase field where it is not. The higher-order densities have since been taken up in
the energy setting from two directions. Plung\.{e} et
al.~\cite{plunge2026} extended the deep Ritz treatment to a family of
anisotropic higher-order crack density functionals, enriching the
trial space with higher-order B-spline basis functions in order to
represent the higher-order gradients without automatic
differentiation, and Dean and Bahtiri~\cite{dean2026} coupled a
fourth-order model to a conduction problem for self-sensing
materials, taking the Laplacian by automatic differentiation on a
fixed grid of Gauss cells. Further routes place the network elsewhere in the
computation, learning solution operators from finite element
data~\cite{kiyani2025}, minimizing a peridynamic potential in place
of the local one~\cite{ning2023}, enriching the trial fields with the
crack-tip asymptotics of linear elastic fracture
mechanics~\cite{gu2023}, predicting the phase field within a finite
element solver that retains the equilibrium
solve~\cite{pantidis2026}, or supplying the bulk constitutive
response to a conventional phase-field code~\cite{dammass2025}. The
wider use of machine learning in fracture and damage mechanics has
been surveyed recently~\cite{ani2026}.

The energy-minimizing accounts are instructive for what they had
to give up. Networks
with smooth activations carry a bias toward smooth functions, so a
localized band at the scale $\lreg$ emerges only slowly in
training~\cite{rahaman2019,jacot2018,wang2022ntk}. Coordinate
encodings address this bias by supplying the network with structured
input features~\cite{tancik2020,sitzmann2020}, and their
multiresolution variants place trainable features on grids of
increasing fineness~\cite{mueller2022,huang2024hash}, an idea whose
scale-splitting logic is familiar from multigrid
methods~\cite{briggs2000} and which has entered physics-informed
computation through domain-decomposed and multilevel
architectures~\cite{moseley2023,dolean2024}. Once the representation
is expressive enough to form the band, however, a second difficulty
appears, since an estimator built on a fixed point set offers the
optimizer configurations that lower the estimated energy while the
fields deteriorate between the points. The two difficulties feed each
other, as every gain in expressiveness sharpens the features the
estimator must observe, so a representation cured of its bias toward
smooth functions is precisely the one that endangers a fixed
quadrature.

In the energy setting the remedies adopted so far anchor
the method back to a mesh. In~\cite{manav2024} the field gradients are
computed numerically as in finite elements on a discretized domain
with a fixed Gauss rule and weight regularization, so that a mesh and
a fixed quadrature re-enter the discretization, and the adaptive
quadrature of~\cite{goswami2020fourth} responds to the same difficulty
with refinement machinery of the kind the network was meant to
remove, as does the subdomain refinement of~\cite{chakraborty2022}.
The B-spline enrichment of~\cite{plunge2026} follows the same
direction one step further, replacing automatic differentiation by
basis functions carrying the higher-order gradients, so that the
derivatives are again read from a fixed set of shape functions.
Where automatic differentiation is retained together with a fixed
quadrature grid, as in~\cite{dean2026}, the trial fields are kept
small enough that the freedom the grid cannot observe does not arise.
None of these devices is without consequences of its own. Weight
regularization has been reported to prevent a network from
representing the rapidly varying displacement close to the
crack~\cite{ning2023}, and an independent evaluation of the deep Ritz
treatment on samples containing ten to twenty interacting cracks,
carried out with element shape functions for the gradients and a
fixed Gauss rule for the integrals, returned a different crack
pattern for every random seed and none that agreed with the finite
element solution~\cite{hamdi2026}. A remedy of a different kind
alters the model instead of the discretization, decoupling the
regularization length from the physical process-zone scale through
modified degradation functions so that a coarser resolution
suffices~\cite{lian2023}. A more radical route discards the length
scale altogether, and reports crack paths that then depend on the
network parameters rather than on the material~\cite{konale2026}.
This work keeps the
standard model and resolves the physical band instead.

Both difficulties appear together in a form that is easily mistaken
for a matter of network design. Ghaffari Motlagh et
al.~\cite{motlagh2023} minimized the phase-field energy on a fixed
collocation set for the single-edge-notched tension and shear tests
and obtained a crack that advanced faster than the finite element
reference, at a rate that depended on which activation function was
used, with the piecewise linear choice reproducing the reference and
the smooth choices failing to do so. Since the same study reports
this rate to be insensitive to the depth and the width of the
network, the deficiency lies in the discretization rather than in the
capacity of the representation. A band that the trial fields cannot
resolve contributes too little fracture energy, and on a fixed point
set the deficit does not appear in the estimate, so the crack
advances at too small a cost in energy.

This work retains the mesh-free, automatic-differentiation form of the
deep energy method and addresses both difficulties within the
discretization itself. One network represents the displacement and
phase fields and receives the coordinates through a multiresolution
feature encoding built from $C^{1}$ quadratic B-spline grids, whose
finest level sets the finest scale the representation can express, so
resolving the band becomes a matter of choosing the level resolutions
rather than an outcome of slow training. The finest scale is
therefore fixed by the encoding rather than by the activation, and
the computations reported here use one smooth activation throughout. The incremental energy
is estimated by stratified Monte Carlo integration on points redrawn
at every optimizer iteration, so no fixed point set exists for the
optimizer to exploit, the estimate is unbiased by construction, and
resolution follows the crack since one stratum concentrates on the
damaged region. The crack therefore requires no treatment of its own
anywhere in the discretization. Nothing is refined ahead of the tip,
no degrees of freedom are inserted along the path, and the geometry
of the crack need not be anticipated before the computation, which is
the property that the refinement and remeshing devices reviewed above
are introduced to recover. Each of the two ingredients has precedents
of its own, and what this work contributes is their combination,
together with the evidence that neither of them suffices
alone.

Essential boundary conditions enter exactly through
lifts, and curved domains enter through an exact map built from
non-uniform rational B-splines (NURBS), with interior
holes carried by an indicator mask of the material region, in the
spirit of isogeometric
analysis~\cite{hughes2005,piegl1997}. Since the field
representation is globally $C^{1}$, the fourth-order density enters the same
discretization by exchanging one term of the energy and adding one
differentiation pass. This continuity holds across the entire domain,
so the fourth-order density requires none of the interface conditions
that a formulation on decomposed subdomains must
supply~\cite{chakraborty2022}.

The contributions of this work are the following. The width of the
crack band is set by the level resolutions of the encoding rather
than reached through training, so the band is representable from the
first iteration onward. Redrawing the integration points at every
iteration removes the fixed point set on which such a fine
representation would otherwise be free to deteriorate between the
points, and the two devices are shown to be required together, since
removing either one of them stops the crack from propagating.
Essential
boundary conditions and curved geometry are imposed exactly on a
single spline patch, with interior holes carried by a mask, and the
second- and the fourth-order fracture energy densities run on one and
the same discretization at the cost of one further differentiation
pass. The method is verified against a closed-form elastic solution
and against staggered finite element references at matched
regularization length, where the peak loads agree to within about
$1\%$ on the single-edge-notched tests, and it is then applied
without any per-configuration adjustment to a
public benchmark dataset on
which the deep Ritz baseline of its authors fails.

None of this is offered as a faster route to a single quasi-static
forward solution. A mature finite element implementation solves these
same problems at a lower cost per load increment, as the measurements
of \ref{app:hyper} report and as others have found for network-based
solvers generally~\cite{grossmann2024,ani2026}. What the
discretization offers in return is of a different kind. The crack needs
no representation of its own, so evolving and merging crack networks
require neither remeshing nor tracking; the fourth-order density,
which a $C^{0}$ element space cannot accommodate without a
higher-continuity construction, follows from one additional
differentiation pass; curved geometry is exact on a single patch; and
the same smooth representation together with a quadrature that fixes
no point set carries over to problems in which further fields are
solved alongside the two considered here.

The paper is organized as follows. Section~\ref{sec:formulation}
presents the phase-field model of brittle fracture in its second- and
fourth-order forms. Section~\ref{sec:method} describes the
discretization, the encoding and the resampled estimator.
Section~\ref{sec:results} reports six numerical examples, from the
single-edge-notched tests to a public benchmark dataset of random
multi-crack configurations. Section~\ref{sec:conclusions} concludes.

% =====================================================================
\section{Phase-field modeling of brittle fracture}
\label{sec:formulation}

Consider a homogeneous, isotropic, linearly elastic body occupying an
open bounded domain $\Omega \subset \R^{2}$, analyzed under plane strain
and quasi-static, displacement-controlled loading. Cracks are described by
a scalar phase field $\pf : \Omega \to [0,1]$, where $\pf = 0$ corresponds
to intact material and $\pf = 1$ to a fully developed crack; part of the
literature works with the complementary variable
$c = 1 - \pf$~\cite{borden2014}. Small strains are assumed throughout,
with the strain tensor
$\straintensor(\ub) = \tfrac{1}{2}(\grad\ub + \grad\ub^{\mathrm{T}})$
derived from the displacement field $\ub = (u, v)^{\mathrm{T}}$.
Table~\ref{tab:notation} collects the main symbols.

\begin{table}[htb]
    \centering
    \caption{Main symbols. Total energies are written with an uppercase
    $\Ene$, energy densities with a lowercase $\edens$.}
    \label{tab:notation}
    \footnotesize
    \begin{tabular}{llll}
        \toprule
        Symbol & Meaning & Symbol & Meaning \\
        \midrule
        $\Omega$, $\widehat{\Omega}$ & physical / parametric domain &
        $\mathcal{N}_{\bm{\theta}}$, $\bm{\theta}$ & network and its parameters \\
        $\ub$, $\straintensor$ & displacement vector, strain tensor &
        $\Fenc$, $\bm{G}_{\ell}$ & feature vector, feature grids \\
        $\pf$, $\pf_{0}$ & phase field, seeded profile &
        $h$ & finest grid spacing \\
        $\lreg$ & regularization length &
        $\bm{G}$, $\Jgeo$ & geometry map, its Jacobian \\
        $\Ene$, $\edens$, $\edens^{\pm}$ & total energy, elastic density, its parts &
        $\chi$ & domain mask \\
        $\Gc$, $\dmg$, $\kres$ & toughness, degradation, residual stiffness &
        $\varrho$, $M$ & sampling density, points per iteration \\
        $\psi_{\mathrm{c},2}$, $\psi_{\mathrm{c},4}$ & fracture energy densities &
        $n_{\mathrm{r}}$ & resampling interval \\
        $\delta$, $F$ & prescribed displacement, reaction &
        $s$ & phase-field squash function \\
        $\gamma_{\mathrm{ir}}$, $\tau$ & penalty weight, dead band &
        $U_{\mathrm{ref}}$ & displacement scale \\
        $\omega$ & boundary envelope of the lifts &
        $\widehat{D}$ & driving-force indicator of the sampler \\
        \bottomrule
    \end{tabular}
\end{table}

\FloatBarrier
% ---------------------------------------------------------------------
\subsection{Energy functional}
\label{subsec:energy}

The energetic description of brittle fracture goes back to Griffith, who
argued that a crack can only advance if the elastic energy released by
its advance supplies the fracture energy of the surface it
creates~\cite{griffith1921}.
Francfort and Marigo recast this balance as a minimization principle for
the total energy of the cracked body~\cite{francfort1998}, and Bourdin et
al.\ made the principle computable by regularizing the unknown crack set
into a phase field~\cite{bourdin2000,bourdin2008}. In this regularized
setting the total energy reads
\begin{equation}
    \Ene(\ub, \pf)
    =
    \int_{\Omega}
    \bigl[
        \dmg(\pf)\, \edens^{+}(\straintensor)
        + \edens^{-}(\straintensor)
    \bigr]
    \dV
    +
    \int_{\Omega}
    \psi_{\mathrm{c},n}(\pf)
    \dV ,
    \label{eq:pi}
\end{equation}
in which $\edens^{+} + \edens^{-} = \edens$ is a decomposition of the
elastic strain energy density into a degradable and a protected part
(Section~\ref{subsec:splits}) and $\psi_{\mathrm{c},n}$ is a fracture
energy density of order $n$ proportional to the
critical energy release rate $\Gc$. All loading enters through prescribed boundary displacements and
body forces are absent, so no external work term appears
in~Eq.~\eqref{eq:pi}. The strain energy density of the undamaged material is
\begin{equation}
    \edens(\straintensor)
    =
    \tfrac{1}{2}\lambda\, \trace^{2}\straintensor
    + \mu\, \straintensor : \straintensor ,
    \label{eq:psi}
\end{equation}
with the Lam\'e constants $\lambda$ and $\mu$, and the stiffness
degradation follows the standard quadratic
form~\cite{bourdin2000,miehe2010}
\begin{equation}
    \dmg(\pf) = (1 - \pf)^{2} + \kres ,
    \label{eq:g}
\end{equation}
where the small residual stiffness $\kres$ keeps the fully broken state
well-conditioned.

For the second-order model ($n = 2$) we use the quadratic fracture
energy density~\cite{ambrosio1990,bourdin2000,miehe2010}
\begin{equation}
    \psi_{\mathrm{c},2}(\pf, \grad\pf)
    =
    \frac{\Gc}{2\lreg}
    \bigl(
        \pf^{2} + \lreg^{2}\, |\grad\pf|^{2}
    \bigr) ,
    \label{eq:at2}
\end{equation}
where $\lreg$ is the regularization length that controls the width of the
diffuse crack. After the approximations of Ambrosio and
Tortorelli~\cite{ambrosio1990}, this density is commonly denoted AT2.
Every computation reported here uses it or its fourth-order
counterpart below. In one dimension, the transverse profile of a fully
developed crack at $x = 0$ follows from minimizing
$\int \psi_{\mathrm{c},2}\,\mathrm{d}x$ subject to $\pf(0) = 1$ and
decay at infinity. The optimality condition
$\pf - \lreg^{2} \pf'' = 0$ then yields
\begin{equation}
    \pf_{2}(x) = \exp\bigl(-|x|/\lreg\bigr) .
    \label{eq:profile2}
\end{equation}
Inserting~Eq.~\eqref{eq:profile2} into~Eq.~\eqref{eq:at2} and integrating
across the band gives exactly $\Gc$, so that the second term
of~Eq.~\eqref{eq:pi} measures $\Gc$ per unit crack length. As $\lreg \to 0$
the regularized energy $\Gamma$-converges to the Griffith
functional~\cite{ambrosio1990,bourdin2000}. At finite $\lreg$ the model
is best understood as a gradient damage model with an internal
length~\cite{pham2011,tanne2018}, and $\lreg$ acts as a model
parameter; all comparisons of Section~\ref{sec:results} are made
between solutions computed at the same $\lreg$.

Borden et al.\ proposed a fourth-order extension whose minimizers are
smoother and whose surface energy is approximated more
accurately~\cite{borden2014}. Translated to our convention, in which
their phase field is $c = 1 - \pf$ and their length scale equals
$\lreg/2$ (see \ref{app:fourth}), the density reads
\begin{equation}
    \psi_{\mathrm{c},4}(\pf, \grad\pf, \lap\pf)
    =
    \frac{\Gc}{2\lreg}
    \Bigl(
        \pf^{2}
        + \frac{\lreg^{2}}{2}\, |\grad\pf|^{2}
        + \frac{\lreg^{4}}{16}\, (\lap\pf)^{2}
    \Bigr) .
    \label{eq:fourth}
\end{equation}
The corresponding one-dimensional optimality condition,
$\pf - \tfrac{1}{2}\lreg^{2}\pf'' + \tfrac{1}{16}\lreg^{4}\pf'''' = 0$,
admits the optimal profile
\begin{equation}
    \pf_{4}(x)
    =
    \exp\bigl(-2|x|/\lreg\bigr)\bigl(1 + 2|x|/\lreg\bigr)
    \label{eq:profile4}
\end{equation}
under the additional condition $\pf'(0) = 0$. The normalization is
shared, since $\int \psi_{\mathrm{c},4}\,\mathrm{d}x = \Gc$ on the
profile, and both models therefore dissipate $\Gc$ per unit crack length
by construction.
The elementary computations behind Eq.~\eqref{eq:profile4} and the numerical
verification of the normalization in our implementation are collected in
\ref{app:fourth}.

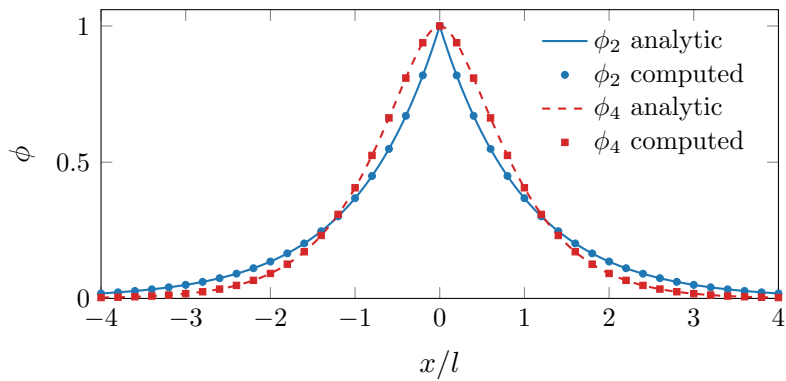
\begin{figure}[htb]
    \centering
    \begin{tikzpicture}
        \begin{axis}[
            paperaxis,
            width=0.62\linewidth, height=5.4cm,
            xlabel={$x/\lreg$}, ylabel={$\pf$},
            xmin=-4, xmax=4, ymin=0, ymax=1.06,
            legend pos=north east,
        ]
            \addplot[cblue, thick]
                table[col sep=comma, x=xl, y=phi2]
                {figures/data/profiles_analytic.csv};
            \addlegendentry{$\pf_{2}$ analytic}
            \addplot[cblue, only marks, mark=*, mark size=1.3pt]
                table[col sep=comma, x=xl, y=phi]
                {figures/data/profiles_pf2.csv};
            \addlegendentry{$\pf_{2}$ computed}
            \addplot[cred, dashed, thick]
                table[col sep=comma, x=xl, y=phi4]
                {figures/data/profiles_analytic.csv};
            \addlegendentry{$\pf_{4}$ analytic}
            \addplot[cred, only marks, mark=square*, mark size=1.2pt]
                table[col sep=comma, x=xl, y=phi]
                {figures/data/profiles_pf4.csv};
            \addlegendentry{$\pf_{4}$ computed}
        \end{axis}
    \end{tikzpicture}
    \caption{Optimal crack profiles of the second-order
    density of Eq.~\eqref{eq:at2} and the fourth-order density of Eq.~\eqref{eq:fourth}
    at equal regularization length. Lines show the analytic
    profiles of Eq.~\eqref{eq:profile2} and~\eqref{eq:profile4}; markers show
    transverse cuts through the relaxed bands computed by the solver of
    Section~\ref{sec:method}. The second-order profile has a slope
    discontinuity at the crack, the fourth-order profile is $C^{1}$
    there and decays faster; both dissipate $\Gc$ per unit crack
    length.}
    \label{fig:profiles}
\end{figure}

The two densities are compared in Fig.~\ref{fig:profiles}. The
second-order profile of Eq.~\eqref{eq:profile2} has a slope
discontinuity at the crack. The fourth-order
profile of Eq.~\eqref{eq:profile4} is $C^{1}$ there and decays faster, so that
for equal $\lreg$ the fourth-order band is visibly more compact, and
Borden et al.\ report improved accuracy of the computed surface energy
together with smoother stress fields adjacent to the
band~\cite{borden2014}. These benefits come at a regularity price. The
term $(\lap\pf)^{2}$ is finite only for $\pf \in H^{2}(\Omega)$, so a
conforming Galerkin discretization requires $C^{1}$ trial functions,
realized in~\cite{borden2014} through isogeometric spline spaces and
otherwise available only through Hermite constructions or mixed
reformulations. This requirement is the main reason second-order models
dominate finite element practice~\cite{wu2020}. A network
representation removes the barrier, and two routes have been taken.
The regularity can be carried by the basis, as in the higher-order
B-spline enrichment of Plung\.{e} et al.~\cite{plunge2026}, or the
derivatives can be taken from the representation itself, as in the
fourth-order multiphysics model of Dean and
Bahtiri~\cite{dean2026}. In the framework
developed in Section~\ref{sec:method} the fields are $C^{1}$ by
construction and the
Laplacian is obtained by one further automatic-differentiation pass,
so the regularity and the derivatives come from the same
representation and no separate construction is needed for either. The
two models then run on identical discretizations and differ by a single
term in the energy density, the choice between them becomes a
modeling question, and the numerical examples report both.

The strong form follows from~Eq.~\eqref{eq:pi} by stationarity.
Stationarity of $\Ene$ with respect to the
displacement field yields the equilibrium equation
\begin{equation}
    \dive \stresstensor = \bm{0},
    \qquad
    \stresstensor
    =
    \dmg(\pf)\, \frac{\partial \edens^{+}}{\partial \straintensor}
    + \frac{\partial \edens^{-}}{\partial \straintensor} ,
    \label{eq:equilibrium}
\end{equation}
and stationarity with respect to the phase field yields, for the two
densities,
\begin{align}
    \frac{\Gc}{\lreg}
    \bigl( \pf - \lreg^{2} \lap\pf \bigr)
    &=
    2 (1 - \pf)\, \edens^{+}
    && (n = 2),
    \label{eq:strongphi2}
    \\
    \frac{\Gc}{\lreg}
    \Bigl( \pf - \tfrac{1}{2}\lreg^{2} \lap\pf
           + \tfrac{1}{16}\lreg^{4} \lap^{2}\pf \Bigr)
    &=
    2 (1 - \pf)\, \edens^{+}
    && (n = 4),
    \label{eq:strongphi4}
\end{align}
a second-order and a fourth-order partial differential equation,
respectively, the latter giving the model its name. The associated
natural boundary conditions are $\grad\pf \cdot \bm{n} = 0$ for $n = 2$
and involve, in addition, the normal derivative of $\lap\pf$ for
$n = 4$~\cite{borden2014}. In a residual-based method these higher-order
conditions must be built into the scheme. In an energy-minimization
method they are never imposed at all, since
minimizing~Eq.~\eqref{eq:pi} over an unconstrained representation does not
require them to be stated. This is a practical advantage of working with the
energy rather than with~Eq.~\eqref{eq:strongphi2}
and~\eqref{eq:strongphi4}.

Since crack growth is irreversible, the phase field must not decrease as
the load increases. Loading is applied incrementally through a monotone
sequence of prescribed boundary displacements
$\delta_{1} < \delta_{2} < \dots < \delta_{N}$, and at step $n$ we
require $\pf \ge \pf_{n-1}$ at every point, where $\pf_{n-1}$ is the
converged field of
the previous step. The constraint is enforced by adding to the energy
density the one-sided quadratic penalty
\begin{equation}
    \psiir
    =
    \gamma_{\mathrm{ir}}\,
    \pospart{\pf_{n-1} - \tau - \pf}^{2} ,
    \label{eq:penalty}
\end{equation}
with penalty weight $\gamma_{\mathrm{ir}}$ and a small dead band
$\tau \ge 0$ that prevents integration noise from slowly ratcheting the
far field upward over many load steps (parameter values are given in
\ref{app:hyper}). A widespread alternative is the history-field
substitution of Miehe et al.~\cite{miehe2010cmame}, in which the crack
driving force is replaced by its maximum over the loading history; it is
also common in energy-based network solvers~\cite{goswami2020tl,
goswami2020fourth}. We prefer the penalty for two reasons. It keeps each
load step the minimization of a single well-defined energy, the property
on which the proposed solver rests, whereas the history substitution
modifies the governing equations in a way that corresponds to no energy;
a detailed analysis of penalized irreversibility is given by Gerasimov
and De Lorenzis~\cite{gerasimov2019}. It also requires the previous
phase field only as a pointwise-evaluable field rather than as state
stored at integration points, a distinction that becomes essential once
the integration points change during the solve
(Section~\ref{subsec:quadrature}).

The $n$-th incremental problem then reads
\begin{equation}
    (\ub_{n}, \pf_{n})
    =
    \arg\min
    \Bigl\{
        \Ene(\ub, \pf; \delta_{n})
        + \int_{\Omega} \psiir\, \dV
        \;:\;
        \ub \ \text{admissible for } \delta_{n}
    \Bigr\} ,
    \label{eq:incremental}
\end{equation}
where admissibility refers to the essential boundary conditions at load
level $\delta_{n}$, which the trial fields of
Section~\ref{subsec:fields} satisfy exactly; the one boundary of
Section~\ref{sec:results} where they do not is treated there.

\FloatBarrier
% ---------------------------------------------------------------------
\subsection{Strain energy decomposition and hybrid formulation}
\label{subsec:splits}

The decomposition $\edens = \edens^{+} + \edens^{-}$ in~Eq.~\eqref{eq:pi}
decides which portion of the strain energy drives, and is degraded by,
damage. Without any decomposition ($\edens^{+} = \edens$,
$\edens^{-} = 0$) compressive and shear states damage the material as
readily as tension does. This is usually undesirable, but not always.
Fully isotropic driving is what permits a fast crack to branch
symmetrically, and we retain the undecomposed model for the branching
example of Section~\ref{subsec:branching}, in line with its use in
dynamic branching studies~\cite{borden2012}.

For the remaining examples a unilateral decomposition is needed, and two
classical constructions are considered. The spectral decomposition of
Miehe et al.~\cite{miehe2010cmame} separates the principal strains,
\begin{equation}
    \edens^{\pm}(\straintensor)
    =
    \tfrac{1}{2}\lambda \bigl\langle \trace\straintensor \bigr\rangle_{\pm}^{2}
    + \mu \sum_{i} \bigl\langle \varepsilon_{i} \bigr\rangle_{\pm}^{2} ,
    \label{eq:spectral}
\end{equation}
where $\varepsilon_{i}$ are the principal strains and
$\langle\cdot\rangle_{\pm}$ denote the positive and negative parts;
under plane strain the out-of-plane principal strain vanishes
identically and drops out of the sums. In pure shear the principal
strains come in opposite pairs, so only half of the shear energy drives
damage under Eq.~\eqref{eq:spectral}, a property that matters for the
mode II example and is examined in \ref{app:failure}. The
volumetric--deviatoric decomposition of Amor et al.~\cite{amor2009}
instead degrades volumetric expansion and all deviatoric deformation,
\begin{equation}
    \edens^{+}
    =
    \tfrac{1}{2} K \pospart{\trace\straintensor}^{2}
    + \mu\, \devop\straintensor : \devop\straintensor ,
    \qquad
    \edens^{-}
    =
    \tfrac{1}{2} K \negpart{\trace\straintensor}^{2} ,
    \label{eq:amor}
\end{equation}
with the bulk modulus $K = \lambda + 2\mu/3$. This split keeps shear fully
damage-driving, but for the same reason it lets the shear-rich stress
concentration at a fixed-edge corner nucleate damage there, an effect that
becomes visible at the resolution reached in this work and is likewise
documented in \ref{app:failure}. Generalizations such as the star-convex decomposition
of Vicentini et al.~\cite{vicentini2024} modulate the
compressive-volumetric contribution and contain Eq.~\eqref{eq:amor} as a
special case; in exploratory runs the additional parameter had no
influence on the behavior relevant here, and we retain the classical
forms.

Our default is the hybrid formulation proposed by Ambati et
al.~\cite{ambati2015}. The mechanical response is governed by isotropic
degradation, so that a developed crack is compliant in all deformation
modes, while the evolution of $\pf$ is driven by the spectral tensile
energy alone, so that compressed material does not damage. The scheme is
defined by the staggered pair
\begin{align}
    \ub^{k+1}
    &=
    \arg\min_{\ub} \int_{\Omega}
    \dmg(\pf^{k})\, \edens(\straintensor)\dV ,
    \label{eq:hybrid-u}
    \\
    \pf^{k+1}
    &=
    \arg\min_{\pf} \int_{\Omega}
    \Bigl[
        \dmg(\pf)\,
        \psi_{\mathrm{e,spec}}^{+}\bigl(\straintensor(\ub^{k+1})\bigr)
        + \psi_{\mathrm{c},n} + \psiir
    \Bigr]\dV .
    \label{eq:hybrid-phi}
\end{align}
Since the solver of Section~\ref{sec:method} performs a single
gradient-based minimization per load step, we fold the pair into one
objective by freezing the cross couplings. Let $\sg{\cdot}$ denote the
stop-gradient operation, whose argument is treated as a constant during
differentiation. The hybrid elastic energy density is implemented as
\begin{equation}
    \psihyb
    =
    \dmg\bigl(\sg{\pf}\bigr)\, \edens(\straintensor)
    +
    \Bigl( \dmg(\pf) - \sg{\dmg(\pf)} \Bigr)\,
    \sg{\psi_{\mathrm{e,spec}}^{+}(\straintensor)} .
    \label{eq:hybrid-folded}
\end{equation}
Differentiating~Eq.~\eqref{eq:hybrid-folded} with respect to the
displacement degrees of freedom recovers~Eq.~\eqref{eq:hybrid-u},
differentiating with respect to the phase-field degrees of freedom
recovers~Eq.~\eqref{eq:hybrid-phi}, and the two updates advance together in
each optimizer iteration. We verified that the resulting parameter
gradients agree with those of an explicitly staggered implementation to
machine precision. The numerical value of~Eq.~\eqref{eq:hybrid-folded}
equals the degraded isotropic strain energy, so energy logs and the
reaction force retain their mechanical meaning. The original
formulation supplements the scheme with a
condition that prevents interpenetration of closed crack faces; the
examples considered are opening- and sliding-dominated, no face
closure occurs, and the condition is omitted. Like the
original, the hybrid scheme does not minimize a single energy, an
inconsistency accepted here as it is in the phase-field
literature~\cite{ambati2015,wu2020}.

\FloatBarrier
% ---------------------------------------------------------------------
\subsection{Pre-existing cracks, loading and reaction force}
\label{subsec:setup}

Most of the cases considered in Section~\ref{sec:results} are modeled
on specimen geometries that contain
pre-existing cracks, either an edge notch or internal flaws. We
represent them diffusely rather than by slitting the domain. The initial
phase field is prescribed as the optimal profile of the governing
density about the crack segments, that is $\pf_{0} = \pf_{2}(d)$ or
$\pf_{4}(d)$ from~Eq.~\eqref{eq:profile2} and~\eqref{eq:profile4} with
$d(\xb)$ being the distance to the nearest segment, and the same function
seeds the field representation of Section~\ref{subsec:fields}. The
crack thereby enters at its energy-consistent diffuse state, and no
artificial equilibration transient occurs at the first load step.
Several pre-existing cracks are handled by taking the minimum distance
over all segments, which is how the coalescence example of
Section~\ref{subsec:coalescence} is set up. Traction-free outer
boundaries require no treatment, being natural boundary conditions
of~Eq.~\eqref{eq:pi}.

The structural response is reported as a load--displacement curve. Let
$\Ene^{*}(\delta) = \min \Ene(\ub, \pf; \delta)$ denote the minimized
energy at load level $\delta$, the minimum being taken over admissible
fields. The reaction force work-conjugate to the prescribed displacement
is
\begin{equation}
    F(\delta)
    =
    \frac{\mathrm{d}\Ene^{*}}{\mathrm{d}\delta}
    =
    \left.
    \frac{\partial \Ene}{\partial \delta}
    \right|_{(\ub,\pf)\,=\,\text{minimizer}} ,
    \label{eq:force}
\end{equation}
where the second equality holds since the energy is stationary at the
minimizer, so only the explicit dependence on $\delta$, carried by the
boundary lift of the displacement representation, survives. By work
conjugacy, Eq.~\eqref{eq:force} equals the resultant of the tractions along
the loaded edge,
\begin{equation}
    F(\delta)
    =
    \int_{\Gamma_{\mathrm{load}}}
    \bigl( \stresstensor \cdot \bm{n} \bigr) \cdot \bm{e}_{\mathrm{load}}
    \dA ,
    \label{eq:forceboundary}
\end{equation}
with $\bm{e}_{\mathrm{load}}$ the unit direction of the prescribed
displacement. We evaluate Eq.~\eqref{eq:force} rather
than~Eq.~\eqref{eq:forceboundary}. The energetic form requires no boundary
quadrature, involves no stress post-processing on a set of measure zero,
and under the Monte Carlo integration of
Section~\ref{subsec:quadrature} it inherits the variance of a domain
integral rather than that of a boundary one; in the discrete setting the
derivative is evaluated exactly by automatic differentiation
(Section~\ref{subsec:training}).

A last ingredient concerns the corners where a fixed edge meets a
loaded edge. The linear elastic solution carries a wedge-type stress
singularity at a corner where the boundary conditions change, whose
order follows from the wedge angle and the conditions on the two faces
in the classical analysis of Williams~\cite{williams1952} and is a
property of the boundary value problem rather than of any
discretization. A phase-field solver that resolves the length $\lreg$
near such a corner therefore sees an unbounded driving force and
nucleates damage there. A coarser discretization does not remove this;
it merely fails to resolve the singular field, so the damage does not
appear in the computation while the singularity remains in the problem.
Since the corner is an artifact of how the specimen is held rather than
a feature of the material, we prescribe a locally elevated toughness at
the fixed-edge corners,
\begin{equation}
    \Gc(\xb) = \Gc \bigl[ 1 + \beta\, b(\xb) \bigr] ,
    \label{eq:gcgrip}
\end{equation}
where $b$ is a smooth bump equal to one at the corner points and
decaying over a radius of a few $\lreg$, and $\beta = 1$ wherever the
device is applied; the runs that use it are listed in \ref{app:hyper}.
This is a numerical device at the grips rather than a material
statement; it keeps the
fracture energy smooth and the incremental problems unchanged in
structure, and the damage it suppresses is documented in
\ref{app:failure}. Its side effects on the post-failure response are noted in
Section~\ref{sec:conclusions}.

% =====================================================================
\section{Mesh-free deep energy method}
\label{sec:method}

The incremental problems of Eq.~\eqref{eq:incremental} are solved with a deep
energy method. The fields are represented by a neural network, the
energy is estimated by Monte Carlo integration, and the network
parameters are updated by a first-order optimizer until the estimated
energy plateaus~\cite{e2018deepritz,samaniego2020,nguyenthanh2020}.
Energy-based losses of this kind involve lower-order derivatives than
strong-form residuals and inherit the natural boundary conditions of the
underlying functional~\cite{samaniego2020,kharazmi2021}, which makes
them a natural fit for the minimization structure of
Section~\ref{sec:formulation}.

\FloatBarrier
% ---------------------------------------------------------------------
\subsection{Field representation and boundary conditions}
\label{subsec:fields}

The discrete fields are built in two layers. A neural network supplies
three raw outputs, and closed-form constructions turn these outputs into
displacement and phase fields that satisfy the essential boundary
conditions exactly and carry the pre-existing crack.

The network is shared by all fields,
\begin{equation}
    \mathcal{N}_{\bm{\theta}} :\;
    \bigl(\xib,\, \Fenc(\xib)\bigr)
    \;\longmapsto\;
    \bigl(\hat{u},\, \hat{v},\, \hat{\pf}\bigr) ,
    \label{eq:net}
\end{equation}
taking the parametric coordinates $\xib = (\xi, \eta) \in [0,1]^{2}$
together with the
feature vector $\Fenc$ of Section~\ref{subsec:encoding}, and returning
two raw displacement components $\hat{u}$, $\hat{v}$ and one raw
phase-field output $\hat{\pf}$. The
network itself is a standard multilayer perceptron. With input
$\bm{z}^{0} = (\xib, \Fenc(\xib))$, the hidden layers apply the
recursion
\begin{equation}
    \bm{z}^{k}
    =
    \sigma\bigl( \bm{W}^{k} \bm{z}^{k-1} + \bm{b}^{k} \bigr),
    \qquad k = 1, \dots, D,
    \label{eq:mlp}
\end{equation}
followed by a linear output layer, with $\sigma$ the GELU activation.
The trainable parameters $\bm{\theta}$ collect the weights $\bm{W}^{k}$,
the biases $\bm{b}^{k}$ and the feature grids of the encoding. All
computations use $D = 4$ hidden layers of width 128; the full
architecture data are tabulated in \ref{app:hyper}. Note that the network is small,
since the spatial resolution of the method resides in the feature grids
rather than in the network itself, and nothing in the formulation is
tied to this particular architecture. Sharing one
network among the fields lets the displacement and phase-field channels
use the same encoding and keeps the whole state in a single optimizer.

The problem is solved in nondimensional form. Coordinates live on the
unit parametric domain, displacements are scaled by a reference
magnitude $U_{\mathrm{ref}}$ chosen so that the raw network outputs and
the prescribed displacement are of order one, and all reported
quantities are converted back to physical units. Scalings of this kind
are routine in network-based solvers, where badly scaled inputs or
outputs distort the optimization landscape long before they would affect
a linear solver~\cite{karniadakis2021}.

Essential boundary conditions are satisfied exactly by construction, in
the manner introduced by Lagaris et al.~\cite{lagaris1998} and developed
systematically in~\cite{sukumar2022}. The raw outputs are multiplied by
a function that vanishes on the Dirichlet boundary and added to a lift
that interpolates the prescribed data. For a specimen with a fixed
bottom edge ($y = 0$) and a displacement-driven top edge ($y = 1$),
loaded in shear, the displacement fields read
\begin{subequations}
\label{eq:lift}
\begin{align}
    u &= U_{\mathrm{ref}}
    \bigl[ \omega(y)\, \hat{u} + y\, \delta \bigr] ,
    \label{eq:liftu}
    \\
    v &= U_{\mathrm{ref}}\, \omega(y)\, \hat{v} ,
    \label{eq:liftv}
\end{align}
\end{subequations}
where $\omega(y) = y(1-y)$ is an envelope that vanishes on both grips; for
tension the linear lift $y\,\delta$ moves to the $v$ component. The
fixed edge and the prescribed displacement thus hold identically for
any network parameters, and no boundary penalty term, and hence no
penalty-weight tuning, is needed. Traction-free lateral boundaries are natural
boundary conditions of the energy and require no treatment.

The phase field is assembled around the seeded pre-existing crack of
Section~\ref{subsec:setup},
\begin{equation}
    \pf(\xib)
    =
    \pf_{0}(\xib) + \bigl[ 1 - \pf_{0}(\xib) \bigr]\,
    s\bigl(\hat{\pf}(\xib)\bigr) ,
    \label{eq:phi-ansatz}
\end{equation}
where $\pf_{0}$ is the optimal profile of the governing density about
the crack segments and $s(\hat{\pf}) = 1/(1 + e^{-\hat{\pf}})$ is a logistic
squash. The output bias of the $\hat{\pf}$ channel is initialized at
$-4$, so that $s \approx 0.02$ at the start of training and the phase
field begins within $2\%$ of the seeded profile. Three properties
of~Eq.~\eqref{eq:phi-ansatz} are used repeatedly. The pre-existing crack
carries $\pf = 1$ exactly, independently of the network. The first load
step starts from the energy-consistent diffuse state rather than from an
arbitrary one, and in practice needs no special treatment beyond a
larger iteration budget. Finally, since $s \in (0,1)$, the
representation satisfies $\pf \ge \pf_{0}$ pointwise, so the seeded
crack cannot heal regardless of the state of the penalty
term of Eq.~\eqref{eq:penalty}. The penalty is not made redundant by this
bound, which involves the seeded profile only; irreversibility with
respect to the previous converged state, including whatever has grown
during loading, is exactly what Eq.~\eqref{eq:penalty} enforces.

\FloatBarrier
% ---------------------------------------------------------------------
\subsection{Multiresolution feature encoding}
\label{subsec:encoding}

Used directly on the raw coordinates, a multilayer perceptron is a poor
match for phase-field solutions. Networks of this kind are biased toward
smooth, slowly varying functions~\cite{rahaman2019,wang2022ntk}, while
the solution consists of smooth fields interrupted by a damage band of
width $O(\lreg)$, two orders of magnitude below the specimen size. The
established remedies enrich the input with global oscillatory features,
either Fourier feature embeddings~\cite{tancik2020} or periodic
activations~\cite{sitzmann2020}. These embeddings add fine-scale
content, but they add it globally, and differentiating globally
supported oscillatory features across a steep gradient produces
spurious spatial oscillations of the kind associated with the Gibbs
phenomenon, as documented for phase-field modeling of brittle fracture
in~\cite{manav2024}. We
therefore attach the fine-scale capacity to local, trainable degrees of
freedom instead, namely learnable feature grids at several resolutions,
read by smooth interpolation. Multiresolution grid encodings of this
kind were popularized in computer graphics by M\"uller et
al.~\cite{mueller2022} and have recently been used to accelerate
physics-informed networks~\cite{huang2024hash}. The variant used here
differs in two respects that matter for energy minimization, dense
rather than hashed grids and $C^{1}$ rather than $C^{0}$
interpolation. The second difference is essential. With the $C^{0}$
linear interpolation of hash encodings the derivatives are
discontinuous, to the point that Huang and Alkhalifah replace automatic
differentiation by finite differences~\cite{huang2024hash}, whereas the
quadratic interpolation used here keeps exact automatic differentiation
available, which the kinematics of Section~\ref{subsec:autodiff} relies
on.

The encoding consists of $L$ levels of feature grids
$\bm{G}_{\ell} \in \R^{C \times n_{\ell} \times n_{\ell}}$ over the
parametric domain, with $C$ channels per level. The feature vector
stacks the per-level contributions,
\begin{equation}
    \Fenc(\xib)
    =
    \bigl( \bm{f}_{1}(\xib),\, \bm{f}_{2}(\xib),\, \dots,\,
           \bm{f}_{L}(\xib) \bigr) ,
    \label{eq:encoding}
\end{equation}
and each contribution interpolates its grid over the local
$3 \times 3$ stencil,
\begin{equation}
    \bm{f}_{\ell}(\xib)
    =
    \sum_{a,b \in \{0,1,2\}}
    w_{a}(t_{\xi})\, w_{b}(t_{\eta})\;
    \bm{G}_{\ell}[\,\cdot\,,\, j{+}b,\, i{+}a] ,
    \label{eq:interp}
\end{equation}
where $(i,j)$ indexes the grid cell containing $\xib$ and
$(t_{\xi}, t_{\eta}) \in [0,1)^{2}$ are the local coordinates within
it. The
interpolation weights are those of the uniform quadratic B-spline,
\begin{equation}
    w_{0}(t) = \tfrac{1}{2}(1-t)^{2},
    \qquad
    w_{1}(t) = -t^{2} + t + \tfrac{1}{2},
    \qquad
    w_{2}(t) = \tfrac{1}{2} t^{2} .
    \label{eq:qspline}
\end{equation}
They form a partition of unity and are continuously differentiable
across cell boundaries, so the features, and with them every mechanical
field, are $C^{1}$. The input dimension of the perceptron is $2 + LC$,
and the cost of the encoding lookup, nine multiply-adds per level and
channel, is negligible next to the network evaluation itself.

The grids are zero-initialized. The encoding starts as an inert
augmentation, the network initially sees only the raw coordinates, and
features grow as residuals where the optimization requires them. In
trained models the coarse grids carry the smooth large-scale response,
while the finest grid concentrates its amplitude in the
crack band. The grids train at a larger learning rate than the
network; the values are listed in \ref{app:hyper}, and the runs
reported here carry no weight regularization on the network. The
feature grids alone carry a penalty
$10^{-8} \sum_{\ell} \| \bm{G}_{\ell} \|^{2}$ on the
squared feature amplitudes, which bounds the amplitudes away
from unbounded drift and is far too small to act on the solution
itself.

Since the resolution limit is set by the finest level
alone, one may ask why a hierarchy is used rather than a single fine
grid. The reason lies in how gradient-based training distributes work
across scales. Networks and trainable feature grids learn large-scale
content first and fine-scale content
late~\cite{rahaman2019,wang2022ntk}, so a single fine grid would have to
assemble the smooth large-scale response from its many small cells,
which is slow to train. A hierarchy instead assigns the smooth part to
coarse levels, which represent it with few parameters, and leaves the
finest level to carry only the fine-scale residual, in close analogy
with the coarse-grid correction of multigrid
methods~\cite{briggs2000}. The same design choice has proven itself
elsewhere, with multiresolution grids outperforming single-resolution
ones at comparable parameter counts in neural
graphics~\cite{mueller2022}, and multilevel architectures accelerating
the training of physics-informed
networks~\cite{moseley2023,dolean2024,huang2024hash}. Since the coarse
grids are small compared with the finest one, the additional cost of
the hierarchy is negligible.

The order of the interpolation matters, since it ties the encoding to
the models of Section~\ref{sec:formulation}. With $C^{0}$ linear
interpolation, as in hash-grid encodings~\cite{mueller2022}, strains and
$\grad\pf$ would jump across cell boundaries and the fourth-order term
$(\lap\pf)^{2}$ would not be integrable at all. With quadratic
interpolation the fields are $C^{1}$, strains and the gradient term of
the fracture energy density are continuous, and the second derivatives
of the features are piecewise constant, so the second derivatives of
the fields are bounded and $\lap\pf$ is square-integrable and the
fourth-order energy of Eq.~\eqref{eq:fourth} is finite and well-posed on the
representation. We note that $C^{1}$ quadratic splines are precisely the
minimal space Borden et al.\ employed for the same functional in the
isogeometric setting~\cite{borden2014}. A cubic, $C^{2}$ variant of the
encoding is the natural fallback should $(\lap\pf)^{2}$ ever exhibit
cell-scale noise; in the fourth-order computations reported here it does
not.

The resolutions $n_{\ell}$ grow from coarse to fine, and the
finest level plays the central role. Its spacing $h$ is prescribed as a
fixed
fraction of the regularization length, small enough that the optimal
profile of Eq.~\eqref{eq:profile2} decays over several cells. The
examples use $h = 0.26\lreg$, so that about four features span one
decay length $\lreg$ of the profile and roughly fifteen span the
half-width $4\lreg$ over which it falls below $2\%$ of its peak; the
per-level resolutions of every run are listed in
\ref{app:hyper}. The spacing $h$ is at the same time a hard limit on
the finest scale the representation can express, a property we refer to
as the resolution cap. The features are piecewise quadratic on fixed
knots, so no localized degree of freedom anywhere in the representation
can vary on scales below $h$. The band is thereby representable and
sharp, while oscillations of the kind produced by global oscillatory
features cannot be expressed at all. Matching $h$ to $\lreg$ rather than to
the specimen size is what keeps the grids affordable. What the cap does
not do by itself is protect a fixed
set of integration points from being exploited by the freedom that
remains above $h$
(Sections~\ref{subsec:quadrature} and~\ref{subsec:pairing}).

\FloatBarrier
% ---------------------------------------------------------------------
\subsection{Geometry map and domain mask}
\label{subsec:iga}

Curved domains are treated in the spirit of isogeometric analysis, in
which the geometry is represented exactly by the spline description used
in computer-aided design and the same parametrization carries the
analysis~\cite{hughes2005,cottrell2009}. Only the geometric half of
this idea is adopted here. The computational domain is the image of a parametric unit
square $\widehat{\Omega}$ under a fixed NURBS map, while the solution
fields continue to live on $\widehat{\Omega}$ through the representation
of Sections~\ref{subsec:fields} and~\ref{subsec:encoding}.

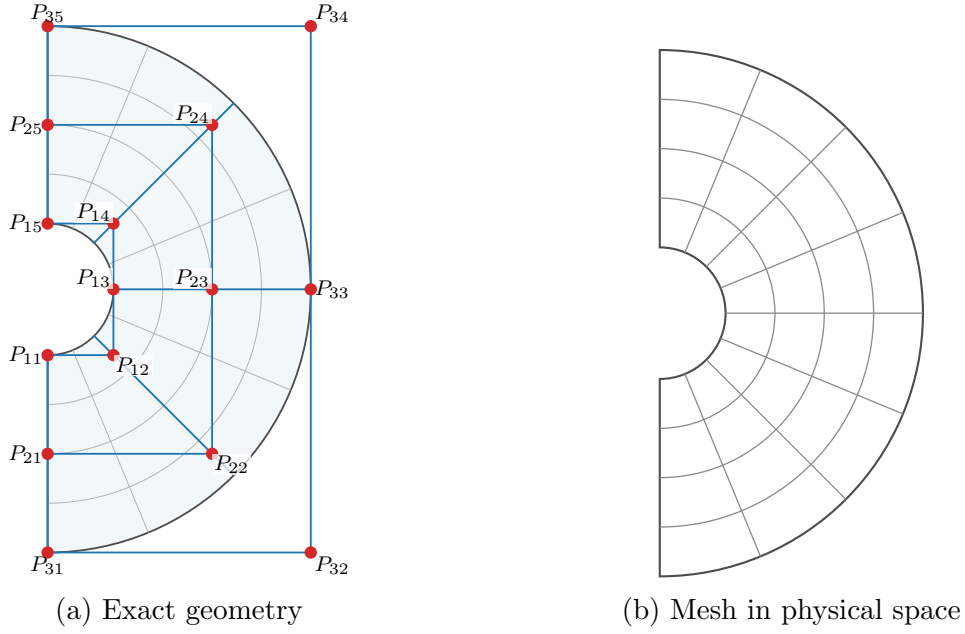
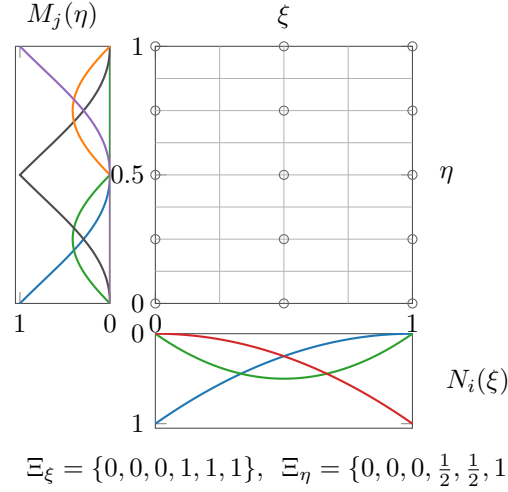
\begin{figure}[htb]
    \centering
    \begin{subfigure}[b]{0.48\linewidth}
        \centering
        \begin{tikzpicture}[scale=1.45]
            \draw[spec] (0, -2.4) arc (-90:90:2.4)
                -- (0, 0.6) arc (90:-90:0.6) -- cycle;
            \foreach \t in {0.25, 0.5, 0.75} {
                \draw[black!30, line width=0.3pt]
                    (0, {-(0.6+\t*1.8)}) arc (-90:90:{0.6+\t*1.8});
            }
            \foreach \a in {-67.5, -45, -22.5, 0, 22.5, 45, 67.5} {
                \draw[black!30, line width=0.3pt]
                    ({0.6*cos(\a)}, {0.6*sin(\a)})
                    -- ({2.4*cos(\a)}, {2.4*sin(\a)});
            }
            % control polygon, three rows (radial order elevated)
            \foreach \r in {0.6, 1.5, 2.4} {
                \draw[cblue, line width=0.7pt]
                    (0, -\r) -- (\r, -\r) -- (\r, 0) -- (\r, \r)
                    -- (0, \r);
            }
            \foreach \i in {0, 1, 2, 3, 4} {
                \draw[cblue, line width=0.7pt]
                    ({0.6*cos(-90+\i*45)}, {0.6*sin(-90+\i*45)})
                    -- ({1.5*cos(-90+\i*45)}, {1.5*sin(-90+\i*45)})
                    -- ({2.4*cos(-90+\i*45)}, {2.4*sin(-90+\i*45)});
            }
            \foreach \pt/\lab/\pos in {
                {(0,-2.4)}/{P_{31}}/below,
                {(2.4,-2.4)}/{P_{32}}/{below right},
                {(2.4,0)}/{P_{33}}/right,
                {(2.4,2.4)}/{P_{34}}/{above right},
                {(0,2.4)}/{P_{35}}/above,
                {(0,-1.5)}/{P_{21}}/left,
                {(0,1.5)}/{P_{25}}/left,
                {(0,-0.6)}/{P_{11}}/left,
                {(0,0.6)}/{P_{15}}/left} {
                \fill[cred] \pt circle (1.6pt);
                \node[font=\scriptsize, \pos, inner sep=1.3pt]
                    at \pt {$\lab$};
            }
            \foreach \pt/\lab/\pos in {
                {(1.5,-1.5)}/{P_{22}}/{below right},
                {(1.5,0)}/{P_{23}}/{above left},
                {(1.5,1.5)}/{P_{24}}/{above left},
                {(0.6,-0.6)}/{P_{12}}/{below right},
                {(0.6,0)}/{P_{13}}/{above left},
                {(0.6,0.6)}/{P_{14}}/{above left}} {
                \fill[cred] \pt circle (1.6pt);
                \node[font=\scriptsize, \pos, inner sep=0.6pt,
                      fill=white, fill opacity=0.75, text opacity=1]
                    at \pt {$\lab$};
            }
        \end{tikzpicture}
        \caption{Exact geometry}
        \label{fig:iga-geom}
    \end{subfigure}
    \hspace{0.01\linewidth}
    \begin{subfigure}[b]{0.42\linewidth}
        \centering
        \begin{tikzpicture}[scale=1.45]
            \draw[black!70, line width=0.8pt] (0, -2.4) arc (-90:90:2.4)
                -- (0, 0.6) arc (90:-90:0.6) -- cycle;
            \foreach \t in {0.25, 0.5, 0.75} {
                \draw[black!45, line width=0.5pt]
                    (0, {-(0.6+\t*1.8)}) arc (-90:90:{0.6+\t*1.8});
            }
            \foreach \a in {-67.5, -45, -22.5, 0, 22.5, 45, 67.5} {
                \draw[black!45, line width=0.5pt]
                    ({0.6*cos(\a)}, {0.6*sin(\a)})
                    -- ({2.4*cos(\a)}, {2.4*sin(\a)});
            }
        \end{tikzpicture}
        \caption{Mesh in physical space}
        \label{fig:iga-mesh}
    \end{subfigure}

    \vspace{1.8ex}
    \begin{subfigure}[b]{0.72\linewidth}
        \centering
        \begin{tikzpicture}
            % square with the control-point grid (Greville points)
            \begin{axis}[
                paperaxis, at={(1.85cm, 1.65cm)}, anchor=south west,
                scale only axis, width=3.4cm, height=3.4cm,
                xmin=0, xmax=1, ymin=0, ymax=1,
                xtick={0, 1}, ytick={0, 0.5, 1},
                axis line style={black!70},
            ]
                \foreach \x in {0.25, 0.5, 0.75} {
                    \edef\temp{\noexpand\addplot[black!30,
                        line width=0.3pt]
                        coordinates {(\x, 0) (\x, 1)};}
                    \temp
                }
                \foreach \e in {0.125, 0.25, 0.375, 0.5, 0.625,
                                0.75, 0.875} {
                    \edef\temp{\noexpand\addplot[black!30,
                        line width=0.3pt]
                        coordinates {(0, \e) (1, \e)};}
                    \temp
                }
                \foreach \e in {0, 0.25, 0.5, 0.75, 1} {
                    \edef\temp{\noexpand\addplot[only marks, mark=o,
                        mark size=1.7pt, black!60]
                        coordinates {(0, \e) (0.5, \e) (1, \e)};}
                    \temp
                }
            \end{axis}
            \node[font=\footnotesize] at (3.55cm, 5.45cm) {$\xi$};
            \node[font=\footnotesize] at (5.70cm, 3.35cm) {$\eta$};
            % angular basis M_j(eta), rotated, left
            \begin{axis}[
                paperaxis, at={(0, 1.65cm)}, anchor=south west,
                scale only axis, width=1.25cm, height=3.4cm,
                xmin=0, xmax=1.05, ymin=0, ymax=1,
                x dir=reverse, xtick={0, 1}, ytick=\empty,
                axis line style={black!70},
            ]
                \addplot[cblue, thick] table[col sep=comma, x=M1, y=eta]
                    {figures/data/iga_basis_eta.csv};
                \addplot[cgreen, thick] table[col sep=comma, x=M2, y=eta]
                    {figures/data/iga_basis_eta.csv};
                \addplot[black!70, thick] table[col sep=comma, x=M3, y=eta]
                    {figures/data/iga_basis_eta.csv};
                \addplot[corange, thick] table[col sep=comma, x=M4, y=eta]
                    {figures/data/iga_basis_eta.csv};
                \addplot[cpurple, thick] table[col sep=comma, x=M5, y=eta]
                    {figures/data/iga_basis_eta.csv};
            \end{axis}
            \node[font=\footnotesize] at (0.62cm, 5.45cm) {$M_{j}(\eta)$};
            % radial basis N_i(xi), below
            \begin{axis}[
                paperaxis, at={(1.85cm, 0)}, anchor=south west,
                scale only axis, width=3.4cm, height=1.25cm,
                xmin=0, xmax=1, ymin=0, ymax=1.05,
                y dir=reverse, xtick=\empty, ytick={0, 1},
                axis line style={black!70},
            ]
                \addplot[cblue, thick]
                    table[col sep=comma, x=xi, y=N1]
                    {figures/data/iga_basis_xi.csv};
                \addplot[cgreen, thick]
                    table[col sep=comma, x=xi, y=N2]
                    {figures/data/iga_basis_xi.csv};
                \addplot[cred, thick]
                    table[col sep=comma, x=xi, y=N3]
                    {figures/data/iga_basis_xi.csv};
            \end{axis}
            \node[font=\footnotesize, anchor=west] at (5.55cm, 0.62cm)
                {$N_{i}(\xi)$};
            \node[font=\footnotesize, anchor=west] at (0, -0.55cm)
                {$\Xi_{\xi} = \{0,0,0,1,1,1\}$, \
                 $\Xi_{\eta} = \{0,0,0,\tfrac12,\tfrac12,1,1,1\}$};
        \end{tikzpicture}
        \caption{Mesh in parametric space, basis functions and knot
        vectors}
        \label{fig:iga-param}
    \end{subfigure}
    \caption{The single-patch NURBS description of the ring, shown with
    the radial direction order-elevated to quadratic, which leaves the
    geometry unchanged. (a)~Exact geometry with the isoparametric grid,
    the control polygon and the control points $\bm{P}_{ij}$; the
    weights reproduce both circles exactly. (b)~Mesh in physical space
    after knot refinement, which likewise leaves the geometry unchanged;
    the knot lines map to concentric arcs and radial lines, and the
    same mesh underlies all three panels. (c)~Parametric domain with the
    control-point grid at the Greville points, the rational quadratic
    basis $M_{j}$ along the angular coordinate $\eta$, the quadratic
    basis $N_{i}$ along the radial coordinate $\xi$, and the knot
    vectors.}
    \label{fig:iga}
\end{figure}

Univariate B-spline basis functions $N_{i,p}$ of degree $p$ on a knot
vector $\Xi = \{\xi_{1}, \dots, \xi_{m}\}$ are defined by the Cox--de
Boor recursion~\cite{piegl1997}, starting from the piecewise constants
\begin{equation}
    N_{i,0}(\xi) =
    \begin{cases}
        1 & \xi_{i} \le \xi < \xi_{i+1}, \\
        0 & \text{otherwise},
    \end{cases}
    \label{eq:coxdeboor0}
\end{equation}
and proceeding for $p \ge 1$ through
\begin{equation}
    N_{i,p}(\xi)
    =
    \frac{\xi - \xi_{i}}{\xi_{i+p} - \xi_{i}}\, N_{i,p-1}(\xi)
    + \frac{\xi_{i+p+1} - \xi}{\xi_{i+p+1} - \xi_{i+1}}\,
    N_{i+1,p-1}(\xi) .
    \label{eq:coxdeboor}
\end{equation}
The bivariate rational basis with weights $w_{ij}$ reads
\begin{equation}
    R_{ij}(\xi, \eta)
    =
    \frac{N_{i,p}(\xi)\, M_{j,q}(\eta)\, w_{ij}}
         {\sum_{k,l} N_{k,p}(\xi)\, M_{l,q}(\eta)\, w_{kl}} ,
    \label{eq:nurbs}
\end{equation}
and with control points $\bm{P}_{ij}$ the geometry map is
\begin{equation}
    \xb
    =
    \bm{G}(\xib)
    =
    \sum_{i,j} R_{ij}(\xi, \eta)\, \bm{P}_{ij} .
    \label{eq:geomap}
\end{equation}
A single patch suffices for every geometry in this paper. The
thick-walled ring of Section~\ref{subsec:ring} is one quadratic NURBS
patch whose weights reproduce the inner and outer circles exactly, the
rational form of Eq.~\eqref{eq:nurbs} being what makes conic sections
representable without approximation~\cite{piegl1997}. Fig.~\ref{fig:iga}
shows this patch together with the basis functions of its two
parametric directions; for the illustration the radial direction, which
is linear in the computations, is order-elevated to quadratic, an exact
operation that leaves the geometry unchanged. The map, its
knots, weights and control points are fixed data of the problem.
$\bm{G}$ does not depend on the trainable parameters, and the
recursion of Eq.~\eqref{eq:coxdeboor0}--\eqref{eq:coxdeboor} is evaluated
inside the differentiation graph, so the Jacobian
$\Jgeo = \partial\xb / \partial\xib$ is available exactly wherever it is
needed. Physical gradients follow from the pullback
\begin{equation}
    \grad_{\xb}(\cdot) = \Jgeo^{-\mathrm{T}}\, \grad_{\xib}(\cdot) ,
    \label{eq:pullback}
\end{equation}
and $\detJ$ enters the integration weights of
Section~\ref{subsec:quadrature}, where it also keeps the sampling
physically uniform however strongly the patch stretches. Distances that
define the seeded profile $\pf_{0}$, corner positions and all plotting
are consistently taken in physical coordinates
through~Eq.~\eqref{eq:geomap}. On the square specimens the map is the
identity, $\Jgeo = \bm{I}$, and the formulation reduces to the plain one
at no cost. Since a single patch covers each domain, no
boundary-fitted mesh is ever built on the physical domain and no
multi-patch coupling or Nitsche-type interface terms
arise~\cite{cottrell2009}. Note that spline descriptions appear on both
sides of~Eq.~\eqref{eq:geomap}, a NURBS map for the geometry and B-spline
feature grids for the solution, with the network supplying the
nonlinear composition in between.

Domains with interior holes are handled by an indicator mask rather than
by boundary fitting. The energy density is multiplied by the
characteristic function $\chi(\xb)$ of the material region, evaluated
exactly at every integration point. The hole boundary carries no
quadrature of its own, and the traction-free condition on it is the
natural boundary condition of the masked energy. The mask is
discontinuous, but this costs nothing under Monte Carlo integration,
where it is the integrand that is sampled and no derivative of $\chi$ is
ever required; under the unbiased scheme of
Section~\ref{subsec:quadrature} the masked integral is recovered exactly
in expectation. The plate-with-hole example of
Section~\ref{subsec:hole} verifies the combination of mask and
quadrature against the Kirsch solution before the same combination is
used for nucleation.

\FloatBarrier
% ---------------------------------------------------------------------
\subsection{Strains and higher-order derivatives}
\label{subsec:autodiff}

All spatial derivatives entering the energy are computed by reverse-mode
automatic differentiation through the field representation. Strains
$\straintensor(\ub)$ and the gradient $\grad\pf$ are the exact
derivatives of the discrete
fields of Eq.~\eqref{eq:lift} and~\eqref{eq:phi-ansatz}; no interpolation onto
shape functions and no finite differencing intervenes between the
representation and its derivatives. Since the energy density contains
first derivatives and is itself differentiated with respect to
$\bm{\theta}$ during training, second-order differentiation through the
computational graph is exercised at every iteration, and the $C^{1}$
regularity of the encoding is what makes this composition well-defined.
The graph retention this requires is the main memory cost of the method,
and it scales linearly with the number of integration points; quantities
evaluated outside the training loop, such as the reaction force and the
field plots, are computed in chunks so that memory never becomes a
constraint there. On mapped geometries the raw parametric gradients are
pulled back through~Eq.~\eqref{eq:pullback} before strains are formed.

The fourth-order model requires one further pass. Differentiating the
components of $\grad\pf$ once more assembles
$\lap\pf = \partial_{xx}\pf + \partial_{yy}\pf$, and this is the entire
cost of switching models. The energy density
exchanges Eq.~\eqref{eq:at2} for Eq.~\eqref{eq:fourth}, one additional
differentiation pass runs per iteration, and the measured time of a
single iteration grows by a factor of about $1.3$. This is the cost of
the iteration alone; the wall-clock times of the complete runs, which
also carry the early stopping of Section~\ref{subsec:training} and are
therefore somewhat higher, are reported in \ref{app:hyper}. The corresponding upgrade in a
Galerkin setting requires a globally $C^{1}$ trial space; here it
requires one additional call to the differentiation engine. The
fourth-order computations in this paper run on the identity geometry
map, where the parametric Laplacian coincides with the physical one. The
pullback of second derivatives through a curved map involves the second
derivatives of $\bm{G}$ and is not needed for the examples considered,
since the one curved-domain example, the ring, runs the second-order
density.

\FloatBarrier
% ---------------------------------------------------------------------
\subsection{Monte Carlo integration and adaptive sampling}
\label{subsec:quadrature}

The energy integral is estimated by importance-sampled Monte Carlo
quadrature. Let $\psitot(\xib; \bm{\theta})$ denote the total energy
density, comprising the elastic, fracture and penalty terms together
with the mask where present. For any sampling density
$\varrho > 0$ on $\widehat{\Omega}$, multiplying and dividing the integrand
by $\varrho$ rewrites the energy as an expectation over points drawn from
$\varrho$,
\begin{equation}
    \Ene(\bm{\theta})
    =
    \int_{\widehat{\Omega}} \psitot\, \detJ \,\mathrm{d}\xib
    =
    \Ex_{\xib \sim \varrho}
    \left[ \frac{\psitot\, \detJ}{\varrho} \right] .
    \label{eq:mc-exact}
\end{equation}
Drawing $M$ independent points from $\varrho$ then gives the estimator
\begin{equation}
    \widehat{\Ene}(\bm{\theta})
    =
    \frac{1}{M} \sum_{i=1}^{M}
    \frac{\psitot(\xib_{i}; \bm{\theta})\, \detJ(\xib_{i})}
         {\varrho(\xib_{i})} ,
    \qquad
    \xib_{i} \overset{\mathrm{iid}}{\sim} \varrho ,
    \label{eq:mc}
\end{equation}
which is unbiased,
$\Ex[\widehat{\Ene}] = \Ene$, whatever density is chosen. The freedom in
$\varrho$ is used to concentrate points where the solution localizes. Our
density is a mixture of three strata,
\begin{equation}
    \varrho
    =
    w_{\mathrm{u}}\, \varrho_{\mathrm{unif}}
    + w_{\mathrm{c}}\, \varrho_{\mathrm{crack}}
    + w_{\mathrm{p}}\, \varrho_{\mathrm{proc}} ,
    \label{eq:mixture}
\end{equation}
each of which is built once per load step from the frozen converged
state of the previous step.
\begin{itemize}
    \item a uniform bulk stratum, realized with a scrambled Sobol
    sequence~\cite{sobol1967} and reweighted by $\detJ$ on curved
    patches so that it is uniform in physical area;
    \item a crack stratum, a piecewise-constant density proportional to
    $\max(\pf_{0}, \pf_{n-1})\,\detJ$ on an auxiliary grid. Since it
    is built from the previous converged phase field and not from the
    pre-existing crack alone, it covers the seeded crack and everything
    that has propagated at one common physical density. The point
    concentration assigned to the pre-existing crack thus transfers
    automatically to the growing crack at a fixed total budget;
    \item a process-zone stratum proportional to
    $\pf(1-\pf) + \eta_{\mathrm{d}}\,\pf_{0} + \beta_{\mathrm{d}}\,
    \widehat{D}$, where $\widehat{D}$ is the normalized driving force
    $\dmg(\pf)\,\edens^{+}$ of the frozen state,
    evaluated in the volumetric--deviatoric form of
    Section~\ref{subsec:splits} whatever decomposition the energy
    uses. Since the estimator divides by the sampling density, the
    choice of indicator moves the variance of $\widehat{\Ene}$ and
    not its expectation; the volumetric--deviatoric form is used
    since it is smooth and, in shear-dominated states, broader than
    the spectral one, so the process zone is never left
    under-sampled. The first term peaks on
    the flanks of the band, and the driving-force term places points
    ahead of the crack tip, where the next increment will localize.
\end{itemize}
When a crack propagates substantially within a simulation, an optional
fourth component appends points on the newly created crack area at a
prescribed physical density, so that the fixed budget is not diluted by
a growing band; the number of appended points is the prescribed density
times the area in which the previous solution exceeds
$\pf = \tfrac{1}{2}$ while the seeded profile does not. This component
is used in the branching example. The mixture weights, the auxiliary
grid resolution and the budget $M$ are listed in \ref{app:hyper}.

The timing of the sampling matters as much as the densities themselves,
and two rules are followed. First, the density $\varrho$ is frozen within
each load step, so every iteration estimates the same energy and the
incremental problem retains the minimization structure
of~Eq.~\eqref{eq:incremental}. Second, the point set is redrawn from $\varrho$
every $n_{\mathrm{r}}$ optimizer iterations, with $n_{\mathrm{r}} = 1$
in all production runs, so the optimizer never interacts twice with the
same integration points. It sees the energy only through fresh unbiased
estimates, in the manner of stochastic gradient descent on the exact
functional. A configuration that lowers the estimate on one
particular point set, without lowering the energy, gains nothing that
survives the next draw. The importance weights $\detJ / \varrho$ are
quadrature data, not decision variables, and are excluded from the
differentiation graph.

The estimator variance is managed by the same construction. The
stratification places the points where the integrand is largest, which
is the classical variance-reduction role of importance sampling, and the
low-discrepancy Sobol stratum keeps the smooth bulk contribution
well-behaved. The residual noise of order $M^{-1/2}$ enters the
optimizer as gradient noise, which Adam tolerates by design, and enters
the convergence test only through a windowed average
(Section~\ref{subsec:training}). Reported quantities are evaluated on a
dedicated sample several times larger than the training budget, so the
noise visible in the load--displacement curves is well below the
physical features they resolve.

\FloatBarrier
% ---------------------------------------------------------------------
\subsection{Stability of the discretization}
\label{subsec:pairing}

Energy-minimizing network solvers for the phase-field description of
brittle fracture are known to
fail in two characteristic ways, both analyzed in~\cite{manav2024}.
First, differentiating a globally supported smooth approximant across
the damage band produces spurious oscillations in the strain and energy
fields. Second, when the representation carries more localized freedom
than the integration rule can observe, minimization finds configurations
that are sharp between the integration points and meaningless as
solutions; following~\cite{manav2024} we refer to these as zero-energy
modes. Reported remedies include taking gradients from a finite element
interpolation with its fixed Gauss rule~\cite{manav2024} and adaptively
refining the integration point set~\cite{goswami2020fourth}.

\begin{figure}[htb]
    \centering
    \begin{subfigure}[b]{0.58\linewidth}
        \centering
        \begin{tikzpicture}[scale=1.50]
            % Each level is drawn as the nodes that carry its features, not as
            % a grid of cells: no background mesh enters the method anywhere.
            % Blue marks the nodes the quadratic B-spline reads at the query
            % point. The pitch between levels must stay above the coarse
            % spacing 1.55/4, or the three lattices read as one field.
            \foreach \k/\n/\qi/\qj in {0/5/2/2, 1/9/5/4, 2/17/10/9} {
                \pgfmathsetmacro{\x}{\k*2.25}
                \foreach \i in {0,...,\the\numexpr\n-1\relax} {
                    \foreach \j in {0,...,\the\numexpr\n-1\relax} {
                        \fill[black!35] ({\x + \i*1.55/(\n-1)},
                            {3.3 + \j*1.55/(\n-1)}) circle (0.45pt);
                    }
                }
                \foreach \a in {-1,0,1} {
                    \foreach \b in {-1,0,1} {
                        \fill[cblue] ({\x + (\qi+\a)*1.55/(\n-1)},
                            {3.3 + (\qj+\b)*1.55/(\n-1)}) circle (0.45pt);
                    }
                }
                \fill[cred] ({\x + 0.96}, 4.15) circle (0.95pt);
            }
            \node[font=\footnotesize] at (0.775, 5.06) {coarse};
            \node[font=\footnotesize] at (3.025, 5.06) {moderate};
            \node[font=\footnotesize] at (5.275, 5.06) {fine};
            \draw[black!45, line width=0.4pt] (0.775, 3.3) -- (3.025, 2.55);
            \draw[black!45, line width=0.4pt] (3.025, 3.3) -- (3.025, 2.55);
            \draw[black!45, line width=0.4pt] (5.275, 3.3) -- (3.025, 2.55);
            \node[font=\small] (cat) at (3.025, 2.30)
                {$(\xib, \Fenc(\xib))$};
            \draw[-{Stealth[length=2mm]}, semithick]
                (3.025, 2.05) -- (3.025, 1.70);
            \node[draw, rounded corners=2pt, fill=cblue!8, font=\small,
                  inner sep=5pt] (nn) at (3.025, 1.35)
                {neural network};
            \draw[-{Stealth[length=2mm]}, semithick]
                (nn.south) -- ++(0, -0.35);
            \node[font=\small] at (3.025, 0.42)
                {$\hat{u},\ \hat{v},\ \hat{\pf}$};
            \node[font=\footnotesize, black!55] at (3.025, 0.02)
                {exact BCs, crack seed};
        \end{tikzpicture}
        \caption{}
        \label{fig:method-enc}
    \end{subfigure}
    \hfill
    \begin{subfigure}[b]{0.38\linewidth}
        \centering
        \includegraphics[width=0.96\linewidth]{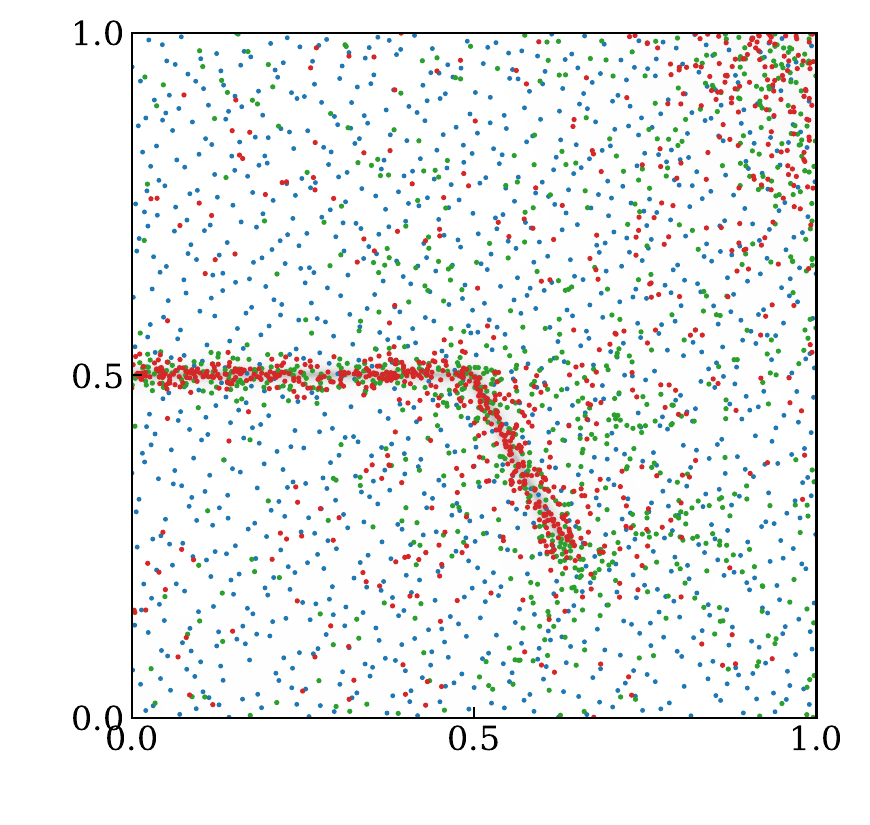}
        \caption{}
        \label{fig:method-smp}
    \end{subfigure}
    \caption{The two components of the discretization. (a)~The three
    feature grids, drawn as the nodes that carry the features. At a query
    point, in red, each level contributes the nine nodes marked in blue,
    which are interpolated with $C^{1}$ B-splines and fed with the
    coordinates to the network; the finest grid sets the resolution cap.
    (b)~One draw of the three sampling strata
    of~Eq.~\eqref{eq:mixture} for the converged shear specimen of
    Section~\ref{subsubsec:sens}, the uniform stratum in blue, the
    process-zone stratum in green and the crack stratum in red, drawn
    over the phase field in light gray; the points are redrawn at every
    iteration.}
    \label{fig:method}
\end{figure}
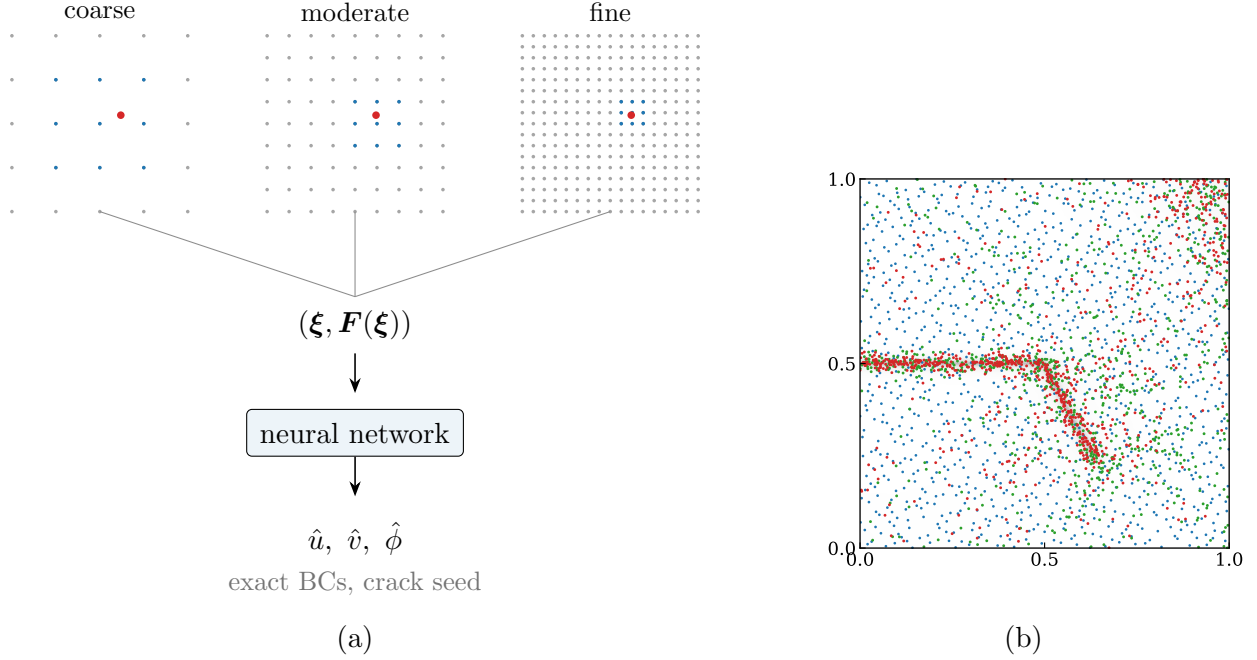

In the proposed method the two components introduced above, summarized
in Fig.~\ref{fig:method}, address the two failure modes directly. The
resolution cap of Section~\ref{subsec:encoding} removes the first mode
at the level of the function space. All localized capacity lives on $C^{1}$ splines of
spacing $h$, so there is no basis on which sub-$h$ oscillations could be
expressed, while the band itself, a few $h$ wide by the choice of the
finest level, remains fully representable. The resampling of
Section~\ref{subsec:quadrature} removes the second mode at the level of
the optimization. Over a load step the optimizer is exposed to millions
of freshly drawn points, so there is no fixed point set with respect to
which a zero-energy mode could be defined or maintained.

It has been argued that a Gauss--Legendre rule laid out on the
NURBS parameterization offers better accuracy and efficiency than
Monte Carlo sampling~\cite{goswami2020fourth}, an assessment a
recent survey repeats~\cite{ani2026}. The claim concerns the
quadrature error of a fixed integrand, and on that measure it is
correct. The estimator of Section~\ref{subsec:quadrature} is
constructed for a different setting, in which the integrand evolves
with the parameters under optimization and any point set that stands
still, deterministic or random, supplies the structure a zero-energy
mode requires. Redrawing the points at every iteration removes that
structure, and the variance this costs is controlled by the
stratification of the sampler. The reuse an estimator of this kind
tolerates is measured in \ref{app:failure}, where holding one point
set for ten iterations leaves the response unchanged and holding it
for a hundred already removes the failure event.

Neither ingredient is sufficient on its own. The cap does not protect a
fixed quadrature rule, since configurations that lower the estimated
energy at the integration points while deteriorating between them do
not require sub-$h$ scales at all; they can be built from the resolved
scales alone, and a fixed rule never observes the space between its
points. The resampling, conversely, does not remove the derivative
oscillations, since a representation able to form arbitrarily fine
features develops them at the band under differentiation whether or not
the points move. With both in place, the representation cannot form
features finer than the integration observes, and the integration
offers no fixed point set that could be exploited. The resulting
sharpness is obtained within the quadratic models of
Section~\ref{sec:formulation}, with no threshold-type damage model
involved. The studies of \ref{app:failure} examine the two directions
separately. Fixing the point set reopens the failure mode, while
enlarging the representation at a fixed budget leaves the response
essentially unchanged, the peak moving by under $5\%$, so the
pairing binds in the first direction and is forgiving in the second.

\FloatBarrier
% ---------------------------------------------------------------------
\subsection{Training strategy}
\label{subsec:training}

Load stepping proceeds by warm starting, the natural continuation
strategy for a quasi-static problem. The parameters converged at step
$n-1$ initialize step $n$, and a frozen copy of the step-$(n-1)$ network
supplies everything the new step needs from the past, namely the field
$\pf_{n-1}$ entering the penalty of Eq.~\eqref{eq:penalty}, the sampler
densities of Section~\ref{subsec:quadrature}, and the driving-force
field of the process-zone stratum. No state is stored at integration
points at any time; the previous solution is available as a field,
evaluable at whatever points the sampler draws next.

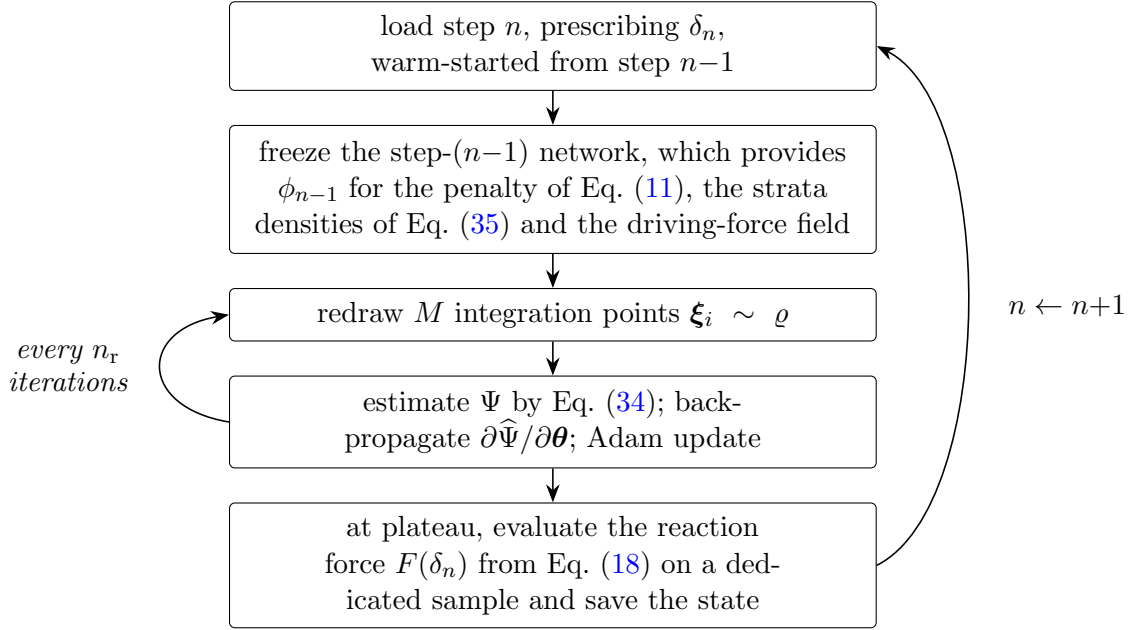
\begin{figure}[htb]
    \centering
    \begin{tikzpicture}[
        node distance = 4.5mm,
        box/.style  = {rectangle, rounded corners = 2pt, draw,
                       align = center, inner sep = 5pt, font = \small,
                       text width = 82mm},
        lbl/.style  = {font = \small\itshape},
        arr/.style  = {-{Stealth[length=2.4mm]}, semithick}
    ]
        \node[box] (step)   {load step $n$, prescribing $\delta_{n}$,
                             warm-started from step $n{-}1$};
        \node[box, below=of step] (freeze)
                            {freeze the step-$(n{-}1)$ network, which
                             provides $\pf_{n-1}$ for the
                             penalty of Eq.~\eqref{eq:penalty}, the strata
                             densities of Eq.~\eqref{eq:mixture} and the
                             driving-force field};
        \node[box, below=of freeze] (draw)
                            {redraw $M$ integration points
                             $\xib_{i} \sim \varrho$};
        \node[box, below=of draw] (est)
                            {estimate $\Ene$ by~Eq.~\eqref{eq:mc};
                             backpropagate
                             $\partial\widehat{\Ene}/\partial\bm{\theta}$;
                             Adam update};
        \node[box, below=of est] (force)
                            {at plateau, evaluate the reaction force
                             $F(\delta_{n})$ from~Eq.~\eqref{eq:force} on a
                             dedicated sample and save the state};
        \draw[arr] (step)   -- (freeze);
        \draw[arr] (freeze) -- (draw);
        \draw[arr] (draw)   -- (est);
        \draw[arr] (est)    -- (force);
        \draw[arr] (est.west) .. controls +(-12mm, 2mm) and +(-12mm, -2mm)
            .. (draw.west);
        \node[lbl, anchor=east] at ($(draw.west)!0.5!(est.west) + (-12mm, 0)$)
            {\shortstack{every $n_{\mathrm{r}}$\\iterations}};
        \draw[arr] (force.east) .. controls +(16mm, 8mm) and +(16mm, -8mm)
            .. (step.east);
        \node[lbl, anchor=west] at ($(step.east)!0.5!(force.east) + (16mm, 0)$)
            {$n \leftarrow n{+}1$};
    \end{tikzpicture}
    \caption{One load step of the solver. The sampling density is frozen
    per load step; the integration points are redrawn every
    $n_{\mathrm{r}}$ optimizer iterations, with $n_{\mathrm{r}} = 1$ in
    all production runs.}
    \label{fig:flowchart}
\end{figure}

Each load step minimizes the estimated energy with
Adam~\cite{kingma2015} in two parameter groups, the perceptron weights
and the encoding grids, the latter at the larger learning rate and
with the small penalty on the feature amplitudes of
Section~\ref{subsec:encoding}. The cold start at step 0, which begins
from $\pf \approx \pf_{0}$, receives roughly twice the iteration budget
of the warm steps. Iterations stop early when the relative range of the
estimated energy over a trailing window falls below a plateau tolerance,
with the window length chosen large enough that estimator noise averages
out of the test. At convergence of each step the reaction force is
evaluated from~Eq.~\eqref{eq:force} by automatic differentiation of the
elastic energy with respect to the prescribed displacement, on a
dedicated sample several times larger than the training budget. All
schedules, learning rates, tolerances and budgets are tabulated in
\ref{app:hyper}. Algorithm~\ref{alg:solver} states the complete loop,
and Fig.~\ref{fig:flowchart} shows the same structure graphically.

\begin{algorithm}[htb]
    \caption{Load-stepping deep energy solver.}
    \label{alg:solver}
    \begin{algorithmic}[1]
        \State initialize $\bm{\theta}$; seed the phase field with
               $\pf_{0}$ (Section~\ref{subsec:fields})
        \For{load step $n = 1, \dots, N$}
            \State prescribe $\delta_{n}$; freeze the previous network
                   ($\pf_{n-1}$, strata densities of Eq.~\eqref{eq:mixture},
                   driving force)
            \For{iteration $t = 1, 2, \dots$ until plateau or budget}
                \If{$t \equiv 0 \pmod{n_{\mathrm{r}}}$}
                    \State redraw $\{\xib_{i}\}_{i=1}^{M} \sim \varrho$
                \EndIf
                \State $\widehat{\Ene} \gets \dfrac{1}{M} \sum_{i}
                       \psitot(\xib_{i}; \bm{\theta})\, \detJ /
                       \varrho(\xib_{i})$
                       \Comment{unbiased estimate of Eq.~\eqref{eq:mc}}
                \State $\bm{\theta} \gets \mathrm{Adam}\bigl(\bm{\theta},
                       \partial\widehat{\Ene}/\partial\bm{\theta}\bigr)$
            \EndFor
            \State $F(\delta_{n}) \gets \partial\Ene_{\mathrm{el}} /
                   \partial\delta$ on a dedicated sample
                   \Comment{reaction force of Eq.~\eqref{eq:force}}
            \State save the converged state
        \EndFor
    \end{algorithmic}
\end{algorithm}

% =====================================================================
\section{Numerical examples}
\label{sec:results}

The capability of the method is assessed on six examples of increasing
difficulty. The single-edge-notched (SEN) tension and shear tests
provide the quantitative comparison with reference solutions, for the
second- and fourth-order models alike, and crack branching and the
coalescence of three pre-existing cracks exercise evolving crack
topology. A plate with a circular hole verifies the elastic machinery
against a closed-form solution and then demonstrates nucleation
without any pre-existing crack, a thick-walled ring exercises the
geometry mapping on a curved domain, and ten random multi-crack
configurations from a recent public benchmark dataset test the
method zero-shot, twenty runs evaluated against the reference
solutions published with it.
Alternatives to the method of
Section~\ref{sec:method} are examined one ingredient at a time in
\ref{app:failure}, together with the failure modes they reproduce.

The finite element references for the notched square specimens and
the coalescence test were computed with an in-house staggered solver
written for this work and distributed with the code of the paper. It
implements the standard second-order model on a uniform mesh of
$512 \times 512$ bilinear quadrilateral elements over the unit square,
an element size of $\lreg/5$, finer than the $\lreg/2$
of common practice~\cite{miehe2010cmame}, with the displacement and
the phase field together carrying $789\,507$ degrees of freedom.
Irreversibility is enforced through the
history field of Miehe et al.~\cite{miehe2010cmame}, and the
pre-existing crack is imposed as a $\pf = 1$ condition on the notch
nodes. Each load increment is solved by alternating the displacement
and the phase-field problem, and geometry, material data, boundary
conditions, regularization length and the strain energy decomposition
are matched to the corresponding deep energy run. The implementation was verified against the
published curves of Tangella et al.~\cite{tangella2022} for the
single-edge-notched tests before being used as the reference here.

The elastic stage of the plate with a hole is compared
against the Kirsch solution~\cite{kirsch1898}. For the ring we compare
the computed crack path against the isogeometric phase-field results of
Si et al.~\cite{si2023}, and the reference fields and curves of the
random multi-crack configurations are those distributed with the
benchmark dataset~\cite{hamdi2026}. Unless stated otherwise, the square specimens
share one material set, $E = 340\,\mathrm{GPa}$, $\nu = 0.22$ and
$\Gc = 42.47\,\mathrm{J/m^{2}}$ under plane strain, a unit side length
of $1\,\mathrm{mm}$, a unit out-of-plane thickness of
$1\,\mathrm{mm}$, so that the reactions reported below are forces in
newtons, and a regularization length
$\lreg = 0.01\,\mathrm{mm}$; the discretization and training parameters
of every run are tabulated in \ref{app:hyper}. All load--displacement
curves are evaluated from the energy derivative of Eq.~\eqref{eq:force}, and
all runs of one example share the seeding, sampling and training
strategy of Section~\ref{sec:method}; the discretization parameters
that differ between examples are the level resolutions, the point
budget and the iteration budget of \ref{app:hyper}, and no quantity
is tuned to a particular run.

\FloatBarrier
% ---------------------------------------------------------------------
\subsection{Single-edge-notched tension and shear}
\label{subsec:sen}

The single-edge-notched square is the standard quantitative benchmark of
the phase-field literature~\cite{miehe2010cmame,ambati2015}. A
horizontal pre-existing crack runs from the left edge to the center at
mid-height, seeded through the optimal
profiles of Eq.~\eqref{eq:profile2} and~\eqref{eq:profile4} as described in
Section~\ref{subsec:setup}. The bottom edge is fixed and the top edge
is displacement-driven through the lift of Eq.~\eqref{eq:lift}, vertically for
the tension (mode I) test and horizontally for the shear (mode II)
test; both set-ups are shown in Fig.~\ref{fig:setup-sen}.

\begin{figure}[!htb]
    \centering
    \begin{subfigure}[b]{0.44\linewidth}
        \centering
        \begin{tikzpicture}[scale=1.45]
            \fill[clamphatch] (0, -0.30) rectangle (3.2, 0);
            \draw[black!70, line width=0.7pt] (0, 0) -- (3.2, 0);
            \draw[spec] (0, 0) rectangle (3.2, 3.2);
            \draw[crackline] (0, 1.6) -- (1.6, 1.6);
            \foreach \x in {0.25, 0.7, 1.15, 1.6, 2.05, 2.5, 2.95} {
                \draw[loadarrow] (\x, 3.28) -- (\x, 3.75);
            }
            \node[cred, font=\normalsize, anchor=south] at (1.6, 3.82)
                {$\bm{\delta}$};
            \draw[dimline] (0, -0.58) -- (3.2, -0.58);
            \node[dimfont, below] at (1.6, -0.60) {$1\,\mathrm{mm}$};
            \draw[dimline] (3.55, 0) -- (3.55, 3.2);
            \node[dimfont, rotate=90, below] at (3.60, 1.6) {$1\,\mathrm{mm}$};
            \draw[dimline] (0, 1.90) -- (1.6, 1.90);
            \node[dimfont, above] at (0.8, 1.92) {$0.5\,\mathrm{mm}$};
        \end{tikzpicture}
        \caption{}
        \label{fig:setup-sent}
    \end{subfigure}
    \hspace{0.06\linewidth}
    \begin{subfigure}[b]{0.44\linewidth}
        \centering
        \begin{tikzpicture}[scale=1.45]
            \fill[clamphatch] (0, -0.30) rectangle (3.2, 0);
            \draw[black!70, line width=0.7pt] (0, 0) -- (3.2, 0);
            \draw[spec] (0, 0) rectangle (3.2, 3.2);
            \draw[crackline] (0, 1.6) -- (1.6, 1.6);
            \draw[black!70, line width=0.7pt] (0, 3.2) -- (3.2, 3.2);
            \foreach \x in {0.10, 0.54, 0.98, 1.42, 1.86, 2.30, 2.74} {
                \draw[loadarrow] (\x, 3.42) -- (\x + 0.32, 3.42);
            }
            \node[cred, font=\normalsize, anchor=south] at (1.6, 3.62)
                {$\bm{\delta}$};
            \draw[dimline] (0, -0.55) -- (3.2, -0.55);
            \node[dimfont, below] at (1.6, -0.57) {$1\,\mathrm{mm}$};
            \draw[dimline] (3.55, 0) -- (3.55, 3.2);
            \node[dimfont, rotate=90, below] at (3.60, 1.6) {$1\,\mathrm{mm}$};
            \draw[dimline] (0, 1.90) -- (1.6, 1.90);
            \node[dimfont, above] at (0.8, 1.92) {$0.5\,\mathrm{mm}$};
        \end{tikzpicture}
        \caption{}
        \label{fig:setup-sens}
    \end{subfigure}
    \caption{Single-edge-notched specimen under (a)~tension and
    (b)~shear. The bottom edge is fixed, the top edge is
    displacement-driven, and the pre-existing crack is seeded
    diffusely through $\pf_{0}$.}
    \label{fig:setup-sen}
\end{figure}
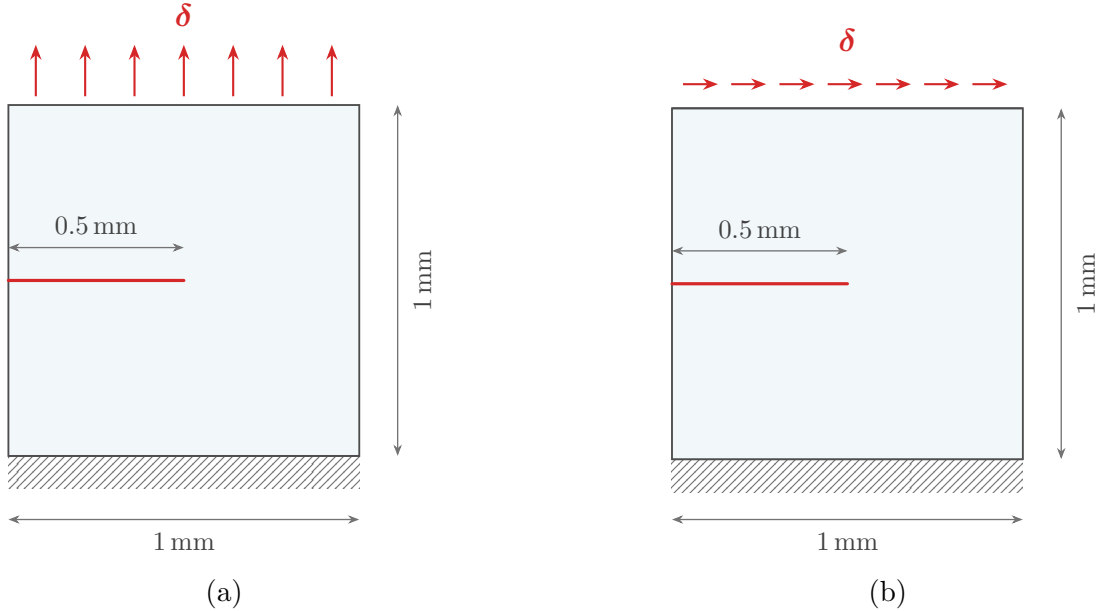 The tension test applies
$80$ increments of $10^{-5}\,\mathrm{mm}$; the shear test applies $126$
increments of $2.5 \times 10^{-5}\,\mathrm{mm}$. Both the second- and
the fourth-order model are run on identical discretizations, differing
only in the density $\psi_{\mathrm{c},n}$ entering~Eq.~\eqref{eq:pi}, so
each test produces three curves to compare, the finite element
reference and the two deep energy solutions.

% ---------------------------------------------------------------------
\subsubsection{Tension}
\label{subsubsec:sent}

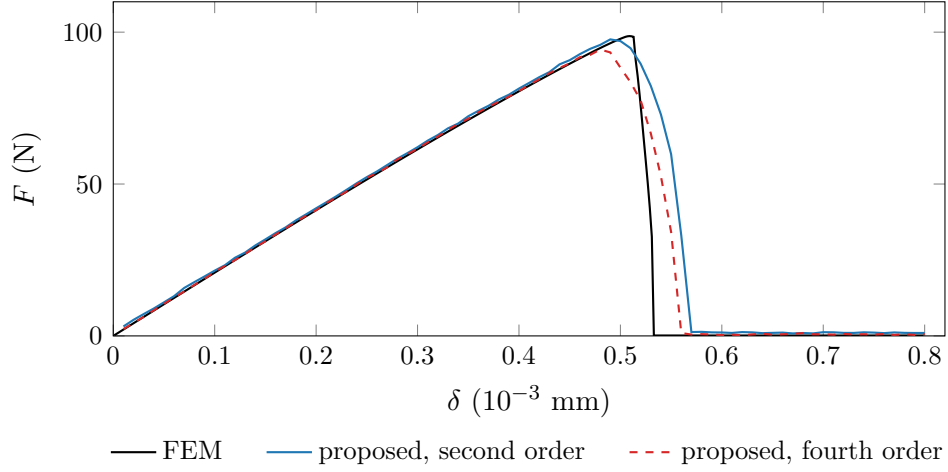
\begin{figure}[!htb]
    \centering
    \begin{tikzpicture}
        \begin{axis}[
            paperaxis, width=0.74\linewidth, height=6.0cm,
            xlabel={$\delta$ ($10^{-3}$ mm)}, ylabel={$F$ (N)},
            xmin=0, xmax=0.82, ymin=0, ymax=110,
            legend style={at={(0.5, -0.28)}, anchor=north,
                          legend columns=3,
                          /tikz/every even column/.append style=
                              {column sep=0.5cm}},
        ]
            \addplot[black, thick, each nth point=2]
                table[col sep=comma,
                      x expr=\thisrow{delta_mm}*1000, y=F]
                {figures/data/sent_fem.csv};
            \addlegendentry{FEM}
            \addplot[cblue, thick]
                table[col sep=comma,
                      x expr=\thisrow{delta_mm}*1000, y=F]
                {figures/data/sent_dem2.csv};
            \addlegendentry{proposed, second order}
            \addplot[cred, thick, dashed]
                table[col sep=comma,
                      x expr=\thisrow{delta_mm}*1000, y=F]
                {figures/data/sent_dem4.csv};
            \addlegendentry{proposed, fourth order}
        \end{axis}
    \end{tikzpicture}
    \caption{Single-edge-notched tension. Load--displacement curves of
    the second- and fourth-order solutions against the finite element
    reference, computed with the in-house staggered solver described
    at the head of Section~\ref{sec:results}.}
    \label{fig:sent-fd}
\end{figure}

Under tension the crack propagates horizontally across the ligament and
the load--displacement curves show a single peak followed by an abrupt
drop to complete failure (Fig.~\ref{fig:sent-fd}). The second-order
solution peaks at $97.6\,\mathrm{N}$ at
$\delta = 4.90 \times 10^{-4}\,\mathrm{mm}$, against the
reference peak of $98.7\,\mathrm{N}$ at
$5.09 \times 10^{-4}\,\mathrm{mm}$, so that the peak load is
reproduced to within $1.1\%$ and the displacement at which it occurs
to within $3.7\%$. The fourth-order solution fails
slightly earlier, at $94.3\,\mathrm{N}$, about $3\%$ below its
second-order counterpart, consistent with its more compact crack band.

\begin{figure}[!htb]
    \centering
    \begin{subfigure}[t]{0.33\linewidth}
        \centering
        \includegraphics[width=\linewidth]{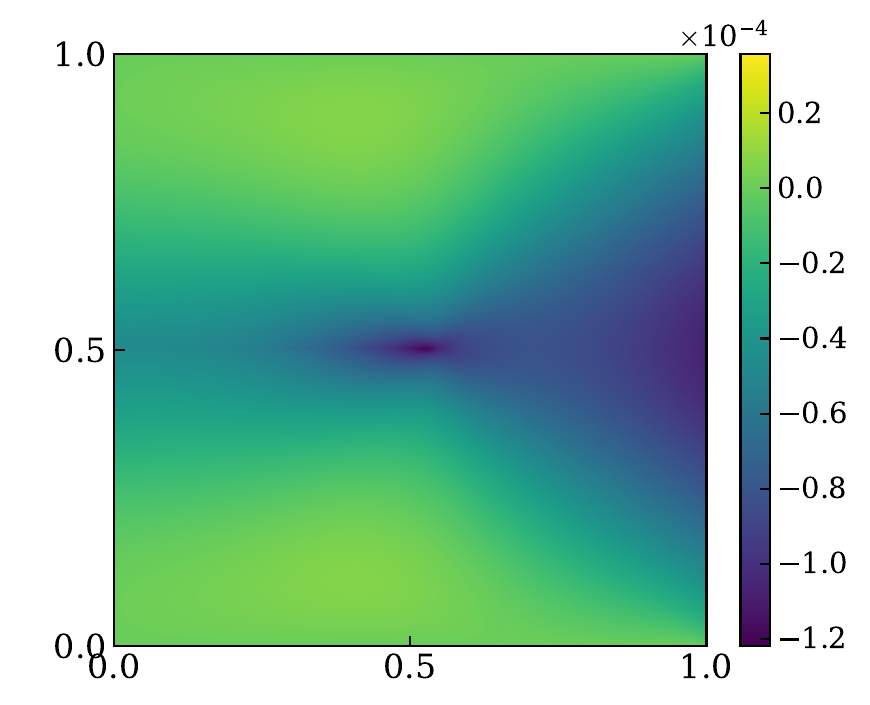}
        \caption{}
    \end{subfigure}%
    \hfill%
    \begin{subfigure}[t]{0.33\linewidth}
        \centering
        \includegraphics[width=\linewidth]{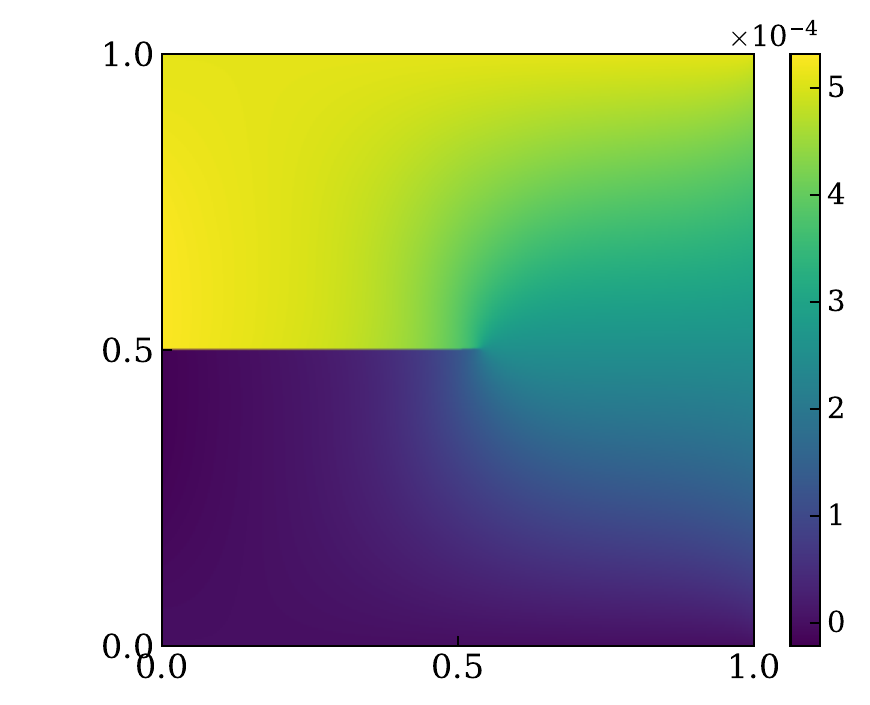}
        \caption{}
    \end{subfigure}%
    \hfill%
    \begin{subfigure}[t]{0.33\linewidth}
        \centering
        \includegraphics[width=\linewidth]{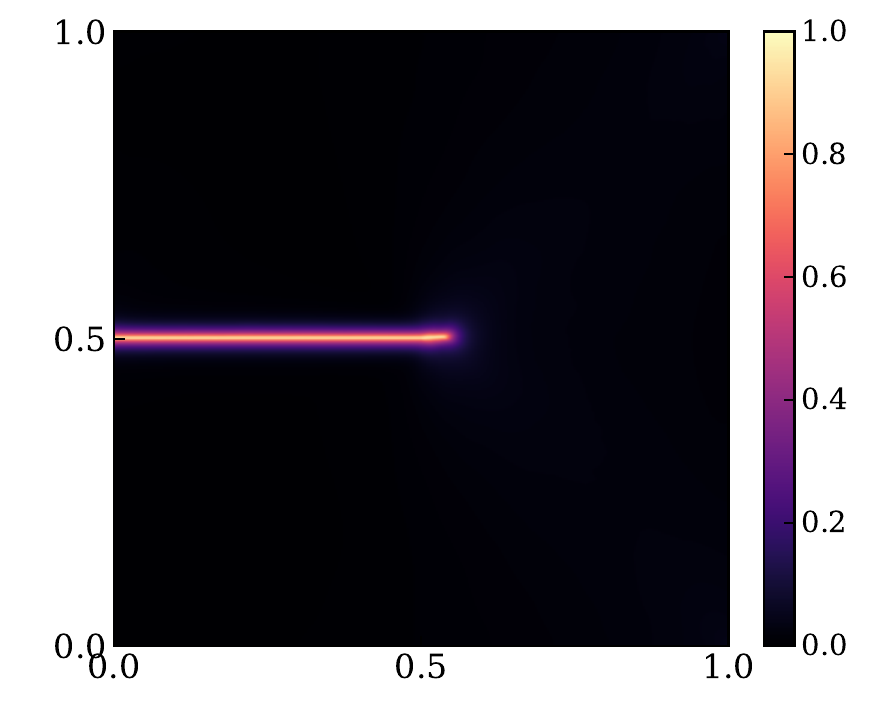}
        \caption{}
    \end{subfigure}

    \vspace{0.6ex}
    \begin{subfigure}[t]{0.33\linewidth}
        \centering
        \includegraphics[width=\linewidth]{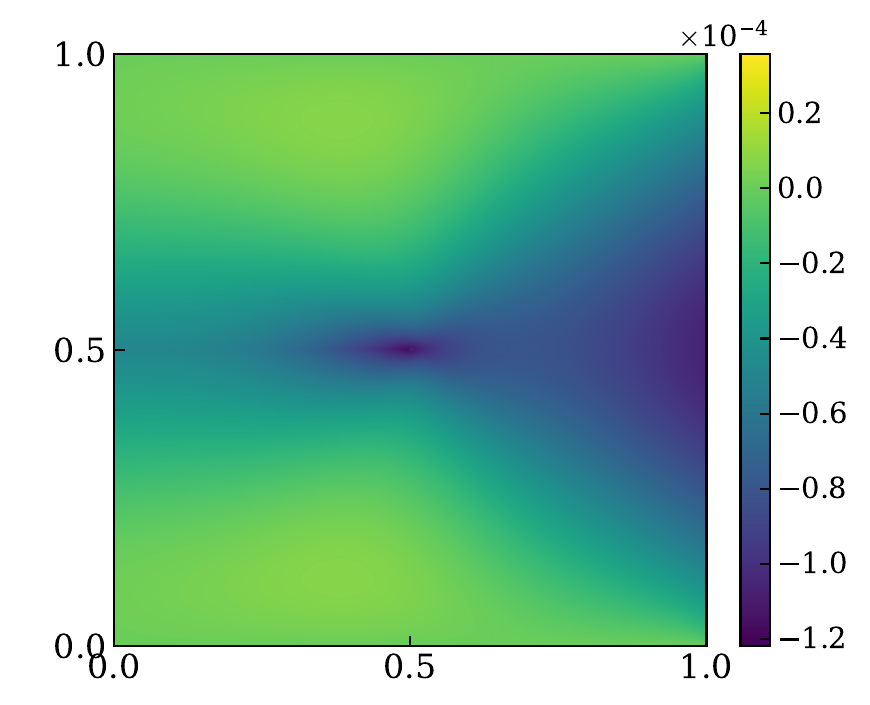}
        \caption{}
    \end{subfigure}%
    \hfill%
    \begin{subfigure}[t]{0.33\linewidth}
        \centering
        \includegraphics[width=\linewidth]{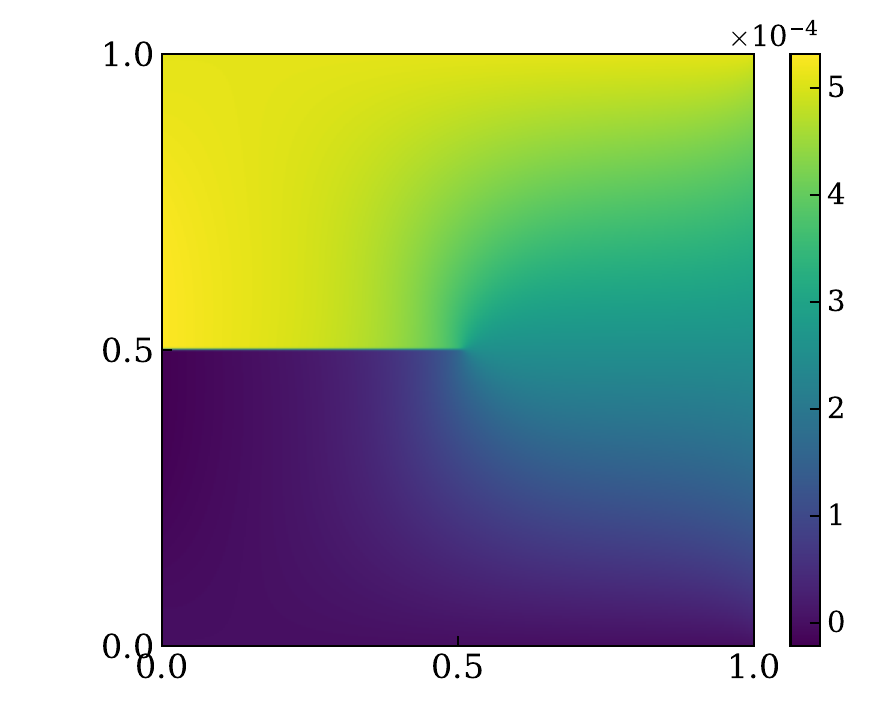}
        \caption{}
    \end{subfigure}%
    \hfill%
    \begin{subfigure}[t]{0.33\linewidth}
        \centering
        \includegraphics[width=\linewidth]{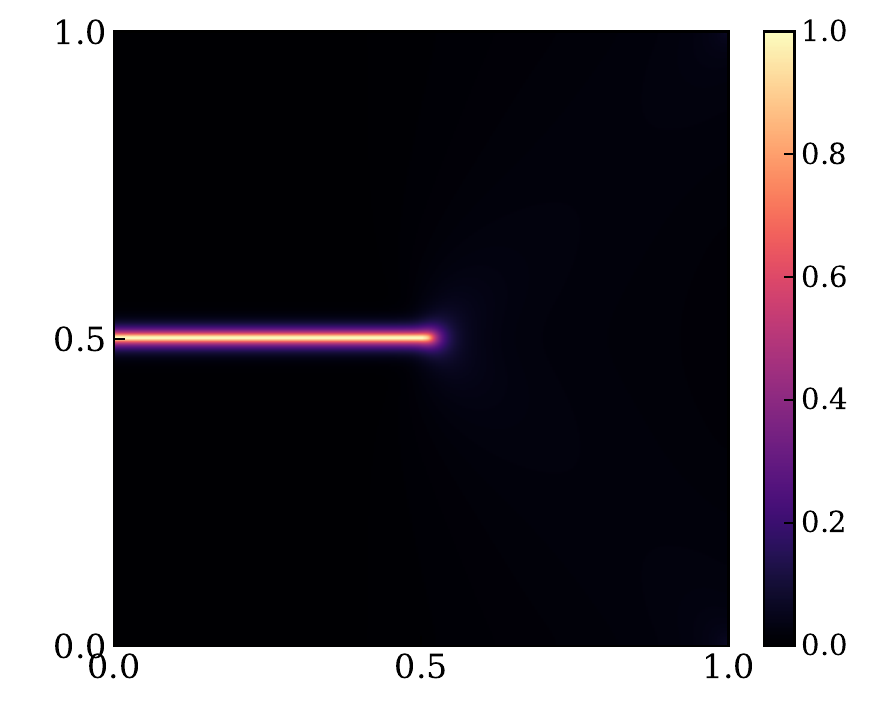}
        \caption{}
    \end{subfigure}
    \caption{Single-edge-notched tension during propagation, at
    $\delta = 5.1 \times 10^{-4}\,\mathrm{mm}$. (a)--(c)~Displacement
    components $u$, $v$ and phase field of the second-order deep
    energy solution; (d)--(f)~the finite element fields at the same
    load level. Displacements in mm; each displacement component
    shares one color scale between the two rows.}
    \label{fig:sent-prop}
\end{figure}

The fields while the crack is running are compared with the finite
element fields in Fig.~\ref{fig:sent-prop}, at
$\delta = 5.1 \times 10^{-4}\,\mathrm{mm}$, with each displacement
component drawn on one color scale shared between the two solutions.
Both components jump across the grown part of the band and stay
smooth ahead of the tip.

\begin{figure}[!htb]
    \centering
    \begin{subfigure}[t]{0.33\linewidth}
        \centering
        \includegraphics[width=\linewidth]{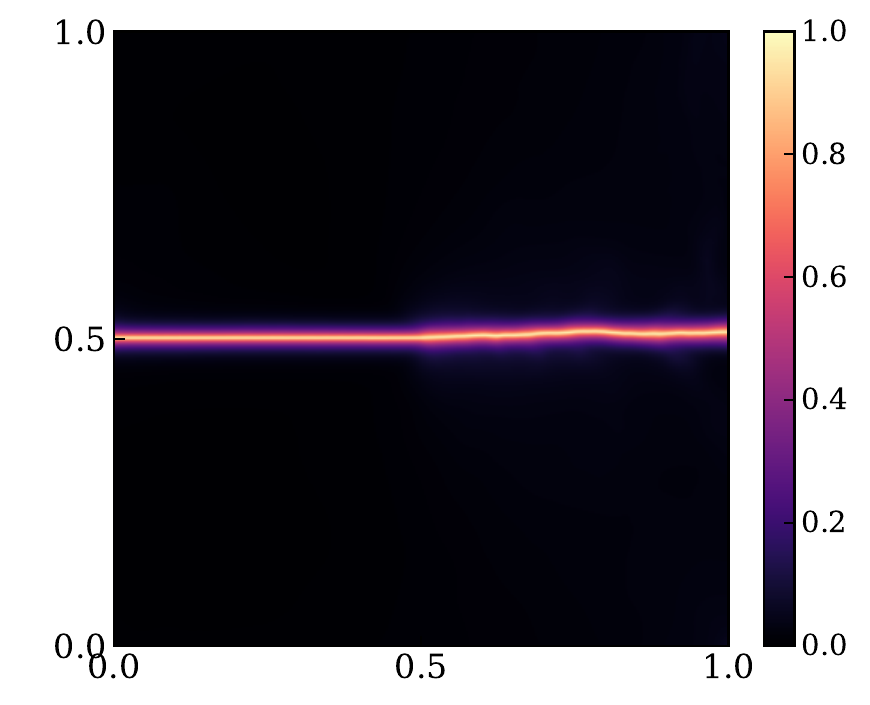}
        \caption{}
    \end{subfigure}%
    \hfill%
    \begin{subfigure}[t]{0.33\linewidth}
        \centering
        \includegraphics[width=\linewidth]{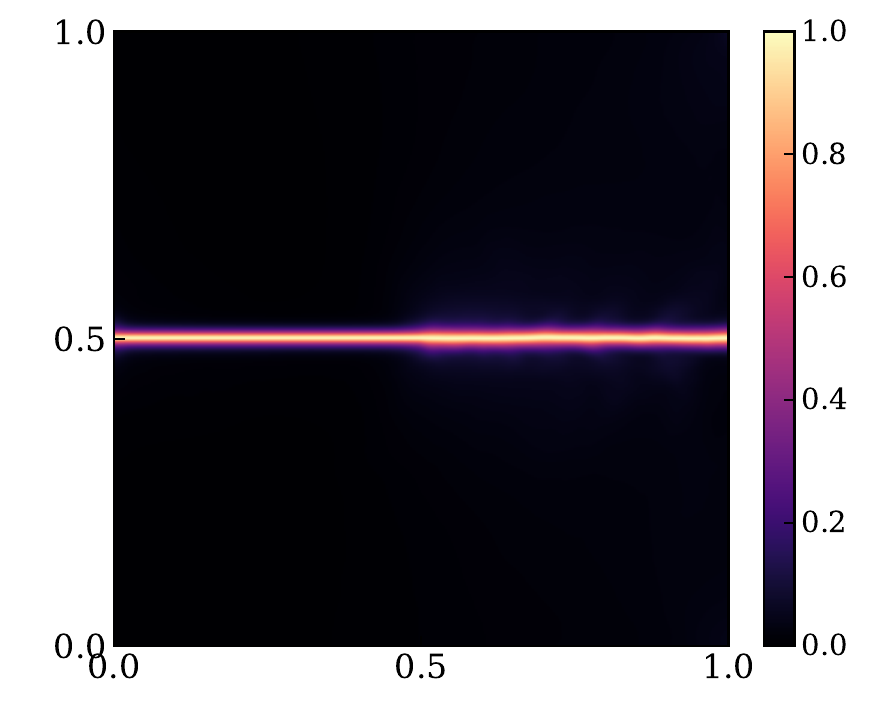}
        \caption{}
    \end{subfigure}%
    \hfill%
    \begin{subfigure}[t]{0.33\linewidth}
        \centering
        \includegraphics[width=\linewidth]{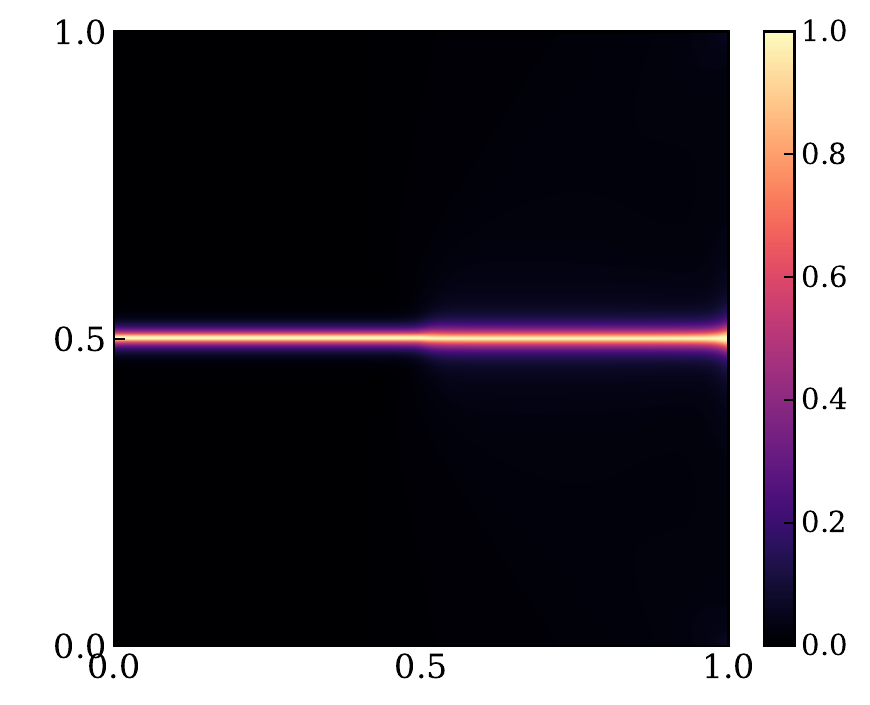}
        \caption{}
    \end{subfigure}
    \caption{Single-edge-notched tension, final phase fields.
    (a)~Second-order and (b)~fourth-order deep energy solutions,
    (c)~finite element reference; all renderings share one color
    scale.}
    \label{fig:sent-fields}
\end{figure}

\begin{figure}[!htb]
    \centering
    \begin{subfigure}[t]{0.33\linewidth}
        \centering
        \includegraphics[width=\linewidth]{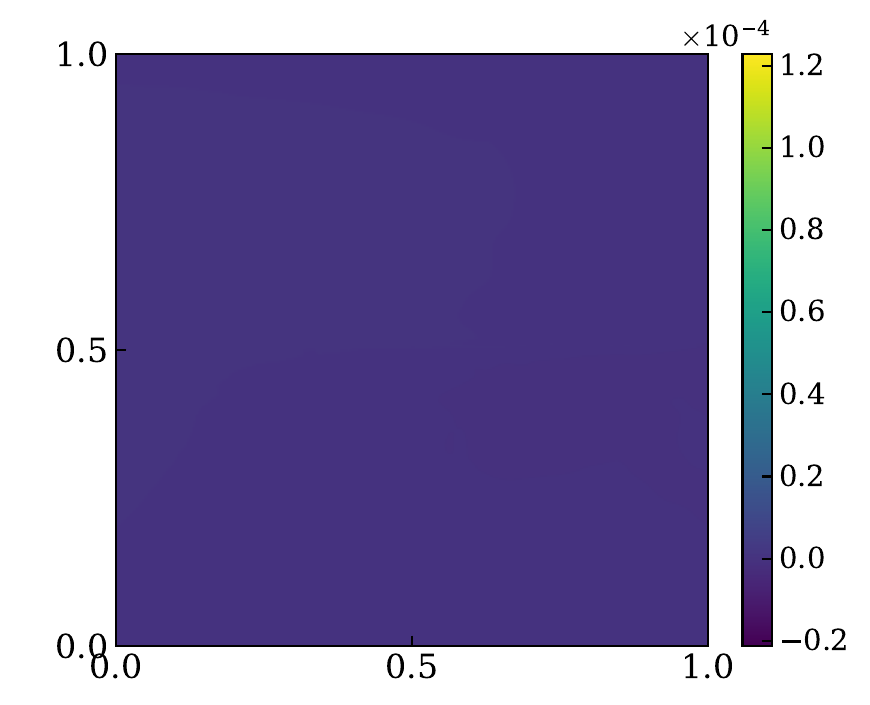}
        \caption{}
    \end{subfigure}%
    \hspace{0.04\linewidth}%
    \begin{subfigure}[t]{0.33\linewidth}
        \centering
        \includegraphics[width=\linewidth]{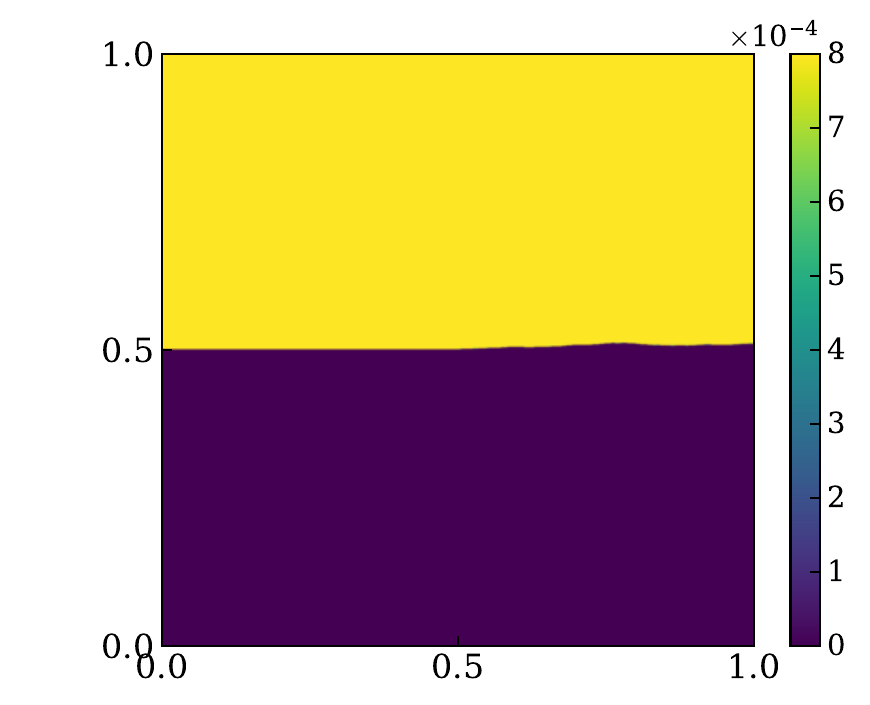}
        \caption{}
    \end{subfigure}

    \vspace{0.6ex}
    \begin{subfigure}[t]{0.33\linewidth}
        \centering
        \includegraphics[width=\linewidth]{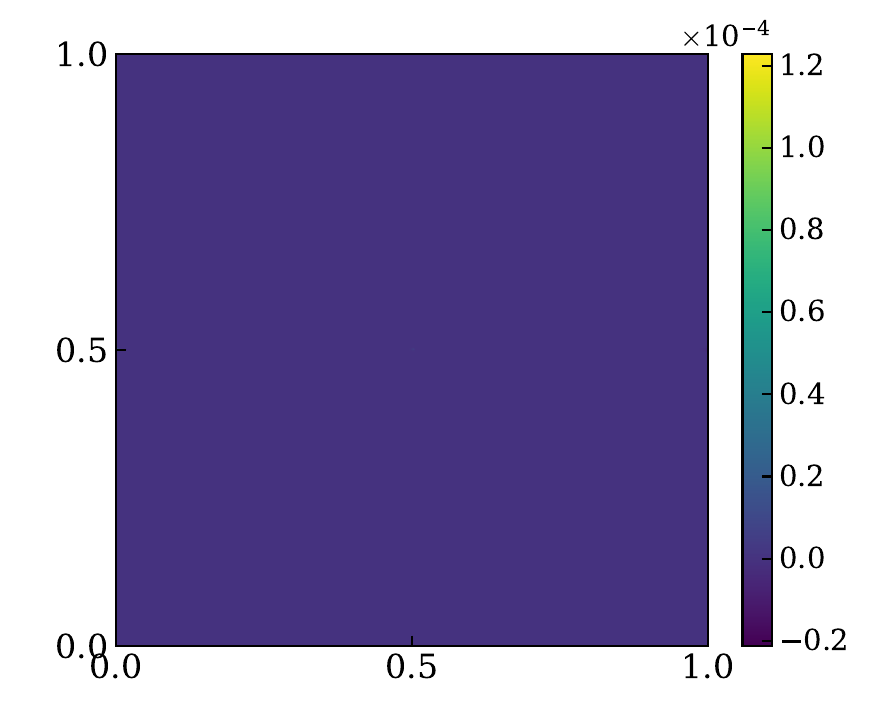}
        \caption{}
    \end{subfigure}%
    \hspace{0.04\linewidth}%
    \begin{subfigure}[t]{0.33\linewidth}
        \centering
        \includegraphics[width=\linewidth]{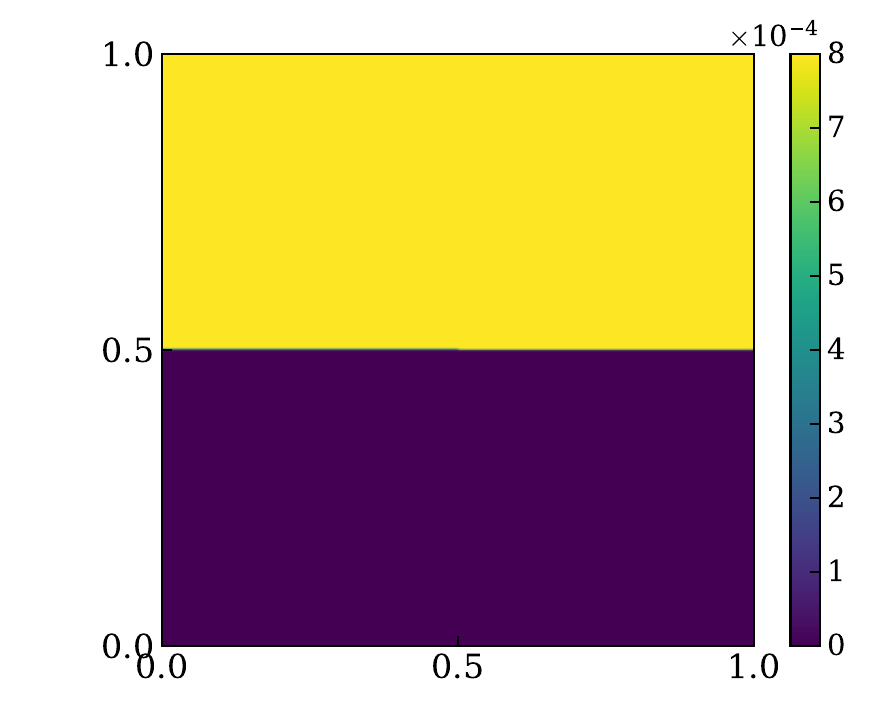}
        \caption{}
    \end{subfigure}
    \caption{Single-edge-notched tension, displacement components of
    the failed state in mm. (a),(b)~$u$ and $v$ of the second-order
    deep energy solution; (c),(d)~the finite element fields; each
    component shares one color scale between the two rows.}
    \label{fig:sent-u}
\end{figure}

The final phase fields are compared in Fig.~\ref{fig:sent-fields}.
The computed path coincides with the finite element path along the
full ligament, and the fourth-order band is visibly more compact than
the second-order one at the same regularization length, consistent
with the narrower optimal profile of Eq.~\eqref{eq:profile4}. The
displacement components of the failed state follow in
Fig.~\ref{fig:sent-u}; the two parts of the specimen separate across
the fully developed band and agree with the finite element fields on
the shared color scales.

\FloatBarrier
% ---------------------------------------------------------------------
\subsubsection{Shear}
\label{subsubsec:sens}

\begin{figure}[!htb]
    \centering
    \begin{tikzpicture}
        \begin{axis}[
            paperaxis, width=0.74\linewidth, height=6.0cm,
            xlabel={$\delta$ ($10^{-3}$ mm)}, ylabel={$F$ (N)},
            xmin=0, xmax=3.25, ymin=0, ymax=75,
            legend style={at={(0.5, -0.28)}, anchor=north,
                          legend columns=3,
                          /tikz/every even column/.append style=
                              {column sep=0.5cm}},
        ]
            \addplot[black, thick, each nth point=2]
                table[col sep=comma,
                      x expr=\thisrow{delta_mm}*1000, y=F]
                {figures/data/sens_fem.csv};
            \addlegendentry{FEM}
            \addplot[cblue, thick]
                table[col sep=comma,
                      x expr=\thisrow{delta_mm}*1000, y=F]
                {figures/data/sens_dem2.csv};
            \addlegendentry{proposed, second order}
            \addplot[cred, thick, dashed]
                table[col sep=comma,
                      x expr=\thisrow{delta_mm}*1000, y=F]
                {figures/data/sens_dem4.csv};
            \addlegendentry{proposed, fourth order}
        \end{axis}
    \end{tikzpicture}
    \caption{Single-edge-notched shear. Load--displacement curves of
    the second- and fourth-order solutions against the finite element
    reference.}
    \label{fig:sens-fd}
\end{figure}
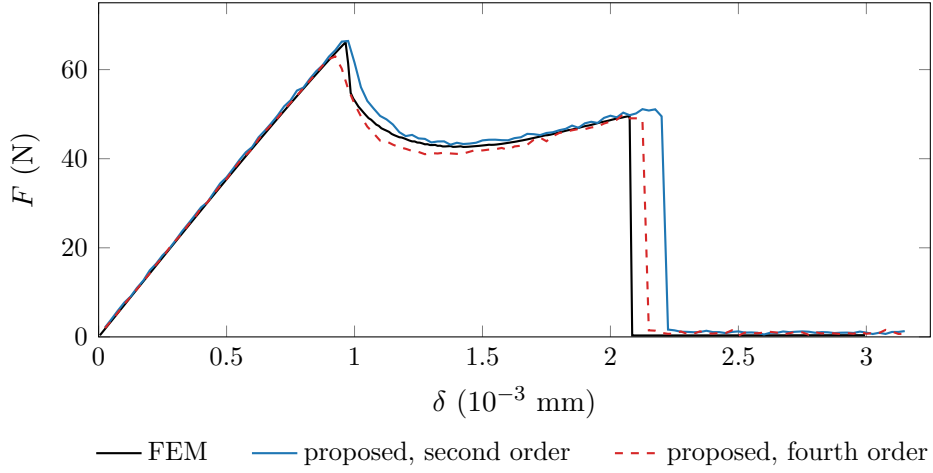

The shear test provides the more demanding comparison. The crack curves
downward toward the bottom edge, and the reference curve exhibits a
first peak, a softening dip, and a second rise as the inclined crack
interacts with the fixed edge before final severance. The computed
curves reproduce all of these features (Fig.~\ref{fig:sens-fd}). The
second-order peak of $66.42\,\mathrm{N}$ at
$9.75 \times 10^{-4}\,\mathrm{mm}$ agrees with the reference value of
$66.35\,\mathrm{N}$ at $9.70 \times 10^{-4}\,\mathrm{mm}$ to within
$0.1\%$, the softening valley ($43.1$ against $42.6\,\mathrm{N}$) and
the second rise ($51.1\,\mathrm{N}$ at $2.13 \times 10^{-3}\,\mathrm{mm}$
against $49.5\,\mathrm{N}$ at $2.07 \times 10^{-3}\,\mathrm{mm}$) follow
the reference to within $3.2\%$, and the phase field remains a
single band throughout, with no spurious secondary branch. The
fourth-order model traces the same curve with a peak of
$63.0\,\mathrm{N}$, about $5\%$ below the second-order one, and an
earlier final severance.

\begin{figure}[!htb]
    \centering
    \begin{subfigure}[t]{0.33\linewidth}
        \centering
        \includegraphics[width=\linewidth]{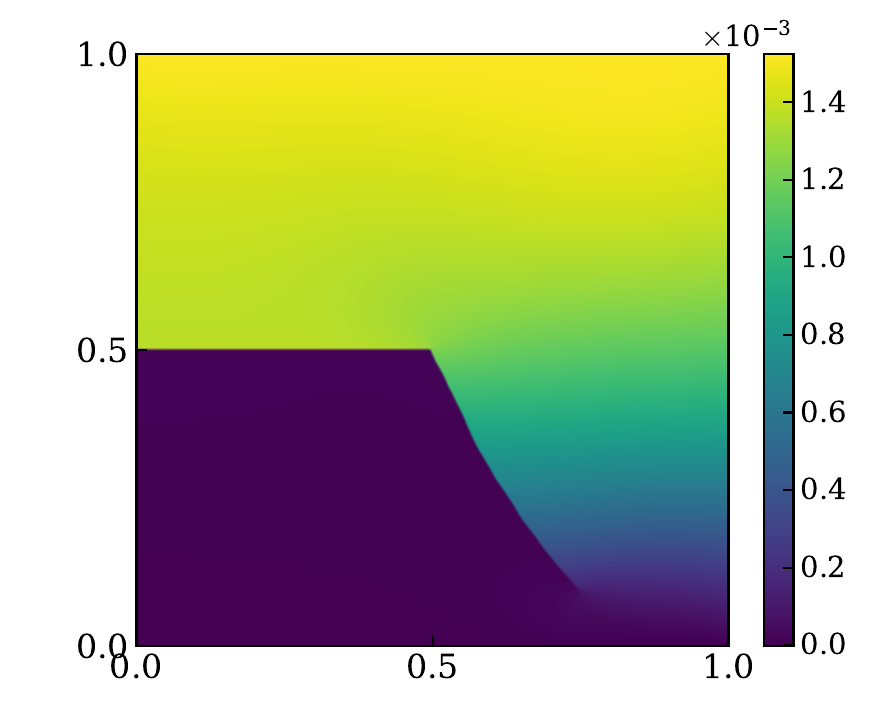}
        \caption{}
    \end{subfigure}%
    \hfill%
    \begin{subfigure}[t]{0.33\linewidth}
        \centering
        \includegraphics[width=\linewidth]{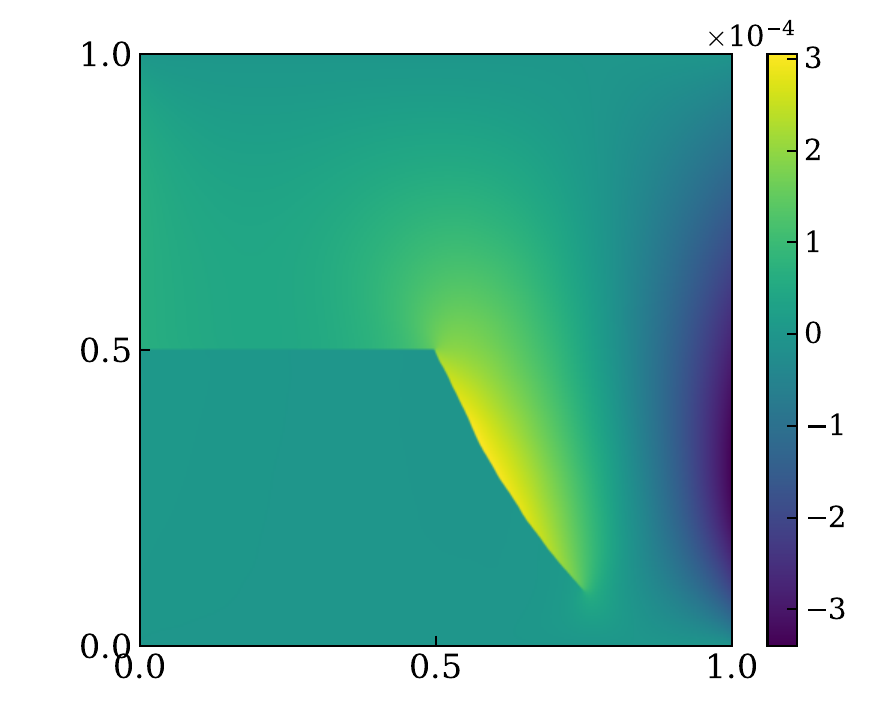}
        \caption{}
    \end{subfigure}%
    \hfill%
    \begin{subfigure}[t]{0.33\linewidth}
        \centering
        \includegraphics[width=\linewidth]{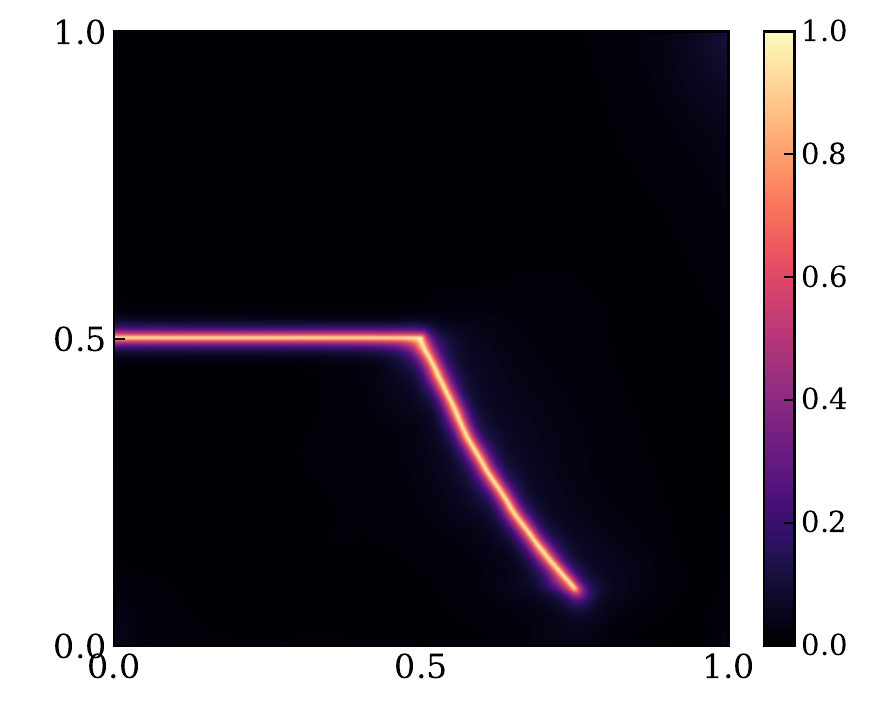}
        \caption{}
    \end{subfigure}

    \vspace{0.6ex}
    \begin{subfigure}[t]{0.33\linewidth}
        \centering
        \includegraphics[width=\linewidth]{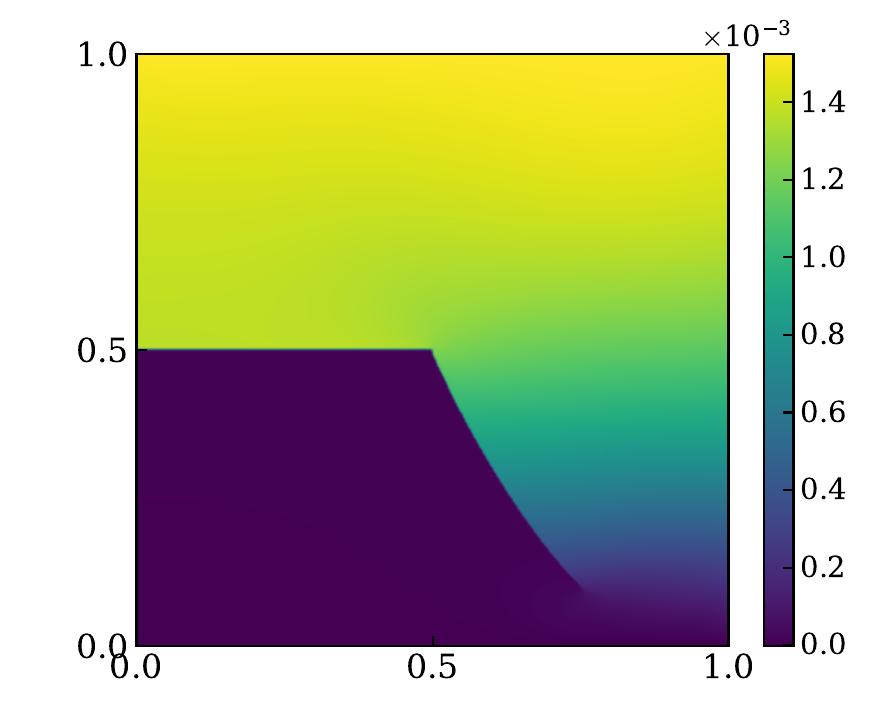}
        \caption{}
    \end{subfigure}%
    \hfill%
    \begin{subfigure}[t]{0.33\linewidth}
        \centering
        \includegraphics[width=\linewidth]{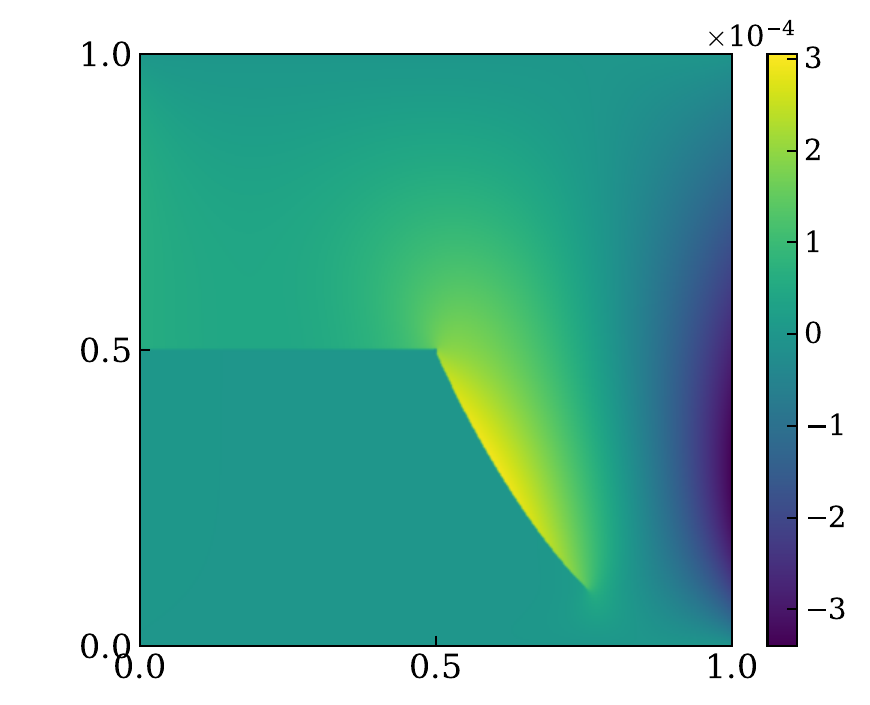}
        \caption{}
    \end{subfigure}%
    \hfill%
    \begin{subfigure}[t]{0.33\linewidth}
        \centering
        \includegraphics[width=\linewidth]{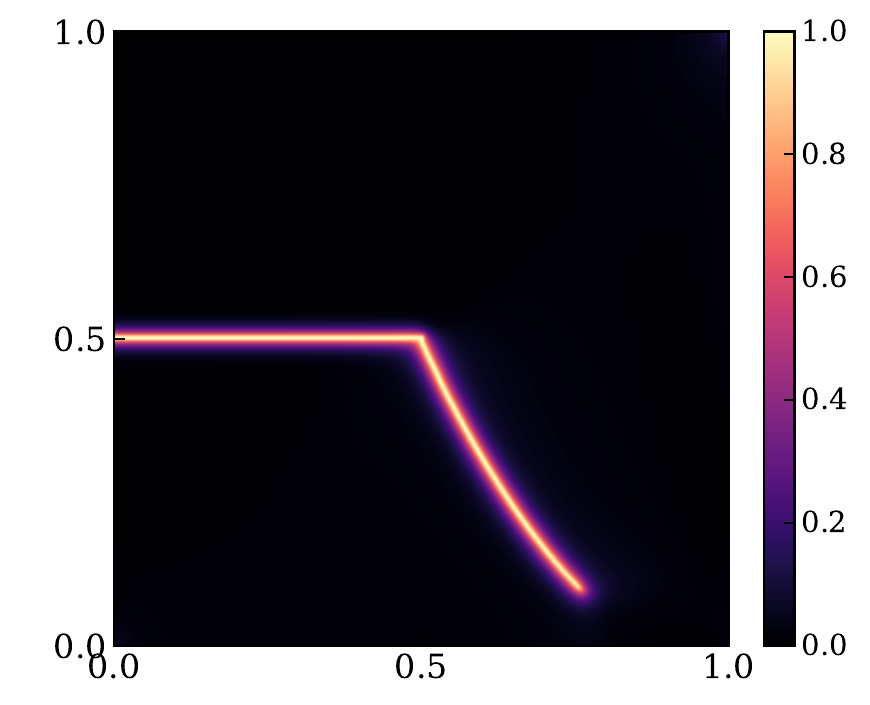}
        \caption{}
    \end{subfigure}
    \caption{Single-edge-notched shear during propagation, at
    $\delta = 1.525 \times 10^{-3}\,\mathrm{mm}$. (a)--(c)~Displacement
    components $u$, $v$ and phase field of the second-order deep
    energy solution; (d)--(f)~the finite element fields at the same
    load level. Displacements in mm; each displacement component
    shares one color scale between the two rows.}
    \label{fig:sens-prop}
\end{figure}

Fig.~\ref{fig:sens-prop} shows the corresponding comparison while the
crack is running, at $\delta = 1.525 \times 10^{-3}\,\mathrm{mm}$;
displacement and phase field agree with the reference at this
intermediate state as well, not only after failure.

\begin{figure}[!htb]
    \centering
    \begin{subfigure}[t]{0.33\linewidth}
        \centering
        \includegraphics[width=\linewidth]{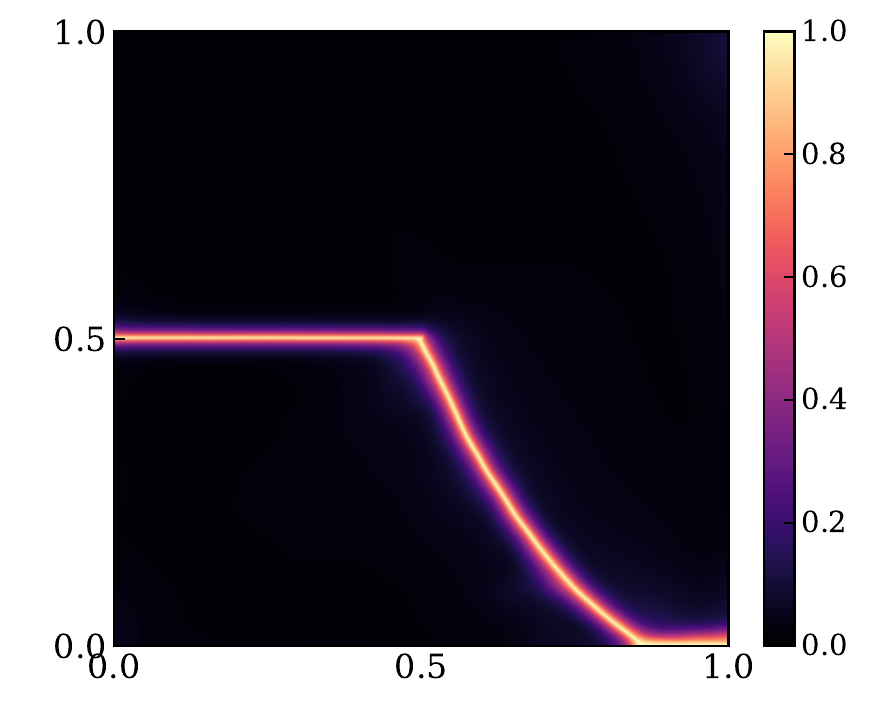}
        \caption{}
    \end{subfigure}%
    \hfill%
    \begin{subfigure}[t]{0.33\linewidth}
        \centering
        \includegraphics[width=\linewidth]{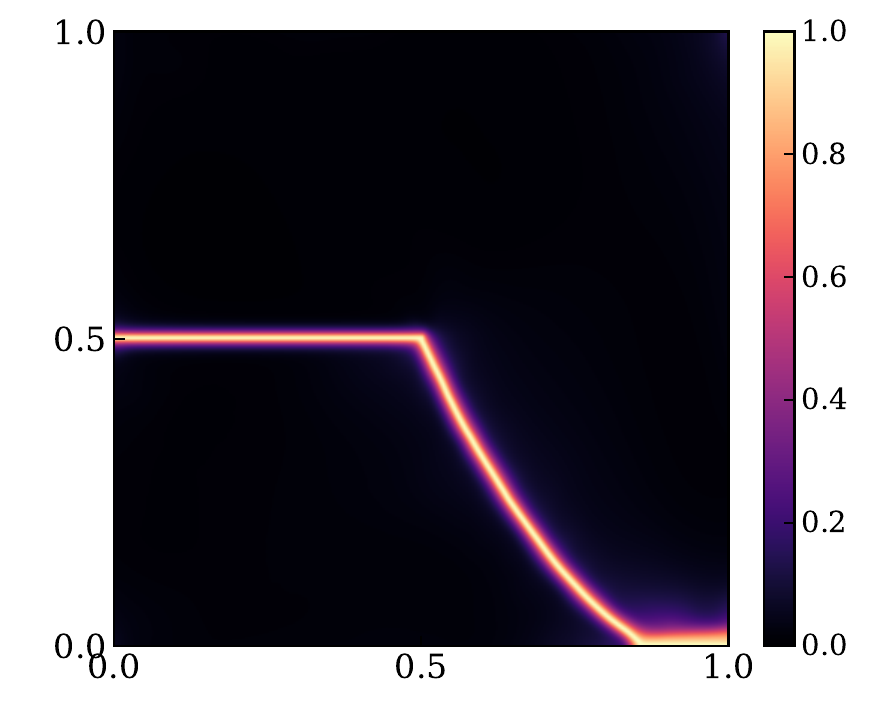}
        \caption{}
    \end{subfigure}%
    \hfill%
    \begin{subfigure}[t]{0.33\linewidth}
        \centering
        \includegraphics[width=\linewidth]{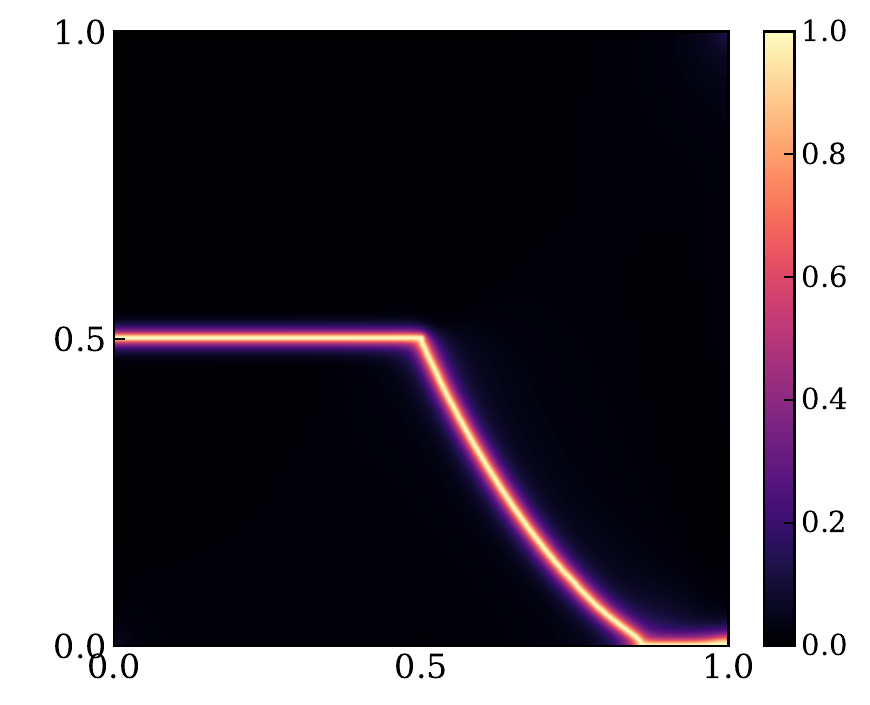}
        \caption{}
    \end{subfigure}
    \caption{Single-edge-notched shear, final phase fields.
    (a)~Second-order and (b)~fourth-order deep energy solutions,
    (c)~finite element reference; all renderings share one color
    scale.}
    \label{fig:sens-fields}
\end{figure}

\begin{figure}[!htb]
    \centering
    \begin{subfigure}[t]{0.33\linewidth}
        \centering
        \includegraphics[width=\linewidth]{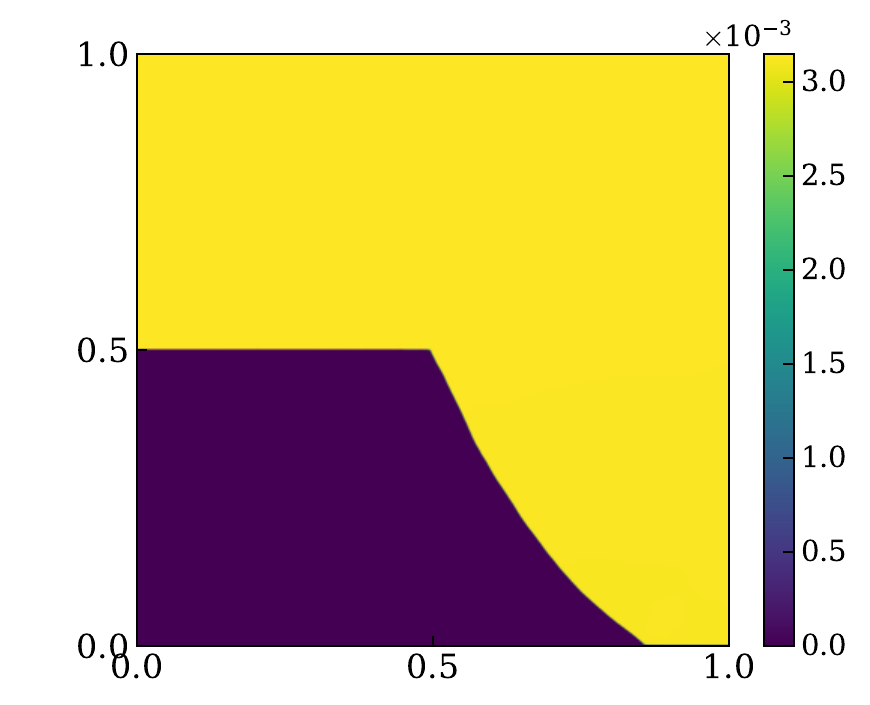}
        \caption{}
    \end{subfigure}%
    \hspace{0.04\linewidth}%
    \begin{subfigure}[t]{0.33\linewidth}
        \centering
        \includegraphics[width=\linewidth]{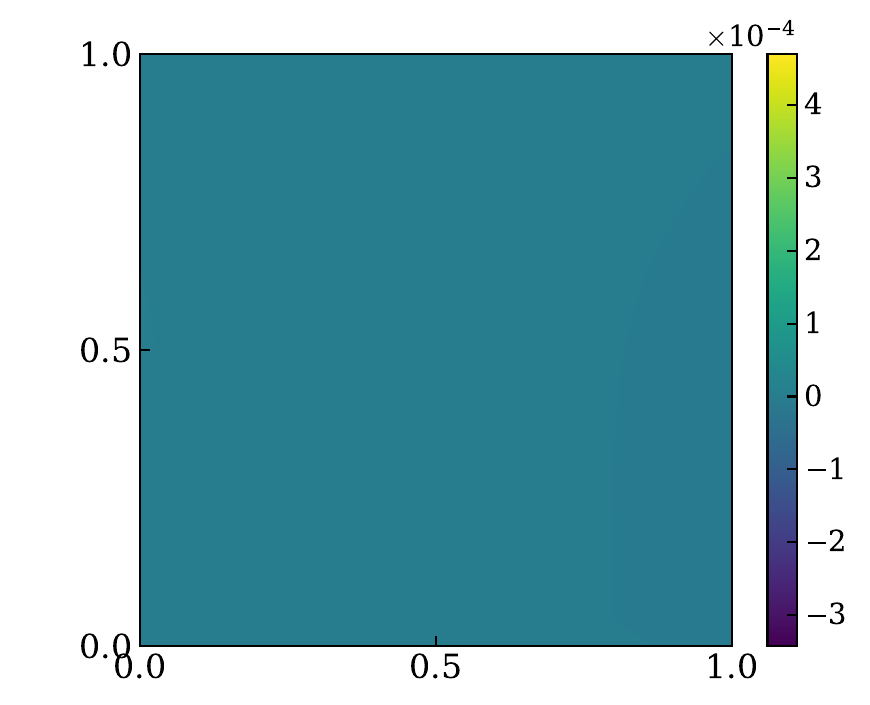}
        \caption{}
    \end{subfigure}

    \vspace{0.6ex}
    \begin{subfigure}[t]{0.33\linewidth}
        \centering
        \includegraphics[width=\linewidth]{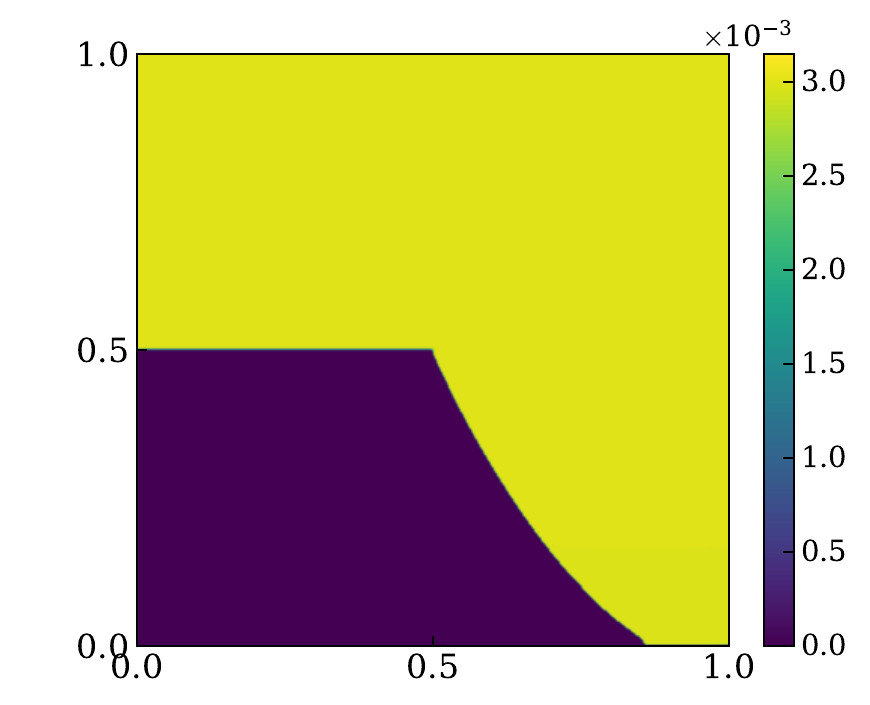}
        \caption{}
    \end{subfigure}%
    \hspace{0.04\linewidth}%
    \begin{subfigure}[t]{0.33\linewidth}
        \centering
        \includegraphics[width=\linewidth]{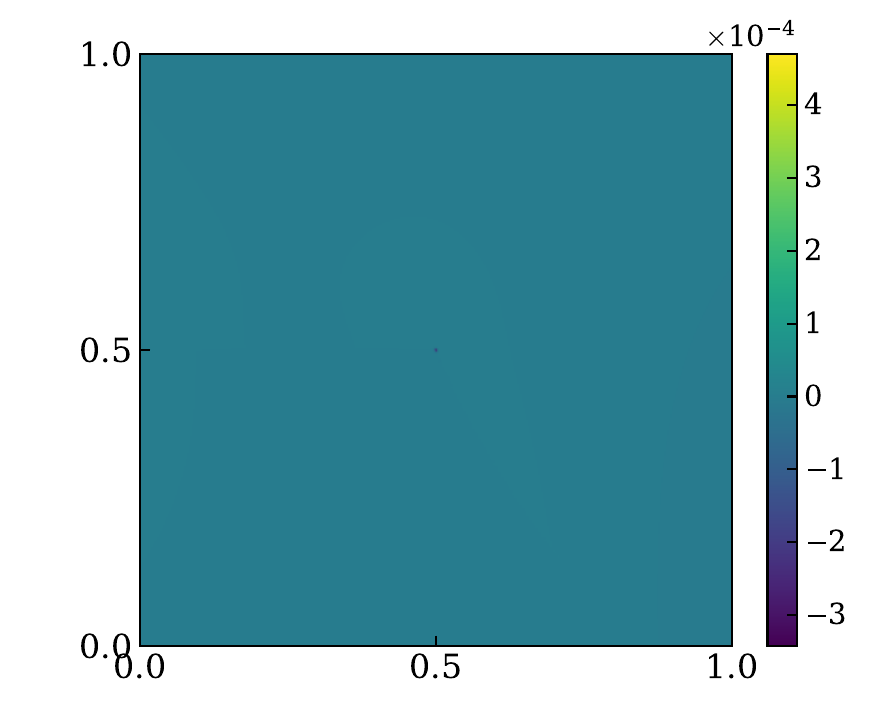}
        \caption{}
    \end{subfigure}
    \caption{Single-edge-notched shear, displacement components of the
    failed state in mm. (a),(b)~$u$ and $v$ of the second-order deep
    energy solution; (c),(d)~the finite element fields; each component
    shares one color scale between the two rows.}
    \label{fig:sens-u}
\end{figure}

The final phase fields are compared in Fig.~\ref{fig:sens-fields} and
the displacement components of the failed state in
Fig.~\ref{fig:sens-u}. The computed path coincides with the reference
along its full trajectory, including the curved approach to the
bottom edge, the fourth-order band is again the more compact one, and
the two parts of the specimen separate across the fully developed
band in agreement with the finite element fields.

Since the minimization problem is non-convex and the optimizer is
stochastic, the robustness of these results against the network
initialization and the sampling sequence is examined by repeating the
shear test with four random seeds, all other settings unchanged. All
four runs complete the full loading program, produce the single curved
crack band, and retain every feature of the response, with a peak load
of $66.3 \pm 0.3\,\mathrm{N}$, a softening valley of
$42.6 \pm 1.0\,\mathrm{N}$ and a second local maximum of
$50.3 \pm 1.3\,\mathrm{N}$ across the ensemble, quoted as mean and
sample standard deviation. The ensemble mean lies within $0.1\%$ of
the reference peak, inside its own scatter of $0.5\%$, so the
single-run agreement quoted above is representative rather than a
fortunate draw; no run is excluded. The same specimen serves as the testbed of
\ref{app:failure}, where the ingredients of Section~\ref{sec:method}
are removed or replaced one at a time.

\FloatBarrier
% ---------------------------------------------------------------------
\subsection{Crack branching}
\label{subsec:branching}

Branching is the first topology-changing example. The specimen and
loading are those of the shear test
(Fig.~\ref{fig:setup-sen}b), but the strain energy is left
undecomposed, $\edens^{+} = \edens$ in~Eq.~\eqref{eq:pi}, so that damage is
driven isotropically, the setting in which fast cracks are known to
branch~\cite{borden2012}. Branching in the physical sense is a dynamic
instability that sets in above a critical crack speed, and a
quasi-static analysis does not reproduce it. What the undecomposed
energy admits here is a symmetric bifurcation of the incremental
minimization, and the comparison below is accordingly made against a
finite element solution of the same functional rather than against
experiment. Two implementation details from
Section~\ref{sec:method} matter here. The crack stratum of the
mixture of Eq.~\eqref{eq:mixture} is replaced by a fixed band that is
symmetric about the pre-existing crack, since the adaptive stratum
follows the current phase field and would otherwise feed points, and
thereby resolution, asymmetrically to whichever branch happens to lead;
and the appended-point component supplies fresh points on the newly
created branches so that the growing pattern does not dilute the fixed
budget.

\begin{figure}[!htb]
    \centering
    \begin{subfigure}[b]{0.74\linewidth}
        \centering
        \begin{tikzpicture}
        \begin{axis}[
            paperaxis, width=\linewidth, height=6.0cm,
            xlabel={$\delta$ ($10^{-3}$ mm)}, ylabel={$F$ (N)},
            xmin=0, xmax=3.25, ymin=0, ymax=72,
            legend style={at={(0.5, -0.28)}, anchor=north,
                          legend columns=3,
                          /tikz/every even column/.append style=
                              {column sep=0.5cm}},
        ]
            \addplot[black, thick, each nth point=2]
                table[col sep=comma,
                      x expr=\thisrow{delta_mm}*1000, y=F]
                {figures/data/bifurc_fem.csv};
            \addlegendentry{FEM}
            \addplot[cblue, thick]
                table[col sep=comma,
                      x expr=\thisrow{delta_mm}*1000, y=F]
                {figures/data/bifurc_dem2.csv};
            \addlegendentry{second order}
            \addplot[cred, thick, dashed]
                table[col sep=comma,
                      x expr=\thisrow{delta_mm}*1000, y=F]
                {figures/data/bifurc_dem4.csv};
            \addlegendentry{fourth order}
        \end{axis}
        \end{tikzpicture}
        \caption{}
        \label{fig:branching-fd}
    \end{subfigure}

    \vspace{0.6ex}
    \begin{subfigure}[t]{0.33\linewidth}
        \centering
        \includegraphics[width=\linewidth]{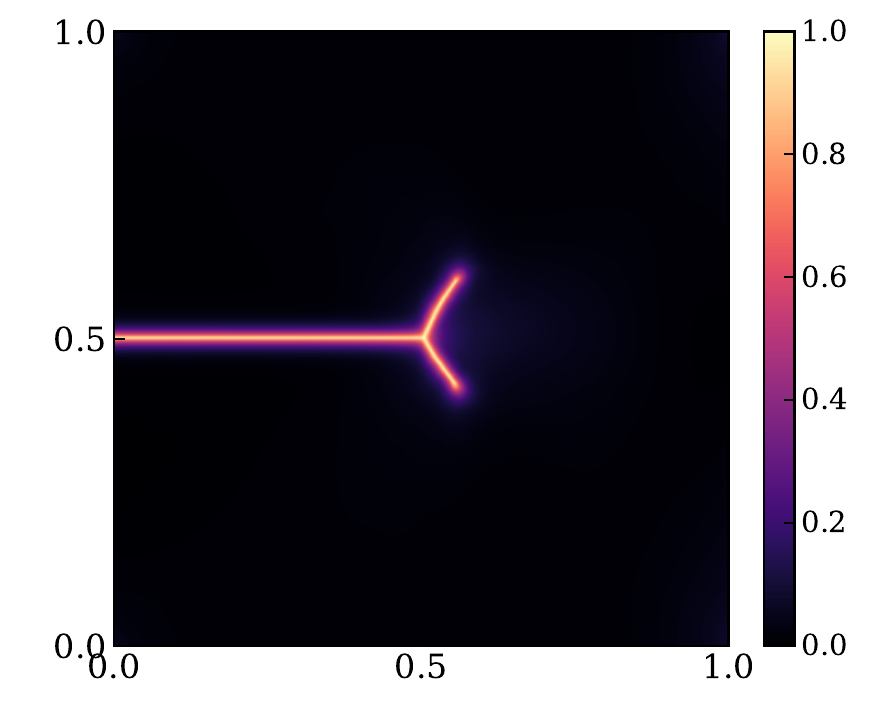}
        \caption{$\delta = 1.03 \times 10^{-3}\,\mathrm{mm}$}
    \end{subfigure}
    \hspace{0.04\linewidth}
    \begin{subfigure}[t]{0.33\linewidth}
        \centering
        \includegraphics[width=\linewidth]{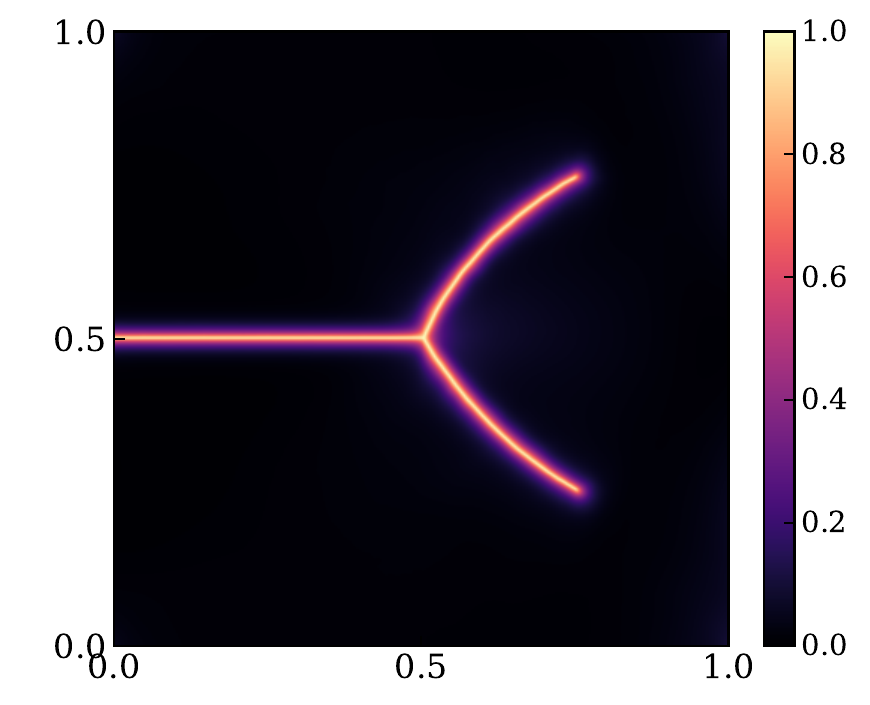}
        \caption{$\delta = 1.53 \times 10^{-3}\,\mathrm{mm}$}
    \end{subfigure}
    \caption{Crack branching under isotropic driving.
    (a)~Load--displacement curves of the second- and fourth-order
    solutions against the finite element reference; (b),(c)~phase field
    of the second-order solution at two load levels. The crack
    splits into a symmetric Y-shaped pattern with no geometric crack
    description involved at any stage.}
    \label{fig:branching}
\end{figure}

\begin{figure}[!htb]
    \centering
    \begin{subfigure}[t]{0.33\linewidth}
        \centering
        \includegraphics[width=\linewidth]{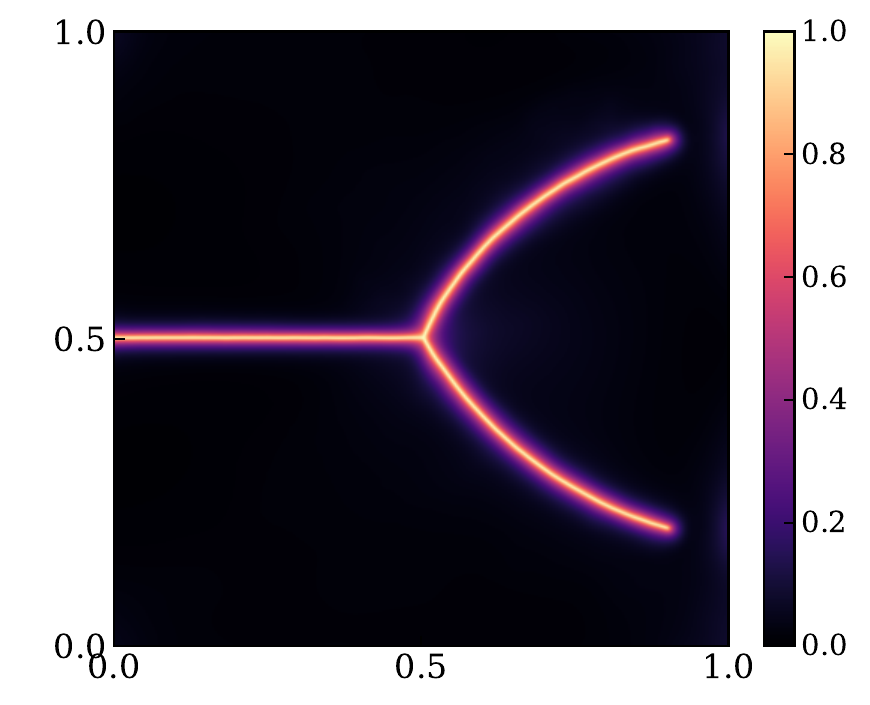}
        \caption{}
    \end{subfigure}%
    \hfill%
    \begin{subfigure}[t]{0.33\linewidth}
        \centering
        \includegraphics[width=\linewidth]{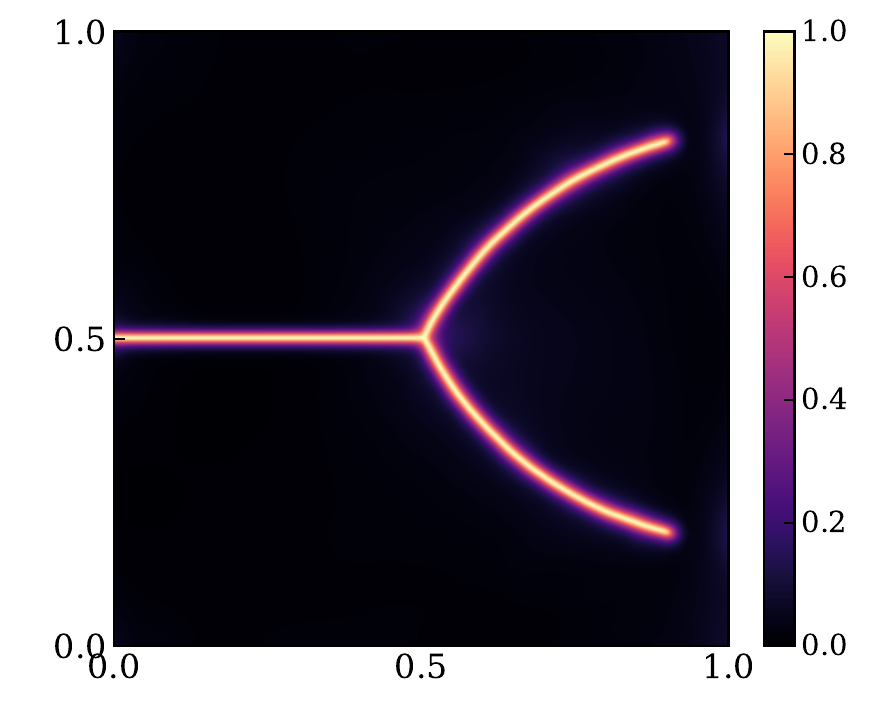}
        \caption{}
    \end{subfigure}%
    \hfill%
    \begin{subfigure}[t]{0.33\linewidth}
        \centering
        \includegraphics[width=\linewidth]{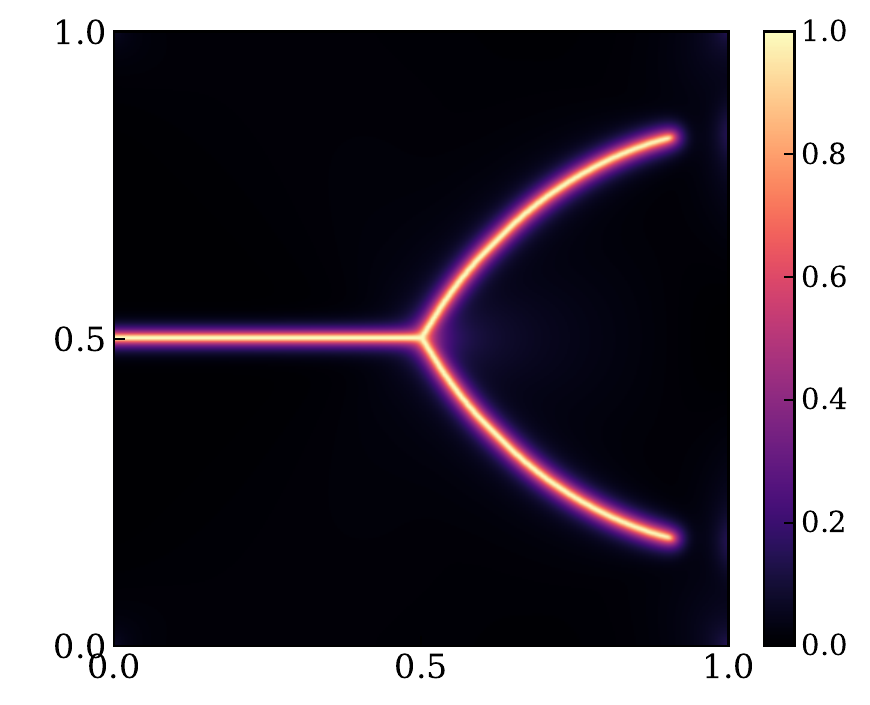}
        \caption{}
    \end{subfigure}

    \vspace{0.6ex}
    \begin{subfigure}[t]{0.33\linewidth}
        \centering
        \includegraphics[width=\linewidth]{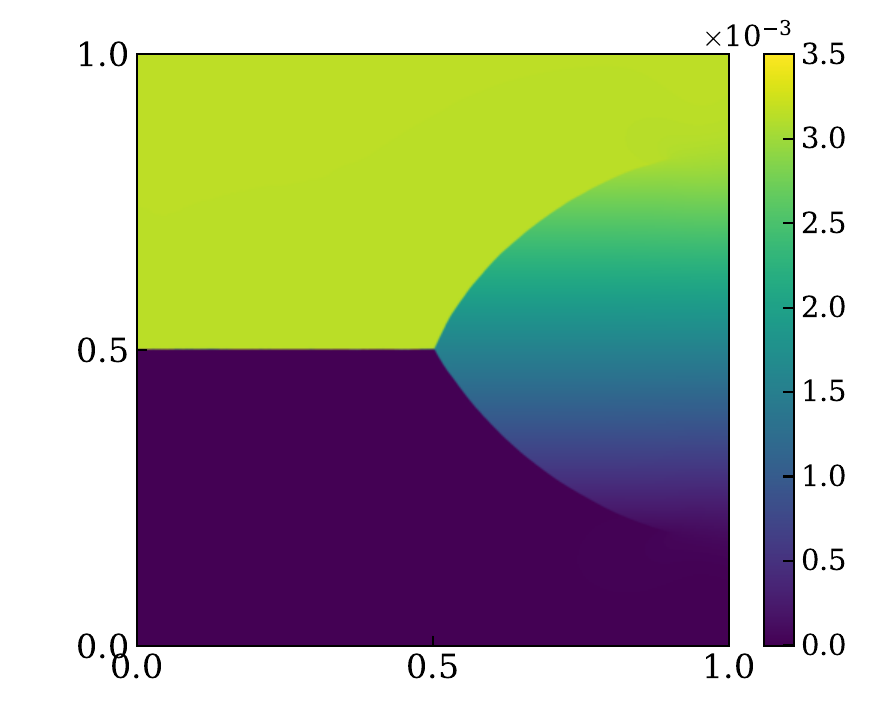}
        \caption{}
    \end{subfigure}%
    \hspace{0.04\linewidth}%
    \begin{subfigure}[t]{0.33\linewidth}
        \centering
        \includegraphics[width=\linewidth]{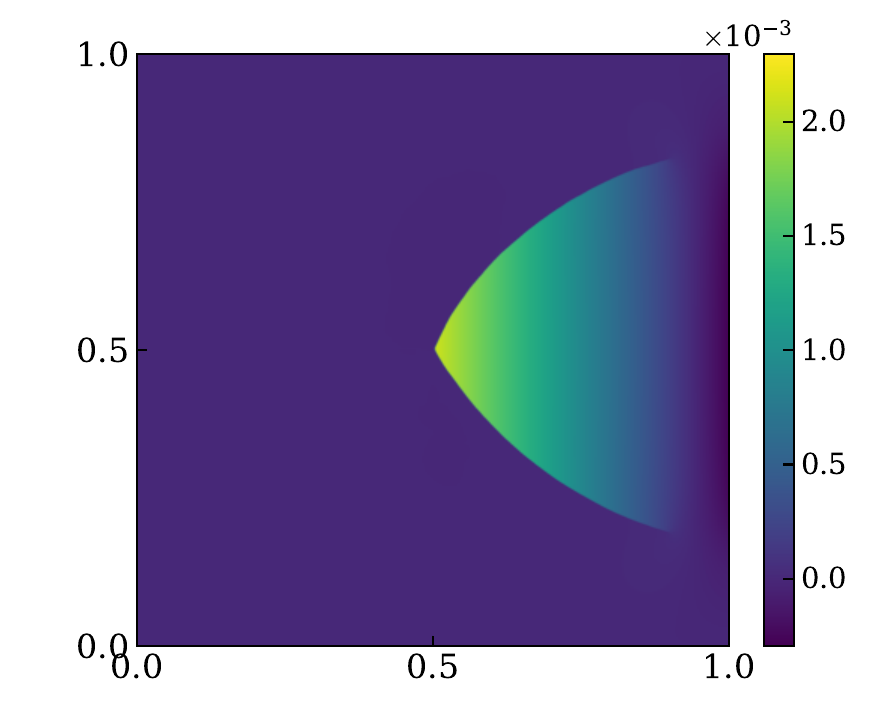}
        \caption{}
    \end{subfigure}

    \vspace{0.6ex}
    \begin{subfigure}[t]{0.33\linewidth}
        \centering
        \includegraphics[width=\linewidth]{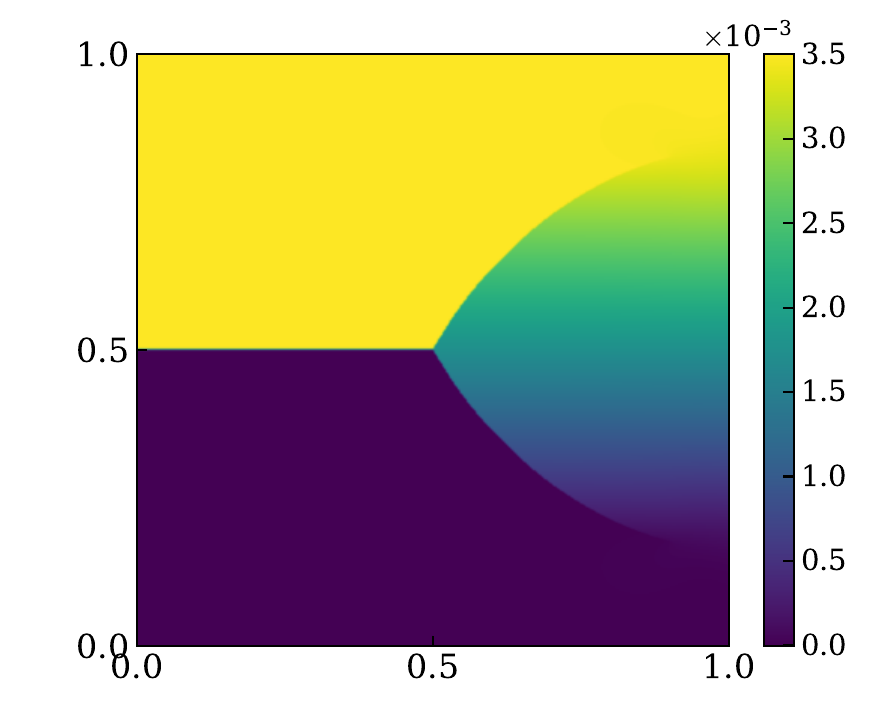}
        \caption{}
    \end{subfigure}%
    \hspace{0.04\linewidth}%
    \begin{subfigure}[t]{0.33\linewidth}
        \centering
        \includegraphics[width=\linewidth]{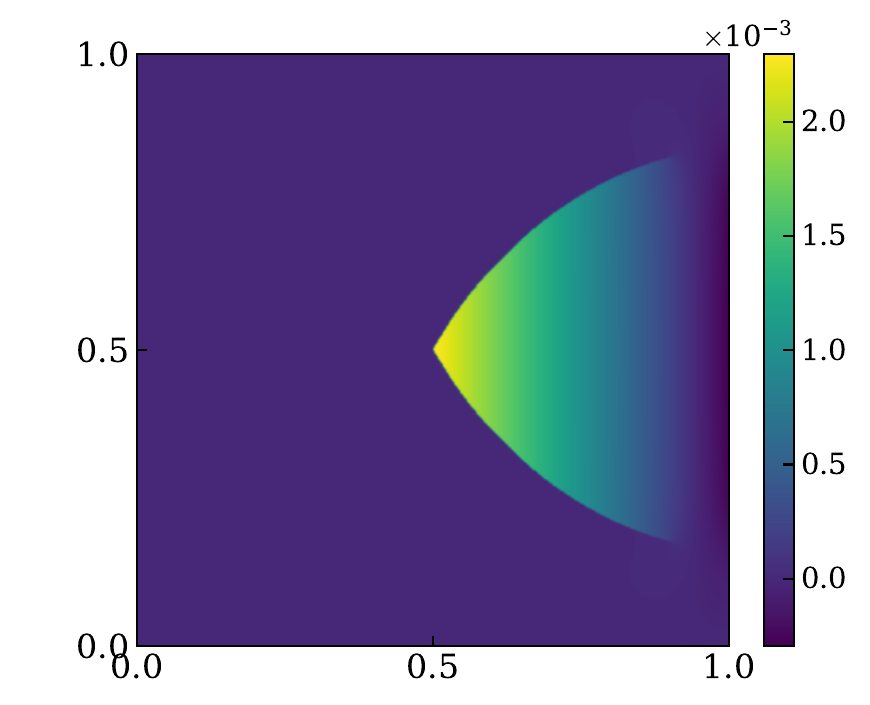}
        \caption{}
    \end{subfigure}
    \caption{Crack branching, final fields. Phase fields of the
    (a)~second-order and (b)~fourth-order deep energy solutions and
    (c)~the finite element reference; the fourth-order solution
    retains the symmetric Y-shaped pattern. Displacement components in
    mm, (d),(e)~$u$ and $v$ of the second-order solution and
    (f),(g)~the finite element fields, each component on one color
    scale shared with its counterpart.}
    \label{fig:branching-fields}
\end{figure}

The crack then leaves the pre-existing tip and splits into a symmetric
pair of branches, forming a Y-shaped pattern without any crack
tracking, kinking criterion or damage threshold
(Fig.~\ref{fig:branching}). The branches curve smoothly apart and
arrest near the right edge, and the pattern remains symmetric about the
crack plane to within the band width. The load--displacement curve
peaks at $59.3\,\mathrm{N}$, about $8\%$ below the finite element
reference computed with the same undecomposed energy, and decays over a
long tail as the branches extend. The fourth-order run on the same
configuration was exploratory, since the higher-order density might
plausibly have suppressed the bifurcation; it does not. The
fourth-order crack follows a nearly identical Y-shaped path
(Fig.~\ref{fig:branching-fields}), with a peak of $57.8\,\mathrm{N}$,
so the branching behavior of the functional carries over unchanged.
The displacement components of the final state are compared with the
finite element fields in Fig.~\ref{fig:branching-fields}d--g.

\FloatBarrier
% ---------------------------------------------------------------------
\subsection{Coalescence of en-echelon cracks}
\label{subsec:coalescence}

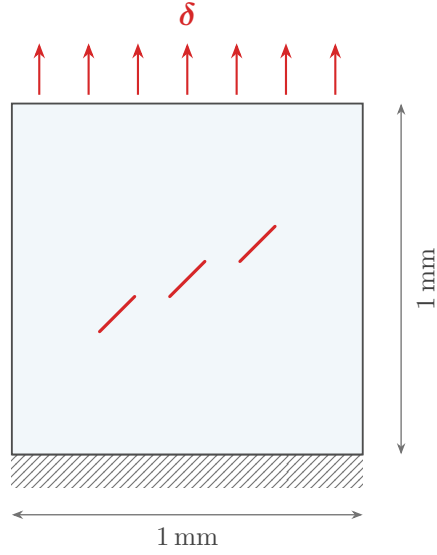
\begin{figure}[!htb]
    \centering
    \begin{tikzpicture}[scale=1.45]
        \fill[clamphatch] (0, -0.30) rectangle (3.2, 0);
        \draw[black!70, line width=0.7pt] (0, 0) -- (3.2, 0);
        \draw[spec] (0, 0) rectangle (3.2, 3.2);
        \draw[crackline] (0.80, 1.12) -- (1.12, 1.44);
        \draw[crackline] (1.44, 1.44) -- (1.76, 1.76);
        \draw[crackline] (2.08, 1.76) -- (2.40, 2.08);
        \foreach \x in {0.25, 0.7, 1.15, 1.6, 2.05, 2.5, 2.95} {
            \draw[loadarrow] (\x, 3.28) -- (\x, 3.75);
        }
        \node[cred, font=\normalsize, anchor=south] at (1.6, 3.82)
            {$\bm{\delta}$};
        \draw[dimline] (0, -0.55) -- (3.2, -0.55);
        \node[dimfont, below] at (1.6, -0.57) {$1\,\mathrm{mm}$};
        \draw[dimline] (3.55, 0) -- (3.55, 3.2);
        \node[dimfont, rotate=90, below] at (3.60, 1.6) {$1\,\mathrm{mm}$};
    \end{tikzpicture}
    \caption{Coalescence specimen. Three parallel pre-existing cracks of
    length $0.14\,\mathrm{mm}$, inclined at $45^{\circ}$ and arranged en
    echelon, in the square plate with a fixed bottom edge under
    vertical tension.
   }
    \label{fig:setup-coal}
\end{figure}

In the second topology example several cracks merge into one, in the
arrangement introduced for rock specimens by Zhou et
al.~\cite{zhou2018} and adopted since as a coalescence
benchmark~\cite{manav2024,hamdi2026}. Three parallel pre-existing
cracks are arranged en echelon across the specimen, each inclined at
$45^{\circ}$, and the specimen is loaded in vertical tension in $100$
increments of $10^{-5}\,\mathrm{mm}$ (Fig.~\ref{fig:setup-coal}).

\begin{figure}[!htb]
    \centering
    \begin{subfigure}[b]{0.74\linewidth}
        \centering
        \begin{tikzpicture}
        \begin{axis}[
            paperaxis, width=\linewidth, height=6.0cm,
            xlabel={$\delta$ ($10^{-3}$ mm)}, ylabel={$F$ (N)},
            xmin=0, xmax=1.02, ymin=0, ymax=215,
            legend style={at={(0.5, -0.28)}, anchor=north,
                          legend columns=3,
                          /tikz/every even column/.append style=
                              {column sep=0.5cm}},
        ]
            \addplot[black, thick, each nth point=2]
                table[col sep=comma,
                      x expr=\thisrow{delta_mm}*1000, y=F]
                {figures/data/coal_fem.csv};
            \addlegendentry{FEM}
            \addplot[cblue, thick]
                table[col sep=comma,
                      x expr=\thisrow{delta_mm}*1000, y=F]
                {figures/data/coal_dem2.csv};
            \addlegendentry{second order}
            \addplot[cred, thick, dashed]
                table[col sep=comma,
                      x expr=\thisrow{delta_mm}*1000, y=F]
                {figures/data/coal_dem4.csv};
            \addlegendentry{fourth order}
        \end{axis}
        \end{tikzpicture}
        \caption{}
        \label{fig:coal-fd}
    \end{subfigure}

    \vspace{0.6ex}
    \begin{subfigure}[t]{0.33\linewidth}
        \centering
        \includegraphics[width=\linewidth]{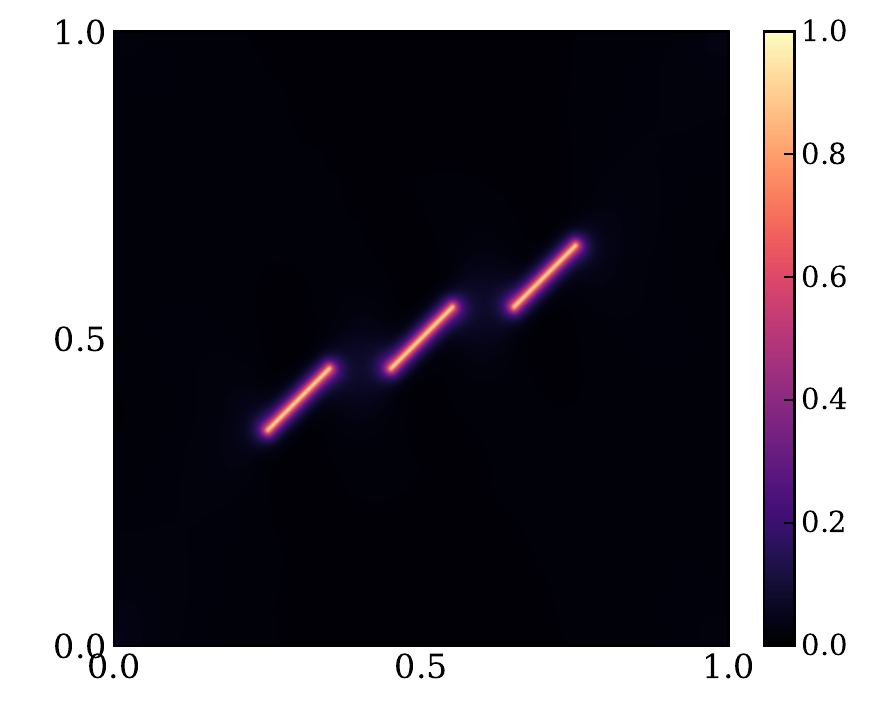}
        \caption{$\delta = 5.1 \times 10^{-4}\,\mathrm{mm}$}
    \end{subfigure}%
    \hfill%
    \begin{subfigure}[t]{0.33\linewidth}
        \centering
        \includegraphics[width=\linewidth]{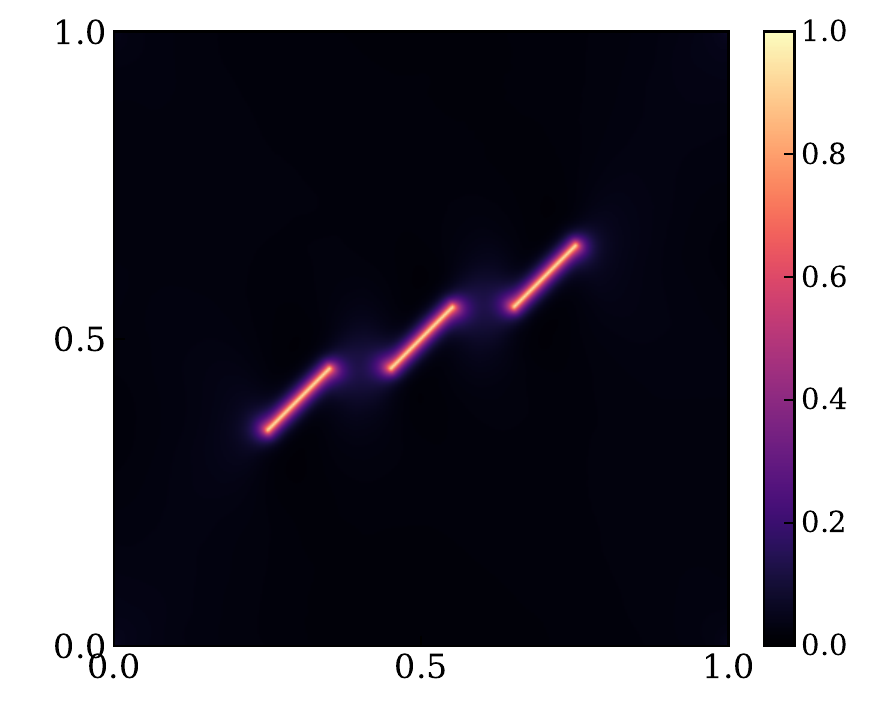}
        \caption{$\delta = 6.1 \times 10^{-4}\,\mathrm{mm}$}
    \end{subfigure}%
    \hfill%
    \begin{subfigure}[t]{0.33\linewidth}
        \centering
        \includegraphics[width=\linewidth]{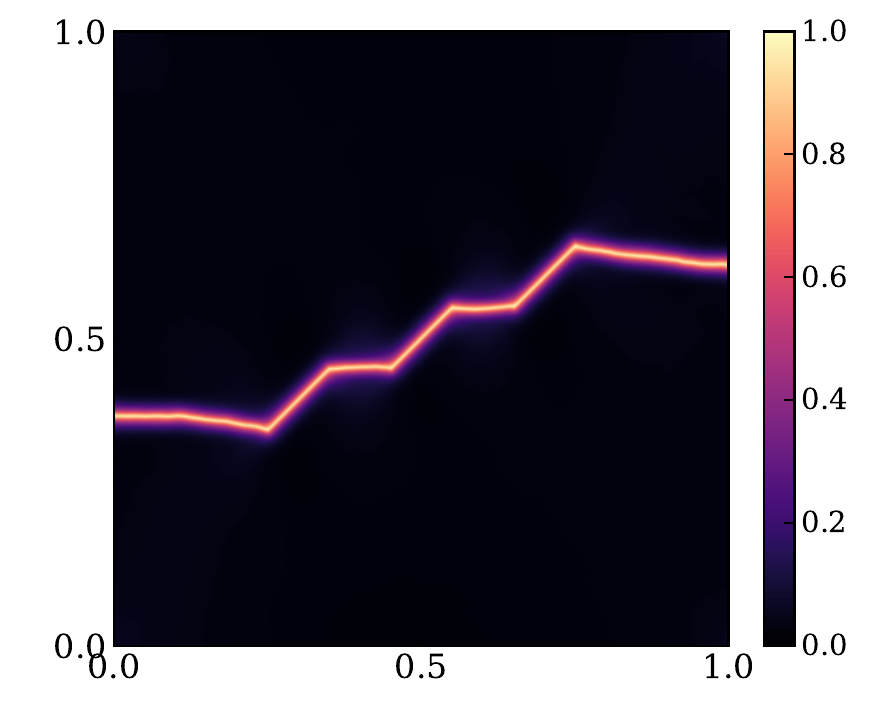}
        \caption{$\delta = 7.1 \times 10^{-4}\,\mathrm{mm}$}
    \end{subfigure}
    \caption{Coalescence of three en-echelon pre-existing cracks under
    tension. (a)~Load--displacement curves of the second- and
    fourth-order solutions against the finite element reference;
    (b)--(d)~phase field of the second-order solution at three load
    levels. The inner tips curve toward one another, link, and
    form a single staircase crack, and the multi-segment seeding
    requires no change to the method.}
    \label{fig:coalescence}
\end{figure}

\begin{figure}[!htb]
    \centering
    \begin{subfigure}[t]{0.33\linewidth}
        \centering
        \includegraphics[width=\linewidth]{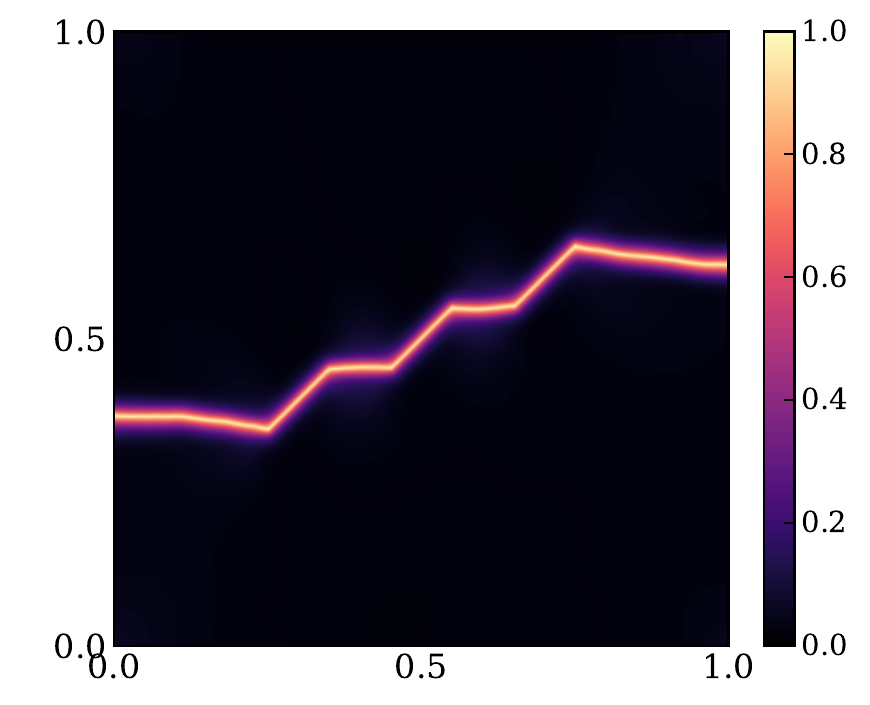}
        \caption{}
    \end{subfigure}%
    \hfill%
    \begin{subfigure}[t]{0.33\linewidth}
        \centering
        \includegraphics[width=\linewidth]{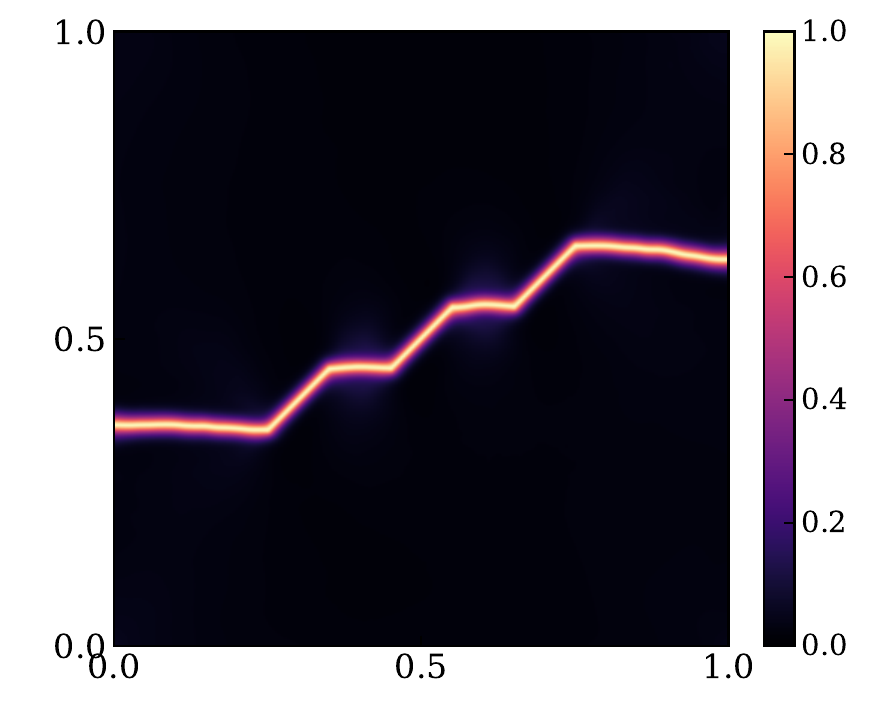}
        \caption{}
    \end{subfigure}%
    \hfill%
    \begin{subfigure}[t]{0.33\linewidth}
        \centering
        \includegraphics[width=\linewidth]{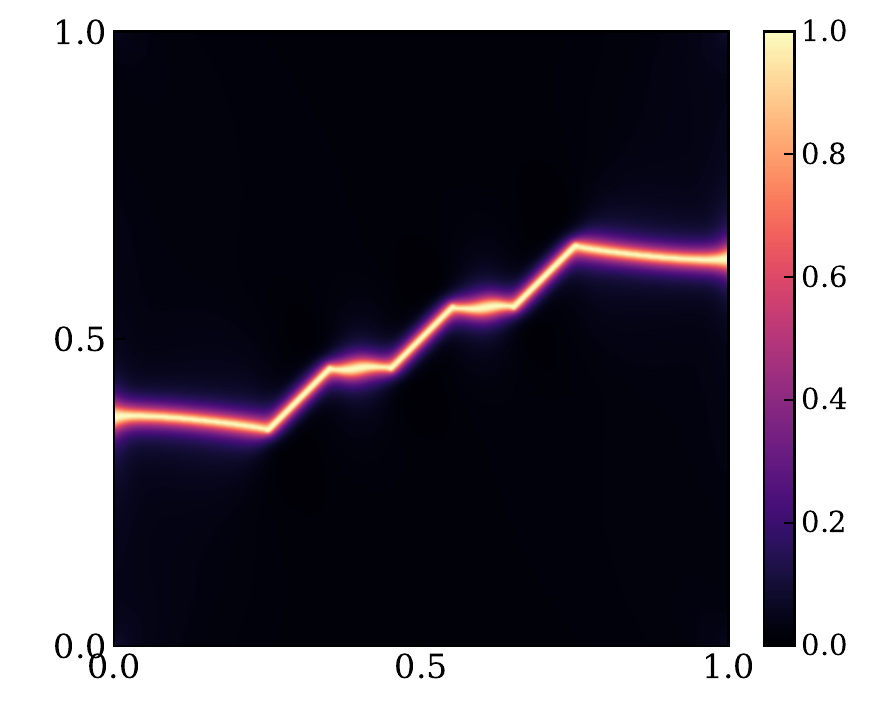}
        \caption{}
    \end{subfigure}

    \vspace{0.6ex}
    \begin{subfigure}[t]{0.33\linewidth}
        \centering
        \includegraphics[width=\linewidth]{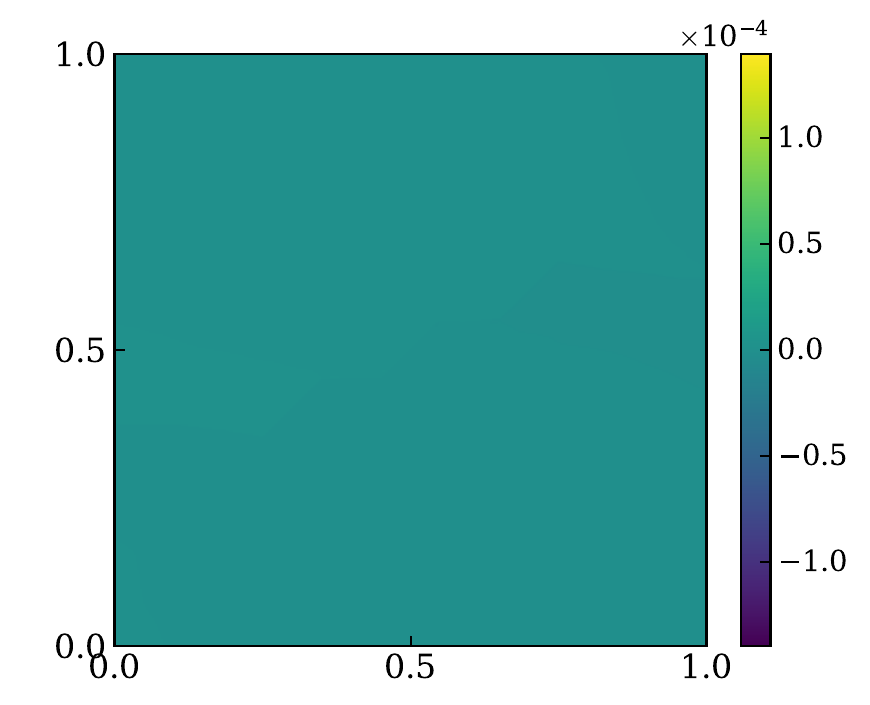}
        \caption{}
    \end{subfigure}%
    \hspace{0.04\linewidth}%
    \begin{subfigure}[t]{0.33\linewidth}
        \centering
        \includegraphics[width=\linewidth]{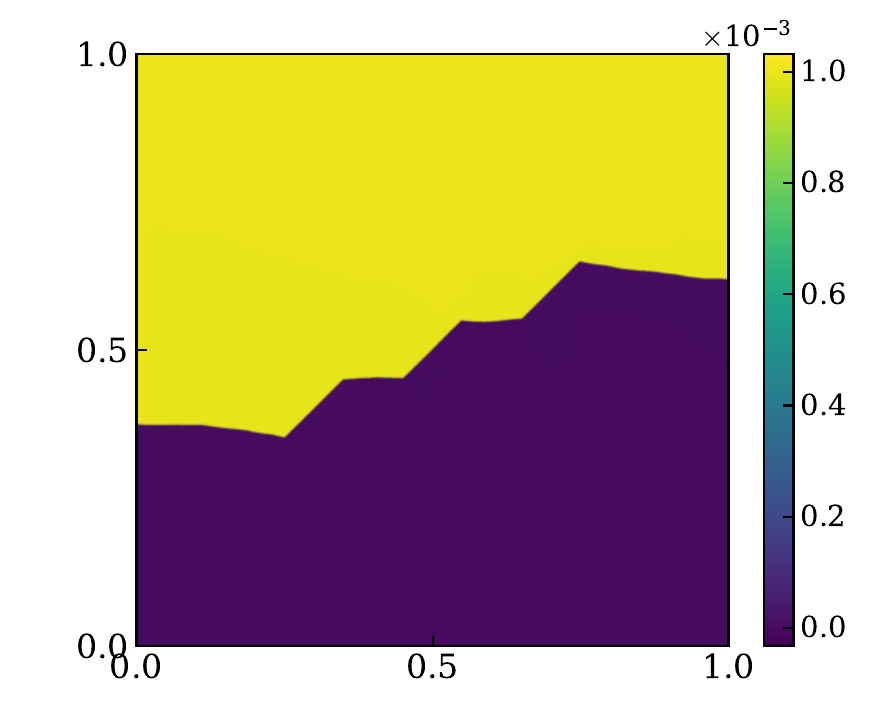}
        \caption{}
    \end{subfigure}

    \vspace{0.6ex}
    \begin{subfigure}[t]{0.33\linewidth}
        \centering
        \includegraphics[width=\linewidth]{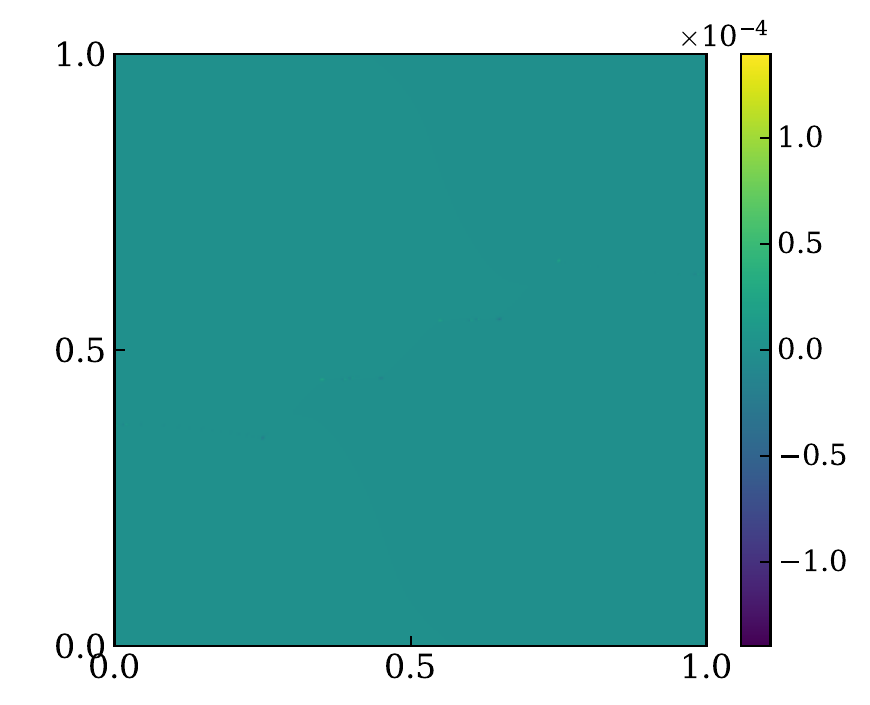}
        \caption{}
    \end{subfigure}%
    \hspace{0.04\linewidth}%
    \begin{subfigure}[t]{0.33\linewidth}
        \centering
        \includegraphics[width=\linewidth]{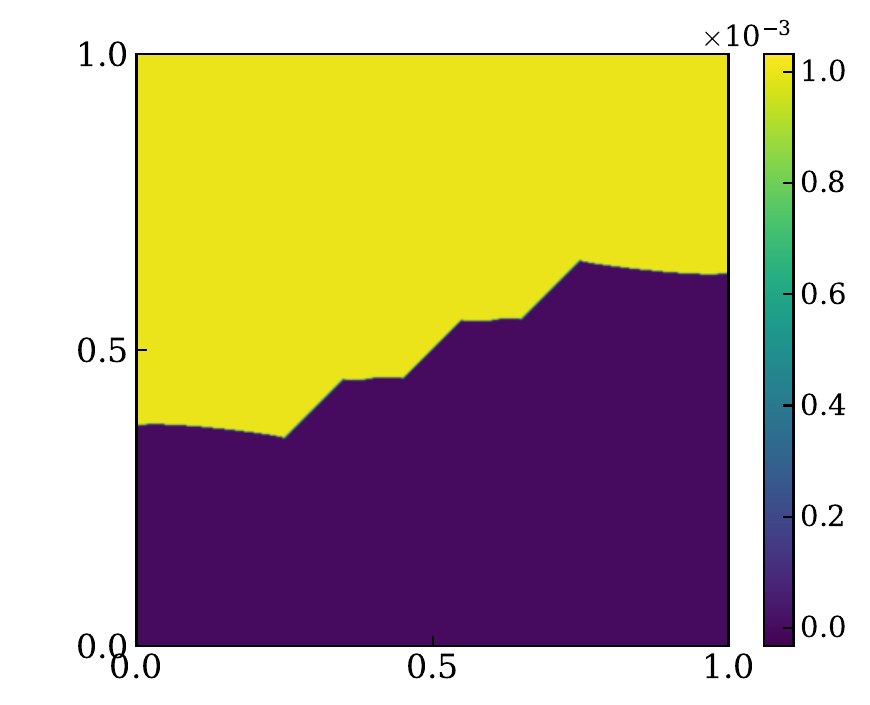}
        \caption{}
    \end{subfigure}
    \caption{Coalescence test, final fields. Phase fields of the
    (a)~second-order and (b)~fourth-order deep energy solutions and
    (c)~the finite element reference; both orders link the three
    cracks along the same staircase. Displacement components in mm,
    (d),(e)~$u$ and $v$ of the second-order solution and (f),(g)~the
    finite element fields, each component on one color scale shared
    with its counterpart.}
    \label{fig:coal-fields}
\end{figure}

 The seeded profile of Eq.~\eqref{eq:profile2} takes
the minimum distance over the three segments, as described in
Section~\ref{subsec:setup}, and no other ingredient changes. Under load the inner crack tips curve
toward one another, link, and form a single staircase crack across the
specimen, the linkage pattern reported for this arrangement in the
phase-field literature~\cite{zhou2018,manav2024}
(Fig.~\ref{fig:coalescence}). The load--displacement curve
peaks at $195.4\,\mathrm{N}$ at $6.4 \times 10^{-4}\,\mathrm{mm}$,
$5.8\%$ above the finite element reference of $184.8\,\mathrm{N}$ at
$6.1 \times 10^{-4}\,\mathrm{mm}$, and drops to complete failure once
the outer segments reach the lateral boundaries. The fourth-order run
links the cracks along the same staircase
(Fig.~\ref{fig:coal-fields}) and peaks at $182.8\,\mathrm{N}$, about
$6\%$ below the second-order value and about $1\%$ below the
reference.
The displacement components of the final state are compared with the
finite element fields in Fig.~\ref{fig:coal-fields}d--g.

\FloatBarrier
% ---------------------------------------------------------------------
\subsection{Plate with a circular hole}
\label{subsec:hole}

Before any cracking, the computed fields of this example are
verified against a closed-form elastic solution, which tests the
combination of domain mask, sampling and quadrature; the same
specimen then nucleates cracks with no pre-existing crack present.

\begin{figure}[!htb]
    \centering
    \begin{tikzpicture}[scale=1.45]
        \fill[clamphatch] (-0.30, 0) rectangle (0, 3.2);
        \draw[black!70, line width=0.7pt] (0, 0) -- (0, 3.2);
        \draw[spec] (0, 0) rectangle (3.2, 3.2);
        \fill[white, draw=black!70, line width=0.7pt]
            (1.6, 1.6) circle (0.32);
        \fill[black!60] (1.6, 1.6) circle (0.9pt);
        \draw[black!55, line width=0.5pt, -{Stealth[length=1.5mm]}]
            (1.6, 1.6) -- ({1.6+0.32*cos(40)}, {1.6+0.32*sin(40)});
        \node[dimfont, anchor=west] at (1.80, 1.95) {$0.1\,\mathrm{mm}$};
        \foreach \y in {0.25, 0.7, 1.15, 1.6, 2.05, 2.5, 2.95} {
            \draw[loadarrow] (3.28, \y) -- (3.75, \y);
        }
        \node[cred, font=\normalsize, anchor=west] at (3.80, 1.6)
            {$\bm{\delta}$};
        \draw[dimline] (0, -0.42) -- (3.2, -0.42);
        \node[dimfont, below] at (1.6, -0.44) {$1\,\mathrm{mm}$};
        \draw[dimline] (-0.72, 0) -- (-0.72, 3.2);
        \node[dimfont, rotate=90, above] at (-0.76, 1.6) {$1\,\mathrm{mm}$};
    \end{tikzpicture}
    \caption{Plate with a circular hole. The left edge is fixed, the
    right edge is pulled horizontally, and the hole is a masked region
    with no boundary discretization.}
    \label{fig:setup-hole}
\end{figure}
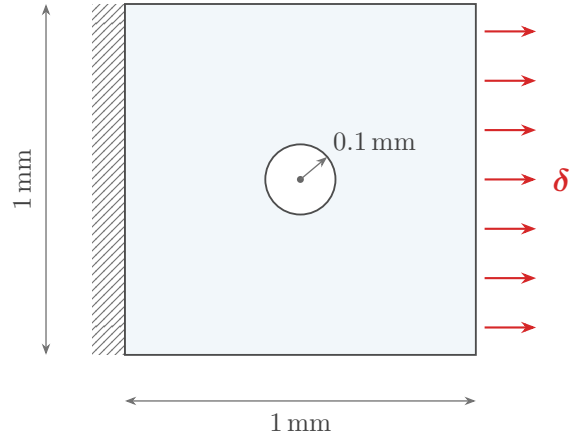

The specimen is the unit square with a circular hole of radius
$0.1\,\mathrm{mm}$ at its center, loaded in uniaxial tension along
the horizontal axis through a boundary lift of the
form of Eq.~\eqref{eq:lift} and shown in Fig.~\ref{fig:setup-hole}. The
hole enters the estimator of Eq.~\eqref{eq:mc} as the indicator mask
$\chi$, with no boundary fitting of any kind. The circular hole admits an exact
boundary-fitted NURBS description, and the plate with a hole is among
the original examples of isogeometric analysis~\cite{hughes2005}, so
the geometry map of Section~\ref{subsec:iga} could represent it with
no mask at all. The runs keep the identity map nonetheless, since the
masked representation is the ingredient under test here; the geometry
map itself is exercised by the ring of Section~\ref{subsec:ring}.

Since there is no pre-existing crack, the seeded profile vanishes,
$\pf_{0} = 0$, and the phase-field representation of Eq.~\eqref{eq:phi-ansatz}
reduces to the squashed network output alone. Nucleation is therefore
carried entirely by the energy competition, which is a recognized
strength of the diffuse description~\cite{tanne2018}.

\begin{figure}[!htb]
    \centering
    \begin{subfigure}[t]{0.48\linewidth}
        \centering
        \begin{tikzpicture}
            \begin{axis}[
                paperaxis,
                width=\linewidth, height=5.4cm,
                xlabel={$\theta$ (deg)},
                ylabel={$\sigma_{\theta\theta}/\sigma_{\infty}$},
                xmin=0, xmax=360, ymin=-1.4, ymax=3.6,
                xtick={0, 90, 180, 270, 360},
            ]
                \addplot[black, thick]
                    table[col sep=comma, x=theta_deg, y=kirsch]
                    {figures/data/kirsch_angular.csv};
                \addplot[cred, only marks, mark=*, mark size=1.1pt,
                         each nth point=4]
                    table[col sep=comma, x=theta_deg, y=dem]
                    {figures/data/kirsch_angular.csv};
            \end{axis}
        \end{tikzpicture}
        \caption{}
        \label{fig:hole-ang}
    \end{subfigure}%
    \hfill%
    \begin{subfigure}[t]{0.48\linewidth}
        \centering
        \begin{tikzpicture}
            \begin{axis}[
                paperaxis,
                width=\linewidth, height=5.4cm,
                xlabel={$r/a$},
                ylabel={$\sigma_{\theta\theta}/\sigma_{\infty}$},
                xmin=1, xmax=3, ymin=-1.4, ymax=3.8,
            ]
                \addplot[black, thick]
                    table[col sep=comma, x=r_over_a, y=kirsch90]
                    {figures/data/kirsch_radial.csv};
                \addplot[cred, only marks, mark=*, mark size=1.1pt,
                         each nth point=3]
                    table[col sep=comma, x=r_over_a, y=dem90]
                    {figures/data/kirsch_radial.csv};
                \addplot[black, dashed, thick]
                    table[col sep=comma, x=r_over_a, y=kirsch0]
                    {figures/data/kirsch_radial.csv};
                \addplot[cblue, only marks, mark=square*, mark size=1.0pt,
                         each nth point=3]
                    table[col sep=comma, x=r_over_a, y=dem0]
                    {figures/data/kirsch_radial.csv};
                \node[font=\footnotesize, anchor=west]
                    at (axis cs:2.30, 1.62) {$\theta = 90^{\circ}$};
                \node[font=\footnotesize, anchor=west]
                    at (axis cs:2.30, -0.62) {$\theta = 0^{\circ}$};
            \end{axis}
        \end{tikzpicture}
        \caption{}
        \label{fig:hole-rad}
    \end{subfigure}
    \caption{Plate with a circular hole, elastic stage.
    (a)~Hoop stress on the circle $r = a + 0.1\lreg$, one tenth of a
    regularization length outside the hole of radius $a$, against the
    Kirsch solution~\cite{kirsch1898}; (b)~radial decay along the
    $\theta = 90^{\circ}$ and $\theta = 0^{\circ}$ rays. In both panels
    the lines are the analytic solution and the markers are the
    computed values. The hole
    is represented by the indicator mask in~Eq.~\eqref{eq:mc}, with no
    boundary fitting; stresses are normalized by the measured far-field
    stress. Small discrepancies from the analytic values are expected,
    since the Kirsch solution describes an infinite plate and the
    specimen is finite.}
    \label{fig:hole}
\end{figure}
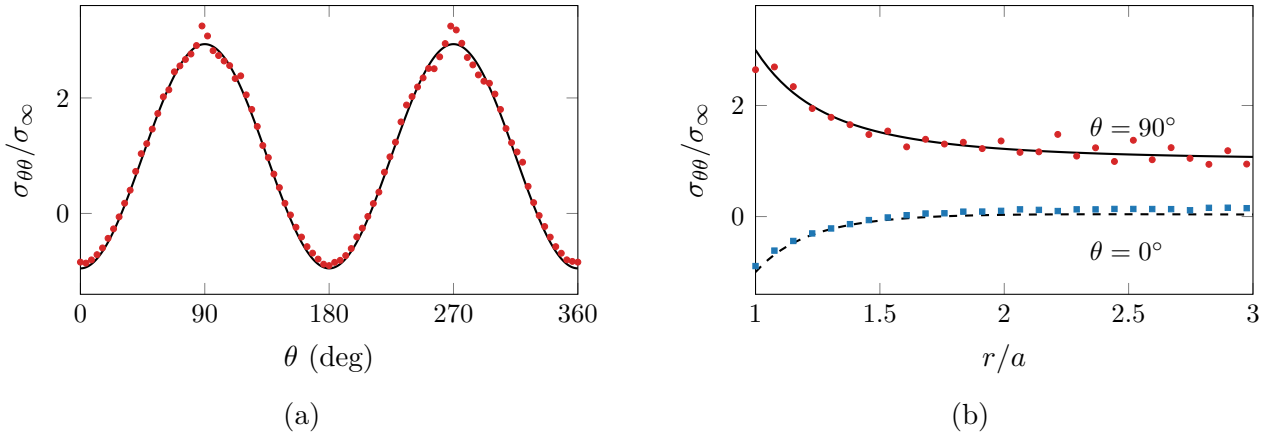

In a first stage the phase field is held at zero, so that
minimizing~Eq.~\eqref{eq:pi} reduces to the elastic problem alone. The Kirsch solution for an infinite plate gives the
hoop stress along the hole edge as
$\sigma_{\theta\theta}(a, \theta) = \sigma_{\infty}(1 - 2\cos 2\theta)$,
with the stress concentration factor of three at
$\theta = \pm 90^{\circ}$, and a characteristic radial decay of
$\sigma_{\theta\theta}$ toward the far field~\cite{kirsch1898}.
Fig.~\ref{fig:hole} compares the computed hoop stress against this
solution, with all stresses obtained by automatic differentiation of
the converged displacement field and normalized by the far-field
stress $\sigma_{\infty} = 342\,\mathrm{MPa}$, measured on the
specimen itself as the mean axial stress over the bulk away from the
hole. The infinite-plate reference applies on the finite specimen
since the disturbance of the hole decays as $(a/r)^{2}$. The outer
boundary lies five radii from the center, where that decay leaves a
disturbance of $4\%$, and the compared fields all lie
within three radii. The angular
profile follows the analytic distribution over the full sweep with a
root-mean-square deviation of $0.09\,\sigma_{\infty}$, about
$3\%$ of the peak value, and the radial decay is reproduced along
both the $\theta = 90^{\circ}$ and the $\theta = 0^{\circ}$ rays. Near
the peaks the computed values slightly exceed the infinite-plate curve,
which is consistent with the finite width of the plate, since the hole
occupies one fifth of the specimen width. Since this stage involves no
fracture, it isolates the ingredients that are new in this example,
namely the mask of Section~\ref{subsec:iga}, the hole-edge sampling
and the quadrature of Section~\ref{subsec:quadrature}.

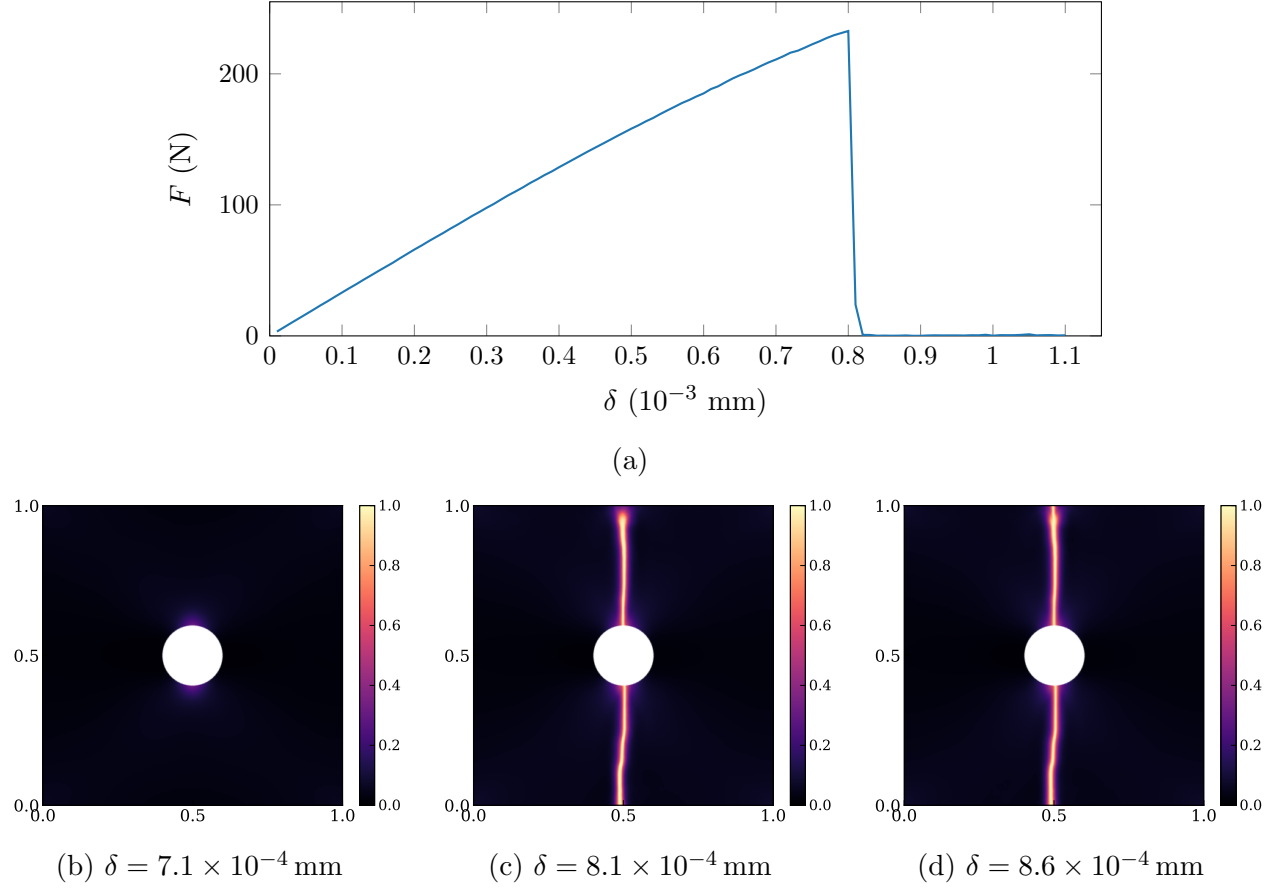
\begin{figure}[!htb]
    \centering
    \begin{subfigure}[b]{0.74\linewidth}
        \centering
        \begin{tikzpicture}
            \begin{axis}[
                paperaxis, width=\linewidth, height=6.0cm,
                xlabel={$\delta$ ($10^{-3}$ mm)}, ylabel={$F$ (N)},
                xmin=0, xmax=1.15, ymin=0, ymax=255,
            ]
                \addplot[cblue, thick]
                    table[col sep=comma,
                          x expr=\thisrow{delta_mm}*1000, y=F]
                    {figures/data/nucl_dem2.csv};
            \end{axis}
        \end{tikzpicture}
        \caption{}
        \label{fig:nucl-fd}
    \end{subfigure}

    \vspace{0.6ex}
    \begin{subfigure}[t]{0.33\linewidth}
        \centering
        \includegraphics[width=\linewidth]{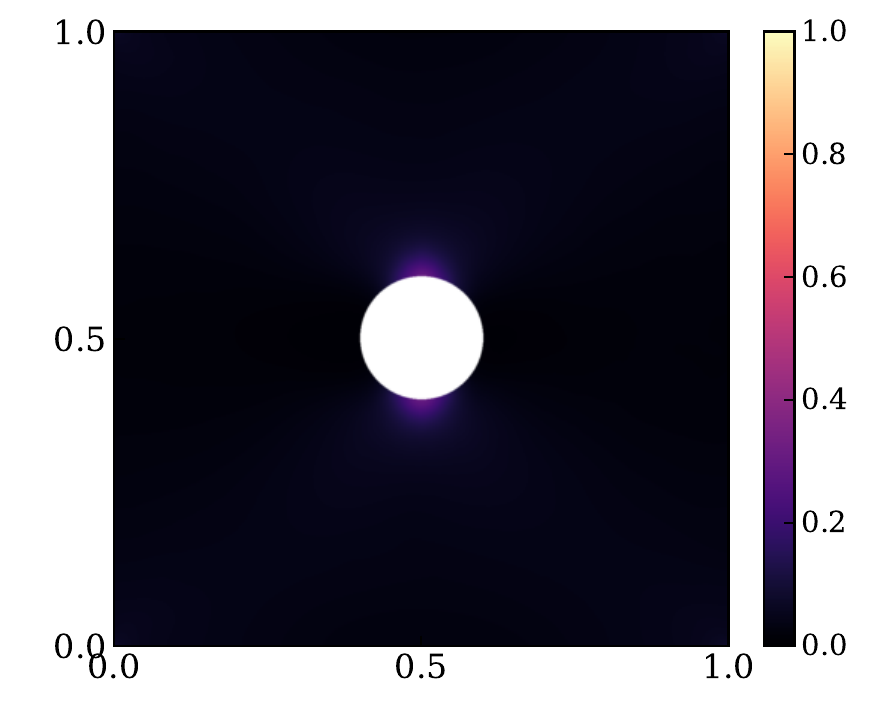}
        \caption{$\delta = 7.1 \times 10^{-4}\,\mathrm{mm}$}
    \end{subfigure}%
    \hfill%
    \begin{subfigure}[t]{0.33\linewidth}
        \centering
        \includegraphics[width=\linewidth]{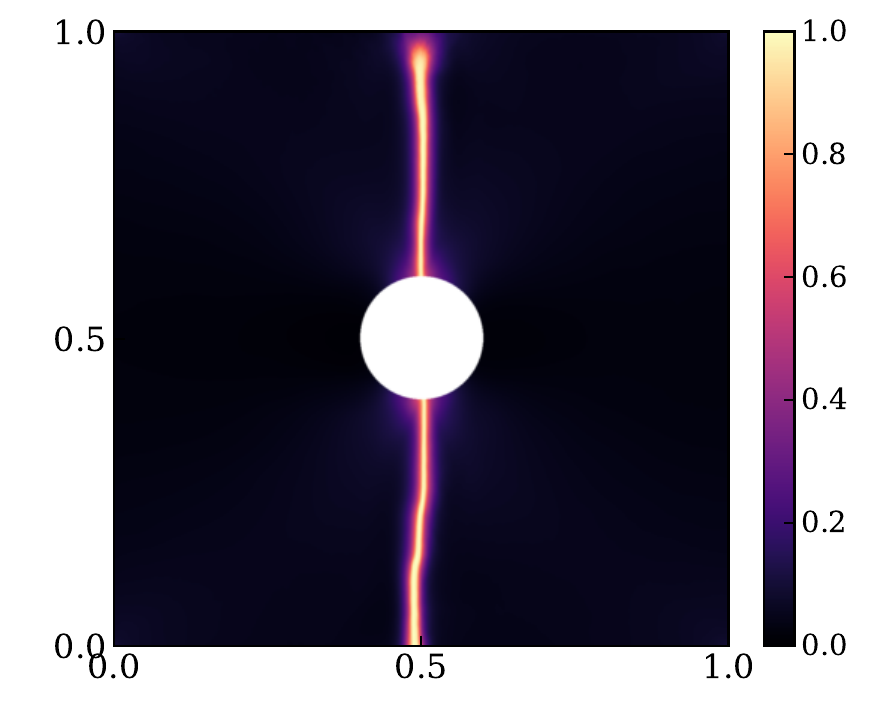}
        \caption{$\delta = 8.1 \times 10^{-4}\,\mathrm{mm}$}
    \end{subfigure}%
    \hfill%
    \begin{subfigure}[t]{0.33\linewidth}
        \centering
        \includegraphics[width=\linewidth]{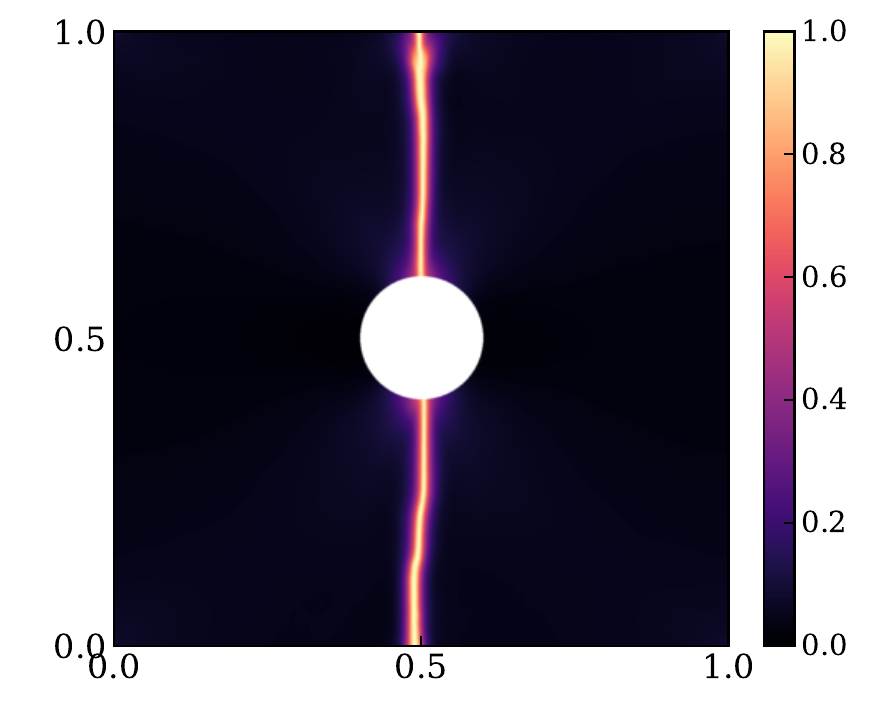}
        \caption{$\delta = 8.6 \times 10^{-4}\,\mathrm{mm}$}
    \end{subfigure}

    \caption{Plate with a circular hole, fracture stage.
    (a)~Load--displacement curve; (b)--(d)~phase field at three load
    levels. Damage nucleates at the stress concentration sites at
    $\theta = \pm 90^{\circ}$, and the two symmetric cracks traverse the
    specimen within a single load increment of the peak.}
    \label{fig:nucleation}
\end{figure}

\begin{figure}[!htb]
    \centering
    \begin{subfigure}[t]{0.33\linewidth}
        \centering
        \includegraphics[width=\linewidth]{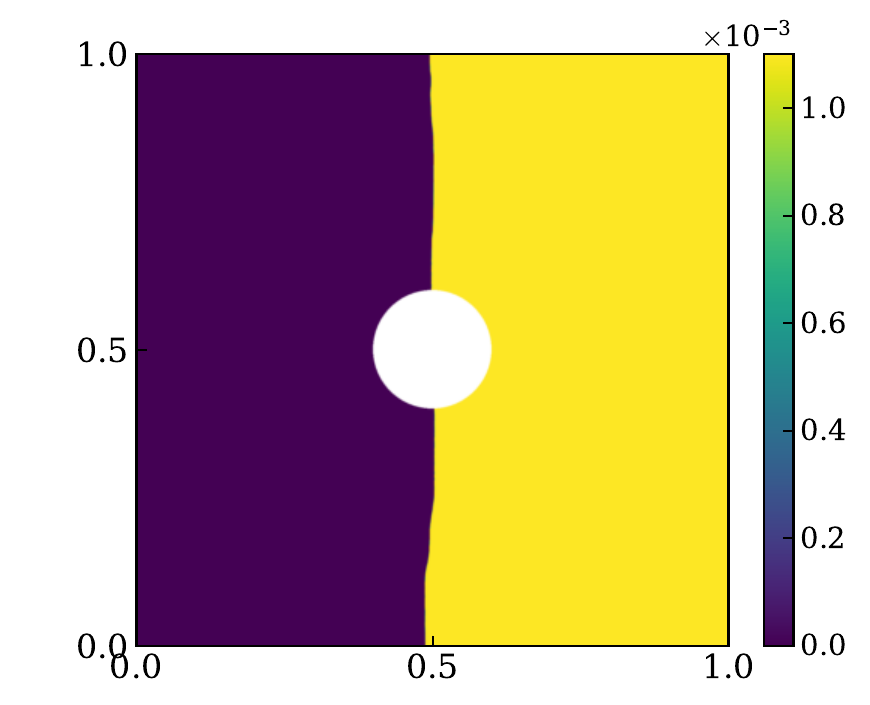}
        \caption{}
    \end{subfigure}%
    \hspace{0.04\linewidth}%
    \begin{subfigure}[t]{0.33\linewidth}
        \centering
        \includegraphics[width=\linewidth]{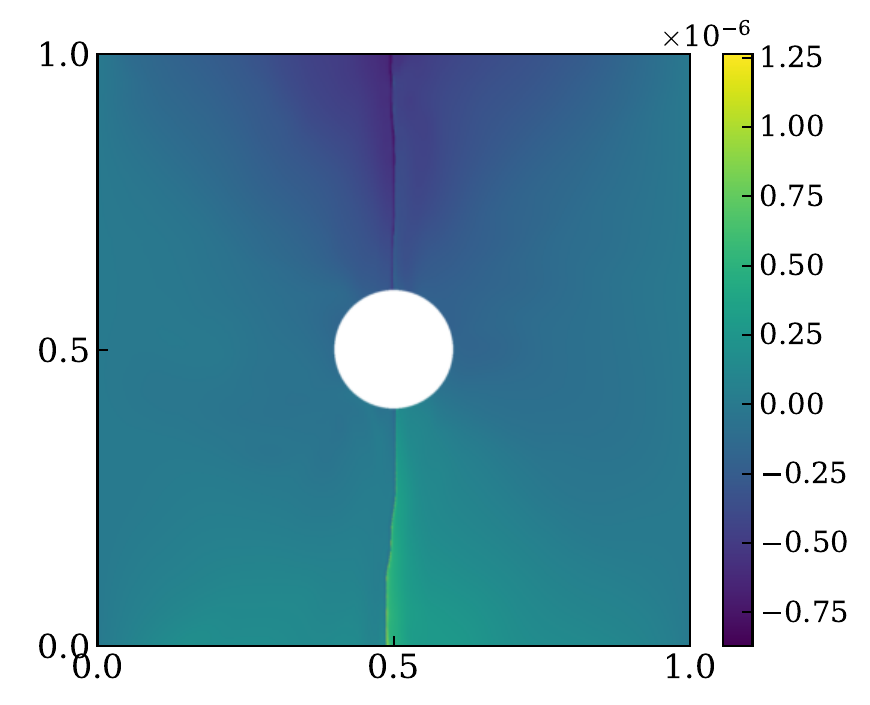}
        \caption{}
    \end{subfigure}
    \caption{Plate with a circular hole, displacement components
    (a)~$u$ and (b)~$v$ of the failed state, in mm.}
    \label{fig:nucleation-u}
\end{figure}

With the elastic stage verified, the full model is released. Damage
nucleates on the hole edge at $\theta = \pm 90^{\circ}$, precisely where
the Kirsch solution concentrates the hoop stress, and grows two
symmetric cracks toward the top and bottom edges. There is no closed-form
reference for this stage; the verification is one of physical
consistency, in that the nucleation site follows the elastic
concentration and the load--displacement curve shows a single sharp
failure event. The site agrees with the variational cavity study of
Tann\'{e} et al.~\cite{tanne2018}, which places nucleation at the
points of maximum hoop stress for holes large on the scale of
$\lreg$; the radius used here is ten $\lreg$. The peak load is $232.7\,\mathrm{N}$ at
$\delta = 8.0 \times 10^{-4}\,\mathrm{mm}$, and the load falls by an
order of magnitude in the single increment that follows, to
$23.8\,\mathrm{N}$, the specimen carrying nothing thereafter
(Fig.~\ref{fig:nucleation}). The displacement components of the
failed state are shown in Fig.~\ref{fig:nucleation-u}, where the
component along the loading direction jumps across the crack pair
while the fields stay smooth elsewhere. A comparable specimen has been treated
within an energy-minimizing network formulation before, where a plate
with an eccentric hole was loaded in tension and a crack was reported
to grow from the hole~\cite{chakraborty2022}. Neither a verification
of the elastic stage against a closed-form solution nor a
quantitative load is reported there, and both are supplied here.

\FloatBarrier
% ---------------------------------------------------------------------
\subsection{Thick-walled ring}
\label{subsec:ring}

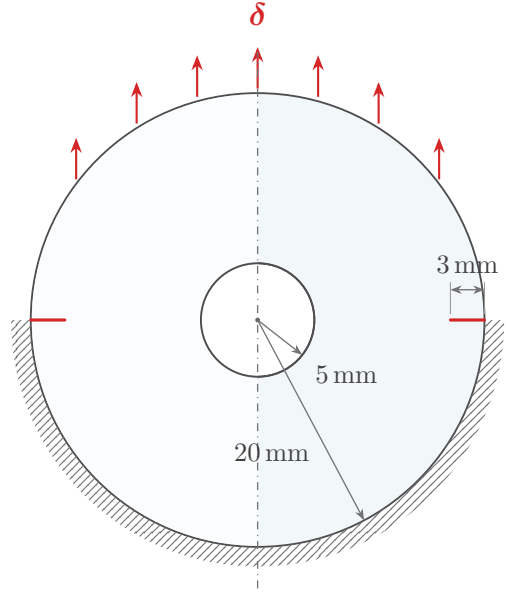
\begin{figure}[!htb]
    \centering
    \begin{tikzpicture}[even odd rule]
        \draw[spec] (0, 0) circle (3.0) (0, 0) circle (0.75);
        \begin{scope}
            \clip (-3.2, -3.2) rectangle (0, 3.2);
            \fill[white, opacity=0.75] (0, 0) circle (2.985);
        \end{scope}
        \draw[black!70, line width=0.7pt] (0, 0) circle (0.75);
        \fill[clamphatch]
            (3.26, 0) arc (0:-180:3.26)
            -- (-3.0, 0) arc (-180:0:3.0) -- cycle;
        \draw[crackline] (2.55, 0) -- (3.0, 0);
        \draw[crackline] (-3.0, 0) -- (-2.55, 0);
        \foreach \x in {-2.4, -1.6, -0.8, 0, 0.8, 1.6, 2.4} {
            \draw[loadarrow] (\x, {sqrt(9.0-\x*\x)+0.06})
                -- (\x, {sqrt(9.0-\x*\x)+0.61});
        }
        \node[cred, font=\normalsize] at (0, 4.05) {$\bm{\delta}$};
        \draw[black!55, dash dot, line width=0.5pt]
            (0, -3.55) -- (0, 3.55);
        % notch length: measured dimension above the right notch
        \draw[black!40, line width=0.4pt] (2.55, 0.08) -- (2.55, 0.50);
        \draw[black!40, line width=0.4pt] (3.0, 0.08) -- (3.0, 0.50);
        \draw[dimline] (2.55, 0.40) -- (3.0, 0.40);
        \node[dimfont, above] at (2.775, 0.48) {$3\,\mathrm{mm}$};
        % radii: measured dimensions from the center
        \fill[black!60] (0, 0) circle (0.9pt);
        \draw[black!55, line width=0.5pt, -{Stealth[length=1.5mm]}]
            (0, 0) -- ({0.75*cos(-38)}, {0.75*sin(-38)});
        \node[dimfont, anchor=west] at (0.62, -0.72) {$5\,\mathrm{mm}$};
        \draw[black!55, line width=0.5pt, -{Stealth[length=1.5mm]}]
            (0, 0) -- ({3.0*cos(-62)}, {3.0*sin(-62)});
        \node[dimfont, anchor=east] at (0.83, -1.75) {$20\,\mathrm{mm}$};
    \end{tikzpicture}
    \caption{Thick-walled ring, with inner radius $5\,\mathrm{mm}$,
    outer radius $20\,\mathrm{mm}$ and two symmetric notches of length
    $3\,\mathrm{mm}$ at the outer boundary. The upper outer semicircle
    is pulled vertically, the lower outer semicircle is fixed, and
    the notches are seeded diffusely. By symmetry about the vertical
    centerline (dash-dotted), one half of the ring is modeled; the
    computed right half is drawn at full tint, and the field
    renderings of Fig.~\ref{fig:ring} show that half.
   }
    \label{fig:setup-ring}
\end{figure}

The thick-walled ring involves a curved domain. Following Si et
al.~\cite{si2023}, a ring with inner radius
$5\,\mathrm{mm}$ and outer radius $20\,\mathrm{mm}$ carries two
symmetric $3\,\mathrm{mm}$ notches on its outer boundary; the upper
outer semicircle is pulled vertically, the lower outer semicircle is
fixed, and the cracks grow from the notch tips horizontally toward
the center. By the symmetry of the set-up, one half of the ring is
modeled, with the symmetry conditions imposed on the vertical
centerline (Fig.~\ref{fig:setup-ring}).

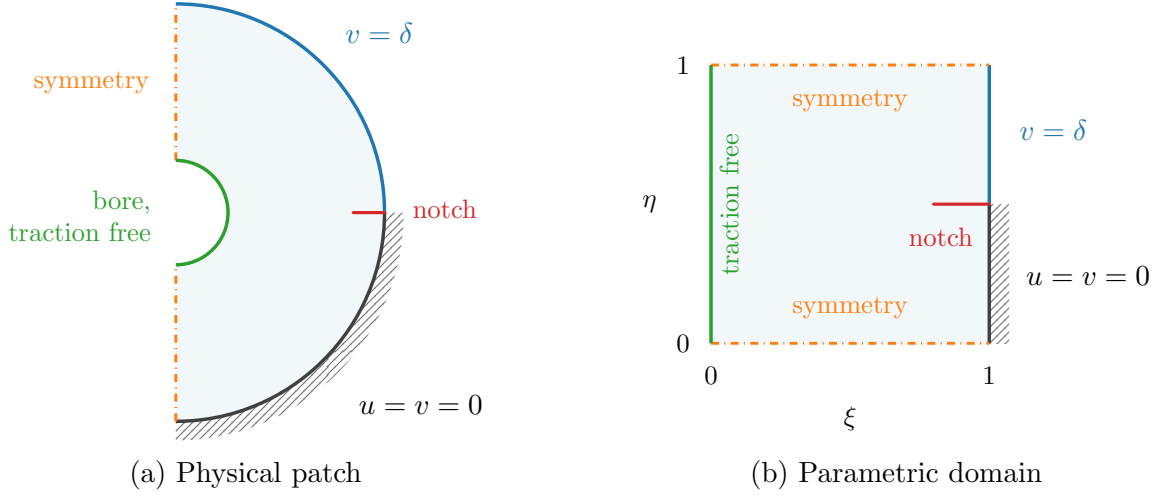
\begin{figure}[!htb]
    \centering
    \begin{subfigure}[b]{0.46\linewidth}
        \centering
        \begin{tikzpicture}[scale=1.15]
            % modeled half-annulus (right half); each boundary is stroked
            % below in the color of its own condition, so the region is
            % filled only
            \fill[cblue!6] (0, -2.4) arc (-90:90:2.4)
                -- (0, 0.6) arc (90:-90:0.6) -- cycle;
            % fixed lower quarter of the outer arc
            \fill[clamphatch] (2.62, 0) arc (0:-90:2.62)
                -- (0, -2.4) arc (-90:0:2.4) -- cycle;
            \draw[black!75, line width=1.3pt] (2.4, 0) arc (0:-90:2.4);
            % mirrors the v = delta label about the horizontal axis, clearing
            % the hatch band that reaches to 2.62
            \node[font=\small, anchor=north west]
                at ({2.80*cos(-45)}, {2.80*sin(-45)}) {$u = v = 0$};
            % pulled upper quarter of the outer arc
            \draw[cblue, line width=1.3pt] (2.4, 0) arc (0:90:2.4);
            \node[cblue, font=\small, anchor=south west]
                at ({2.58*cos(45)}, {2.58*sin(45)}) {$v = \delta$};
            % symmetry: the two straight edges
            \draw[corange, dash dot, line width=1.1pt] (0, 0.6) -- (0, 2.4);
            \draw[corange, dash dot, line width=1.1pt]
                (0, -2.4) -- (0, -0.6);
            \node[corange, font=\footnotesize, anchor=east, align=right]
                at (-0.18, 1.50) {symmetry};
            % traction-free bore, labelled in the empty mirrored half
            \draw[cgreen, line width=1.3pt] (0, 0.6) arc (90:-90:0.6);
            \node[cgreen, font=\footnotesize, anchor=east, align=right]
                at (-0.16, -0.04) {bore,\\traction free};
            % notch
            \draw[crackline] (2.04, 0) -- (2.4, 0);
            \node[cred, font=\footnotesize, anchor=west] at (2.60, 0.04)
                {notch};
        \end{tikzpicture}
        \caption{Physical patch}
        \label{fig:ring-bc-phys}
    \end{subfigure}
    \hspace{0.03\linewidth}
    \begin{subfigure}[b]{0.46\linewidth}
        \centering
        \begin{tikzpicture}[scale=1.15]
            \fill[cblue!6] (0, 0) rectangle (3.2, 3.2);
            % xi = 1: fixed below, pulled above
            \fill[clamphatch] (3.2, 0) rectangle (3.42, 1.6);
            \draw[black!75, line width=1.3pt] (3.2, 0) -- (3.2, 1.6);
            \draw[cblue, line width=1.3pt] (3.2, 1.6) -- (3.2, 3.2);
            \node[cblue, font=\small, anchor=west] at (3.42, 2.45)
                {$v = \delta$};
            \node[font=\small, anchor=west] at (3.50, 0.78)
                {$u = v = 0$};
            % eta = 0 and eta = 1: symmetry
            \draw[corange, dash dot, line width=1.1pt] (0, 0) -- (3.2, 0);
            \draw[corange, dash dot, line width=1.1pt]
                (0, 3.2) -- (3.2, 3.2);
            \node[corange, font=\footnotesize, anchor=north]
                at (1.60, 3.04) {symmetry};
            \node[corange, font=\footnotesize, anchor=south]
                at (1.60, 0.16) {symmetry};
            % xi = 0: traction-free bore
            \draw[cgreen, line width=1.3pt] (0, 0) -- (0, 3.2);
            \node[cgreen, font=\footnotesize, rotate=90, anchor=south]
                at (0.44, 1.6) {traction free};
            % notch
            \draw[crackline] (2.56, 1.6) -- (3.2, 1.6);
            \node[cred, font=\footnotesize, anchor=north east]
                at (3.14, 1.46) {notch};
            % parametric axes
            \node[font=\footnotesize, anchor=north] at (0, -0.14) {$0$};
            \node[font=\footnotesize, anchor=north] at (3.2, -0.14) {$1$};
            \node[font=\footnotesize, anchor=north] at (1.6, -0.60)
                {$\xi$};
            \node[font=\footnotesize, anchor=east] at (-0.12, 0) {$0$};
            \node[font=\footnotesize, anchor=east] at (-0.12, 3.2)
                {$1$};
            \node[font=\footnotesize, anchor=east] at (-0.50, 1.6)
                {$\eta$};
        \end{tikzpicture}
        \caption{Parametric domain}
        \label{fig:ring-bc-param}
    \end{subfigure}
    \caption{Boundary conditions of the ring on the single NURBS patch.
    (a)~The modeled half in physical space; (b)~the parametric domain
    it is mapped from, drawn with the same colors and conventions.
    Each boundary of the specimen is one edge of the
    patch, the symmetry line being the pair of edges $\eta = 0$ and
    $\eta = 1$, the traction-free bore $\xi = 0$, and the outer arc
    $\xi = 1$, pulled above the notch and fixed below it. The two
    arc conditions meet at the interior knot $\eta = \tfrac{1}{2}$,
    which the map sends to the notched point of the outer boundary.}
    \label{fig:ring-bc}
\end{figure}

\begin{figure}[!htb]
    \centering
    \begin{subfigure}[t]{0.33\linewidth}
        \centering
        \includegraphics[width=\linewidth]{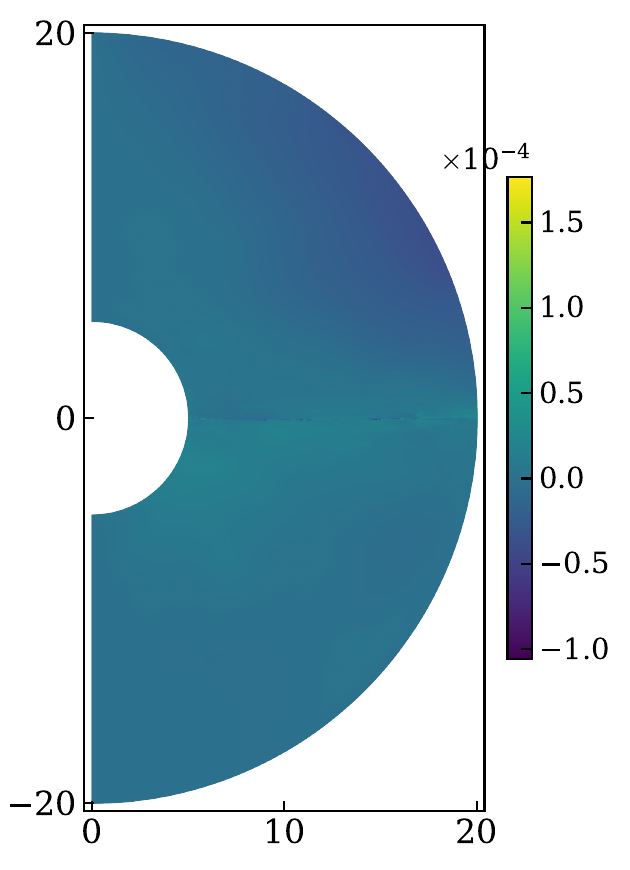}
        \caption{}
    \end{subfigure}%
    \hfill%
    \begin{subfigure}[t]{0.33\linewidth}
        \centering
        \includegraphics[width=\linewidth]{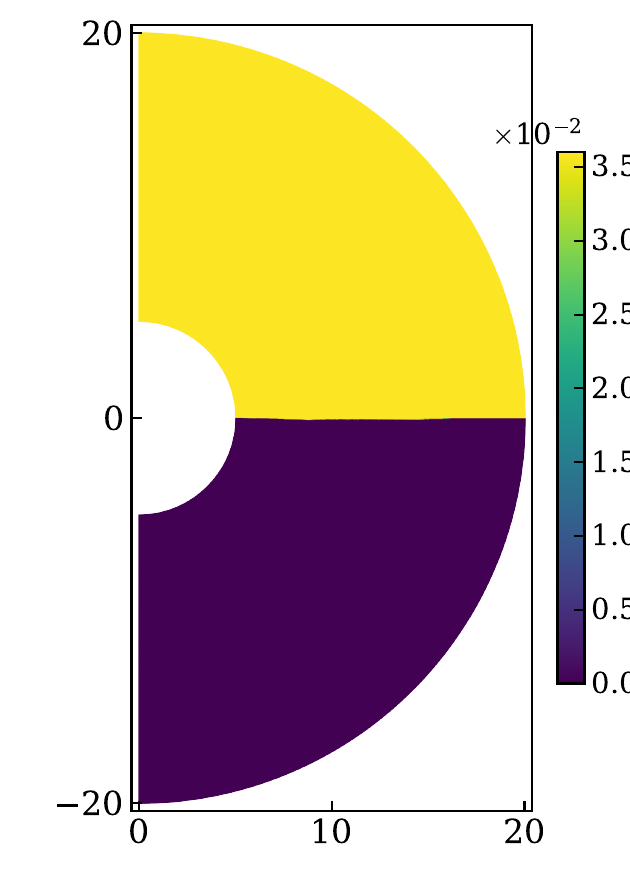}
        \caption{}
    \end{subfigure}%
    \hfill%
    \begin{subfigure}[t]{0.33\linewidth}
        \centering
        \includegraphics[width=\linewidth]{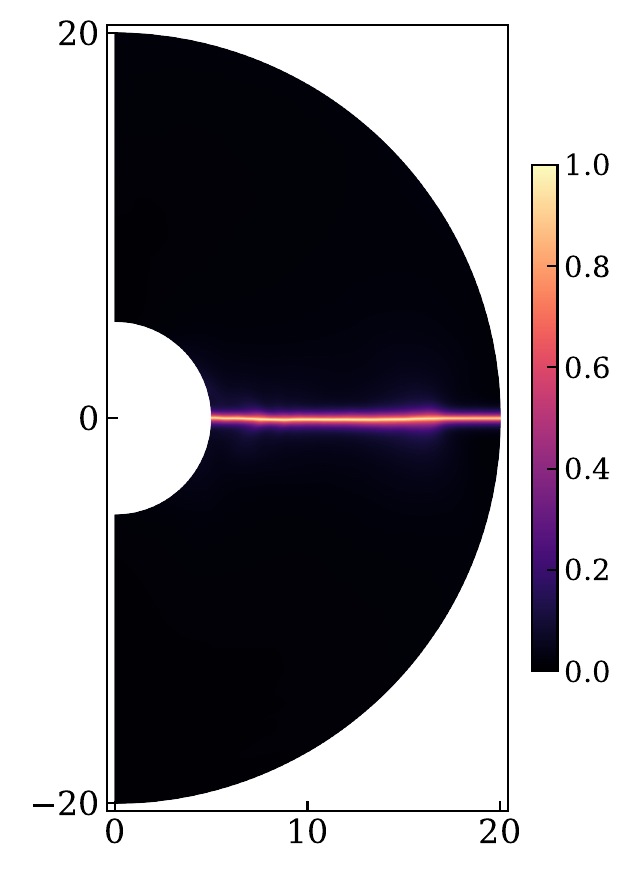}
        \caption{}
    \end{subfigure}
    \caption{Thick-walled ring on a single NURBS patch, final state on
    the computed half. (a),(b)~Displacement components $u$ and $v$ in
    mm; (c)~phase field. The crack grows from the notch tip
    horizontally toward the center and severs the full wall down to
    the inner bore, in agreement with the path reported
    in~\cite{si2023}.}
    \label{fig:ring}
\end{figure}

 The domain is represented by a
single quadratic NURBS patch of the form of Eq.~\eqref{eq:geomap} whose
weights reproduce both circles exactly, the solution fields and the
encoding live on the parametric square with strains obtained through
the pullback of Eq.~\eqref{eq:pullback}, and the sampler operates in physical
measure through the $\detJ$ weight of~Eq.~\eqref{eq:mc}, all as described
in Section~\ref{subsec:iga}; the patch, its control net and its basis
functions are the ones shown in Fig.~\ref{fig:iga}. Every boundary of
the specimen is one edge of that patch, and each carries its condition
there (Fig.~\ref{fig:ring-bc}). The two straight edges on the symmetry
line are the images of $\eta = 0$ and $\eta = 1$, so the symmetry
condition $u = 0$ holds exactly through the envelope
of~Eq.~\eqref{eq:lift} with $v$ left free, and the inner bore is the image
of $\xi = 0$ and is traction free. The outer arc is the image of
$\xi = 1$ and carries the prescribed vertical displacement above the
notch and the fixed condition below it. This arc is the one boundary in this
work that is not treated by the exact construction of
Section~\ref{subsec:fields}. That construction multiplies the network
output by an envelope vanishing on the whole Dirichlet boundary and
adds a lift carrying the data, which presumes one condition along the
edge; here the pulled and the fixed part meet at the interior knot
$\eta = \tfrac{1}{2}$ of Fig.~\ref{fig:iga}c, which the map sends to
the notched point of the outer boundary, and a lift interpolating both
would have to jump there. The two conditions are imposed instead
through a quadratic boundary term on $\xi = 1$ of fixed stiffness
$10^{6}$ in the nondimensional units of Section~\ref{subsec:fields},
integrated in physical arc length by the sampling scheme of
Section~\ref{subsec:quadrature}. The term is admissible in a method
that minimizes the energy since it is non-negative, so the total
energy stays bounded below and the incremental problem keeps the
minimization structure of~Eq.~\eqref{eq:incremental}, with the arc
displacement approaching the prescribed data as the stiffness grows;
a boundary term able to lower the energy would remove that lower bound.
The prescribed displacement enters the energy through this term alone,
so the reaction force is again the derivative of the
minimized energy with respect to $\delta$, as in Eq.~\eqref{eq:force}. No boundary-fitted
mesh, no multi-patch coupling and no interface terms are involved, which
may be contrasted with the multi-patch adaptive scheme of Si et
al.~\cite{si2023}, where the same geometry is assembled from four
quarter-ring patches joined by Nitsche interface terms. The run uses the second-order model with
the material data of the reference, $E = 210\,\mathrm{GPa}$,
$\nu = 0.3$ and $\Gc = 2.7\,\mathrm{kJ/m^{2}}$, and a regularization
length $\lreg = 0.2\,\mathrm{mm}$, with the level resolutions scaled to
it as in Section~\ref{subsec:encoding}. The crack grows from the notch
tip horizontally toward the center, reproducing the path reported
in~\cite{si2023}, and severs the full ligament down to the inner bore
(Fig.~\ref{fig:ring}c). The computed reaction force rises to
$3.91\,\mathrm{kN}$ at $\delta = 2.67 \times 10^{-2}\,\mathrm{mm}$ and
then softens over the remaining loading as the crack crosses the wall.

The displacement components behave as the set-up prescribes. The
pulled upper semicircle carries the prescribed vertical displacement,
the fixed lower semicircle stays at rest, and both components jump
across the severed ligament (Fig.~\ref{fig:ring}a,b).

\FloatBarrier
% ---------------------------------------------------------------------
\subsection{Random multi-crack configurations}
\label{subsec:benchmark}

The examples so far are the standard tests of the phase-field
literature. Hamdi and Lejeune have recently argued that these tests
no longer separate solvers, and have published a benchmark dataset
constructed to be harder~\cite{hamdi2026}. The dataset
comprises one thousand random configurations of $10$--$20$ interior
cracks in a $2 \times 2\,\mathrm{mm}$ plate, each solved under
biaxial tension and under shear by a staggered finite element scheme
on an $800 \times 800$ mesh ($h = 0.25\,\lreg$), with one hundred
snapshots of the fields and the reaction force stored per run. The
dataset provides each configuration under three choices of the
driving energy, and the comparisons here use the spectral one. Its
relevance here is twofold. First, the physics matches
Section~\ref{subsec:splits} exactly. Their displacement solve
degrades the full isotropic stress and drives the damage with the
spectral tensile energy, which is the hybrid formulation adopted
here, and the runs of this section take the second-order fracture
energy density and the material of their study,
$E = 10^{6}\,\mathrm{N/mm^{2}}$, $\nu = 0.3$,
$\Gc = 1\,\mathrm{N/mm}$ and
$\lreg = 0.01\,\mathrm{mm}$. Second, their deep Ritz baseline, which
runs the architecture and code of~\cite{manav2024}, failed on these
configurations, producing a different crack pattern from every
network initialization~\cite{hamdi2026}. Running the proposed method on their
data is therefore a stringent external test.

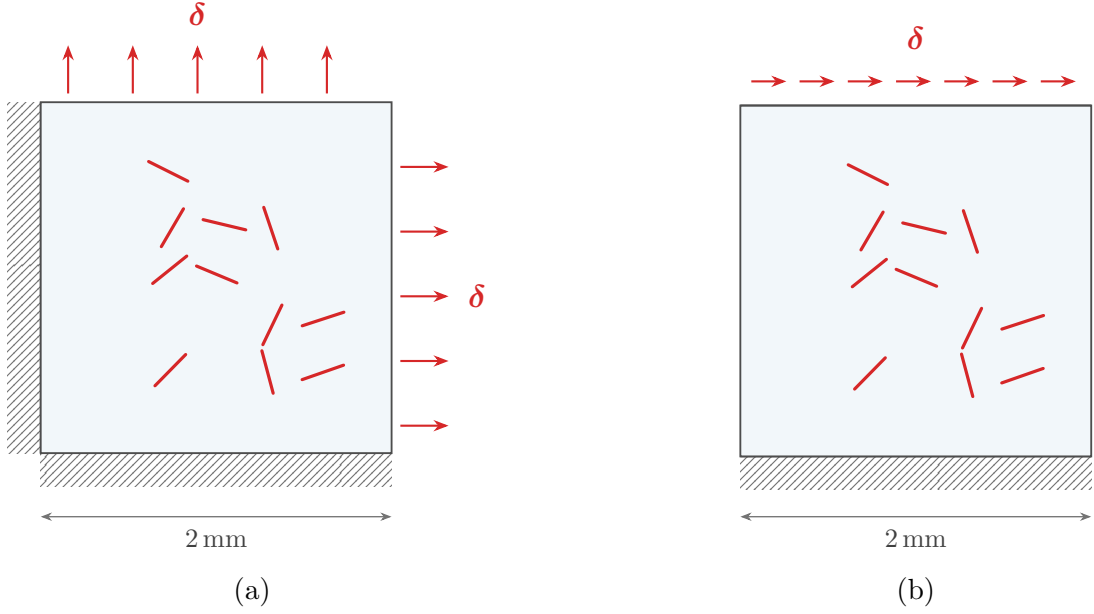
\begin{figure}[!htb]
    \centering
    \begin{subfigure}[b]{0.44\linewidth}
        \centering
        \begin{tikzpicture}[scale=1.45]
            \fill[clamphatch] (0, -0.30) rectangle (3.2, 0);
            \draw[black!70, line width=0.7pt] (0, 0) -- (3.2, 0);
            \fill[clamphatch] (-0.30, 0) rectangle (0, 3.2);
            \draw[black!70, line width=0.7pt] (0, 0) -- (0, 3.2);
            \draw[spec] (0, 0) rectangle (3.2, 3.2);
            \draw[crackline] (1.342, 2.480) -- (0.984, 2.659);
            \draw[crackline] (1.021, 1.547) -- (1.332, 1.797);
            \draw[crackline] (2.384, 0.670) -- (2.762, 0.802);
            \draw[crackline] (2.120, 0.546) -- (2.017, 0.933);
            \draw[crackline] (1.790, 1.552) -- (1.421, 1.706);
            \draw[crackline] (1.870, 2.036) -- (1.481, 2.129);
            \draw[crackline] (2.026, 0.989) -- (2.200, 1.350);
            \draw[crackline] (1.100, 1.883) -- (1.302, 2.228);
            \draw[crackline] (2.162, 1.862) -- (2.034, 2.241);
            \draw[crackline] (1.042, 0.614) -- (1.322, 0.899);
            \draw[crackline] (2.385, 1.161) -- (2.765, 1.285);
            \foreach \x in {0.25, 0.84, 1.43, 2.02, 2.61} {
                \draw[loadarrow] (\x, 3.28) -- (\x, 3.72);
            }
            \foreach \y in {0.25, 0.84, 1.43, 2.02, 2.61} {
                \draw[loadarrow] (3.28, \y) -- (3.72, \y);
            }
            \node[cred, font=\normalsize, anchor=south] at (1.43, 3.80)
                {$\bm{\delta}$};
            \node[cred, font=\normalsize, anchor=west] at (3.80, 1.43)
                {$\bm{\delta}$};
            \draw[dimline] (0, -0.58) -- (3.2, -0.58);
            \node[dimfont, below] at (1.6, -0.60) {$2\,\mathrm{mm}$};
        \end{tikzpicture}
        \caption{}
        \label{fig:setup-bench-t}
    \end{subfigure}
    \hspace{0.06\linewidth}
    \begin{subfigure}[b]{0.44\linewidth}
        \centering
        \begin{tikzpicture}[scale=1.45]
            \fill[clamphatch] (0, -0.30) rectangle (3.2, 0);
            \draw[black!70, line width=0.7pt] (0, 0) -- (3.2, 0);
            \draw[spec] (0, 0) rectangle (3.2, 3.2);
            \draw[crackline] (1.342, 2.480) -- (0.984, 2.659);
            \draw[crackline] (1.021, 1.547) -- (1.332, 1.797);
            \draw[crackline] (2.384, 0.670) -- (2.762, 0.802);
            \draw[crackline] (2.120, 0.546) -- (2.017, 0.933);
            \draw[crackline] (1.790, 1.552) -- (1.421, 1.706);
            \draw[crackline] (1.870, 2.036) -- (1.481, 2.129);
            \draw[crackline] (2.026, 0.989) -- (2.200, 1.350);
            \draw[crackline] (1.100, 1.883) -- (1.302, 2.228);
            \draw[crackline] (2.162, 1.862) -- (2.034, 2.241);
            \draw[crackline] (1.042, 0.614) -- (1.322, 0.899);
            \draw[crackline] (2.385, 1.161) -- (2.765, 1.285);
            \draw[black!70, line width=0.7pt] (0, 3.2) -- (3.2, 3.2);
            \foreach \x in {0.10, 0.54, 0.98, 1.42, 1.86, 2.30, 2.74} {
                \draw[loadarrow] (\x, 3.42) -- (\x + 0.32, 3.42);
            }
            \node[cred, font=\normalsize, anchor=south] at (1.6, 3.62)
                {$\bm{\delta}$};
            \draw[dimline] (0, -0.55) -- (3.2, -0.55);
            \node[dimfont, below] at (1.6, -0.57) {$2\,\mathrm{mm}$};
        \end{tikzpicture}
        \caption{}
        \label{fig:setup-bench-s}
    \end{subfigure}
    \caption{Benchmark configuration $106244$ ($11$ seeded cracks)
    under (a)~biaxial tension, where the left and bottom edges are
    supported by rollers ($u = 0$ and $v = 0$ respectively) and the
    right and top edges are pulled by the same $\delta$, and
    (b)~shear, where the bottom edge is fixed, the top edge slides
    horizontally, and both constrained edges carry the $\pf = 0$
    condition of the dataset.}
    \label{fig:setup-bench}
\end{figure}

Ten configurations are evaluated
under both loading cases, giving twenty runs with identical solver
settings, no training data and no per-sample adjustment. They are the
first ten of the one thousand the repository publishes, in the order
in which it lists them, rather than a selection made here, and they
carry between $11$ and $17$ cracks. The
seeded cracks enter through the diffuse initial profile of
Section~\ref{subsec:setup}, its width calibrated once against the
history-field initialization of the dataset. The phase fields agree
at the first stored snapshot, reaching Dice $0.96$--$0.98$ on the
pixel grid of the dataset under its own threshold-$0.5$ metric, so
the seeding is faithful and the differences that develop later come
from the computed evolution. In the shear case the dataset imposes
$\pf = 0$ on the two constrained edges to suppress boundary
fracture; the same condition is imposed here exactly, through a
multiplicative envelope on the learnable part of the phase-field
representation, in the spirit of the lift of Eq.~\eqref{eq:lift}. The loading
schedule places one increment on every stored snapshot, one hundred
increments of $5 \times 10^{-5}\,\mathrm{mm}$ on each pulled edge in
tension and of $10^{-4}\,\mathrm{mm}$ in shear. The discretization
uses three feature levels $48/192/768$ with $M = 160000$ points and
$3000$ iterations per step. The shear runs widen the
irreversibility dead band of Section~\ref{subsec:energy} to
$\tau = 0.02$, which keeps the integration noise accumulated over the
long elastic stage of that case from ratcheting the phase field
upward.
Fig.~\ref{fig:setup-bench} shows the crack layout and the boundary
conditions of configuration $106244$.

Two of the dataset's own measures are reported for every run, namely
the Dice coefficient of the final phase field against the reference
on their pixel grid, and a per-crack classification in which a seeded
crack counts as active if damage extends beyond its seeded band,
evaluated by the same rule on both fields. The classification is the
discriminating measure, since in every configuration the reference
leaves several seeded cracks dormant while their neighbors grow, link
and percolate. The trained surrogates of the dataset are not evaluated
on this distinction, and the count is insensitive to the
crack-number bias the dataset authors themselves report for the Dice
score~\cite{hamdi2026}. Table~\ref{tab:benchmark} lists all
twenty runs. The computed fields classify $252$ of $280$ seeded
cracks correctly ($90.0\%$, where marking every crack active would
attain $67\%$), with
ten of the twenty runs classifying every crack correctly, and reach
mean Dice scores of
$0.739$ in tension and $0.824$ in shear. For calibration, the
surrogates trained by the dataset authors on $800$ solved samples
reach Dice $0.680$ (UNet) and $0.733$ (Fourier neural operator with
ensembling)~\cite{hamdi2026}; the numbers reported here are obtained
without seeing a single solved sample.

\begin{table}[!htb]
    \centering
    \caption{Zero-shot results on all twenty benchmark runs. Dice is
    computed at the final load step against the reference field with
    the dataset's threshold-$0.5$ metric; the classification column
    counts seeded cracks whose active/dormant state matches the
    reference.}
    \label{tab:benchmark}
    \footnotesize
    \begin{tabular}{lccccc}
        \toprule
        & & \multicolumn{2}{c}{tension} & \multicolumn{2}{c}{shear} \\
        \cmidrule(lr){3-4} \cmidrule(lr){5-6}
        configuration & cracks & Dice & correct & Dice & correct \\
        \midrule
        $100192$  & $16$ & $0.839$ & $16/16$ & $0.888$ & $16/16$ \\
        $100651$  & $16$ & $0.855$ & $15/16$ & $0.838$ & $15/16$ \\
        $101617$  & $13$ & $0.760$ & $13/13$ & $0.817$ & $13/13$ \\
        $103891$  & $13$ & $0.619$ & $8/13$  & $0.829$ & $13/13$ \\
        $105657$  & $14$ & $0.667$ & $11/14$ & $0.756$ & $7/14$  \\
        $106004$  & $17$ & $0.733$ & $16/17$ & $0.885$ & $16/17$ \\
        $106244$  & $11$ & $0.698$ & $11/11$ & $0.796$ & $11/11$ \\
        $106679$  & $12$ & $0.712$ & $9/12$  & $0.741$ & $8/12$  \\
        $1005939$ & $15$ & $0.744$ & $15/15$ & $0.826$ & $13/15$ \\
        $1077563$ & $13$ & $0.767$ & $13/13$ & $0.867$ & $13/13$ \\
        \midrule
        mean / total & & $0.739$ & $127/140$ & $0.824$ & $125/140$ \\
        \bottomrule
    \end{tabular}
\end{table}

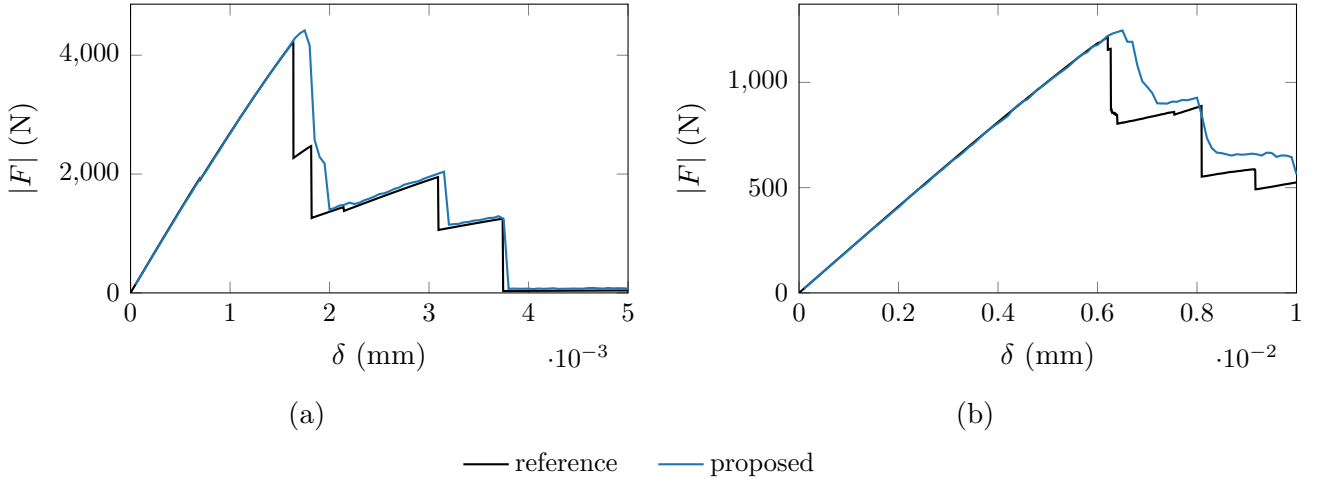
\begin{figure}[!htb]
    \centering
    \begin{subfigure}[t]{0.48\linewidth}
        \centering
        \begin{tikzpicture}
            \begin{axis}[
                paperaxis,
                width=\linewidth, height=5.4cm,
                xlabel={$\delta$ (mm)},
                ylabel={$|F|$ (N)},
                xmin=0, xmax=5e-3, ymin=0,
                legend to name=leg:benchfd,
                legend columns=2,
                legend style={/tikz/every even column/.append style=
                                  {column sep=0.45cm}},
            ]
                \addplot[black, thick]
                    table[col sep=comma, x=delta_mm, y=F]
                    {figures/data/bench_ref_tension.csv};
                \addlegendentry{reference}
                \addplot[cblue, thick]
                    table[col sep=comma, x=delta_mm, y=F]
                    {figures/data/bench_dem_tension.csv};
                \addlegendentry{proposed}
            \end{axis}
        \end{tikzpicture}
        \caption{}
    \end{subfigure}%
    \hfill%
    \begin{subfigure}[t]{0.48\linewidth}
        \centering
        \begin{tikzpicture}
            \begin{axis}[
                paperaxis,
                width=\linewidth, height=5.4cm,
                xlabel={$\delta$ (mm)},
                ylabel={$|F|$ (N)},
                xmin=0, xmax=1e-2, ymin=0,
            ]
                \addplot[black, thick]
                    table[col sep=comma, x=delta_mm, y=F]
                    {figures/data/bench_ref_shear.csv};
                \addplot[cblue, thick]
                    table[col sep=comma, x=delta_mm, y=F]
                    {figures/data/bench_dem_shear.csv};
            \end{axis}
        \end{tikzpicture}
        \caption{}
    \end{subfigure}

    \vspace{0.4ex}
    \ref*{leg:benchfd}
    \caption{Load--displacement response of configuration $106244$
    under (a)~biaxial tension, where the plotted force
    $|F_{x}| + |F_{y}|$ is the work conjugate of the common
    $\delta$, and (b)~shear, against the finite element reference of
    the dataset.}
    \label{fig:bench-fd}
\end{figure}

\begin{figure}[!htb]
    \centering
    \begin{subfigure}[t]{0.33\linewidth}
        \centering
        \includegraphics[width=\linewidth]{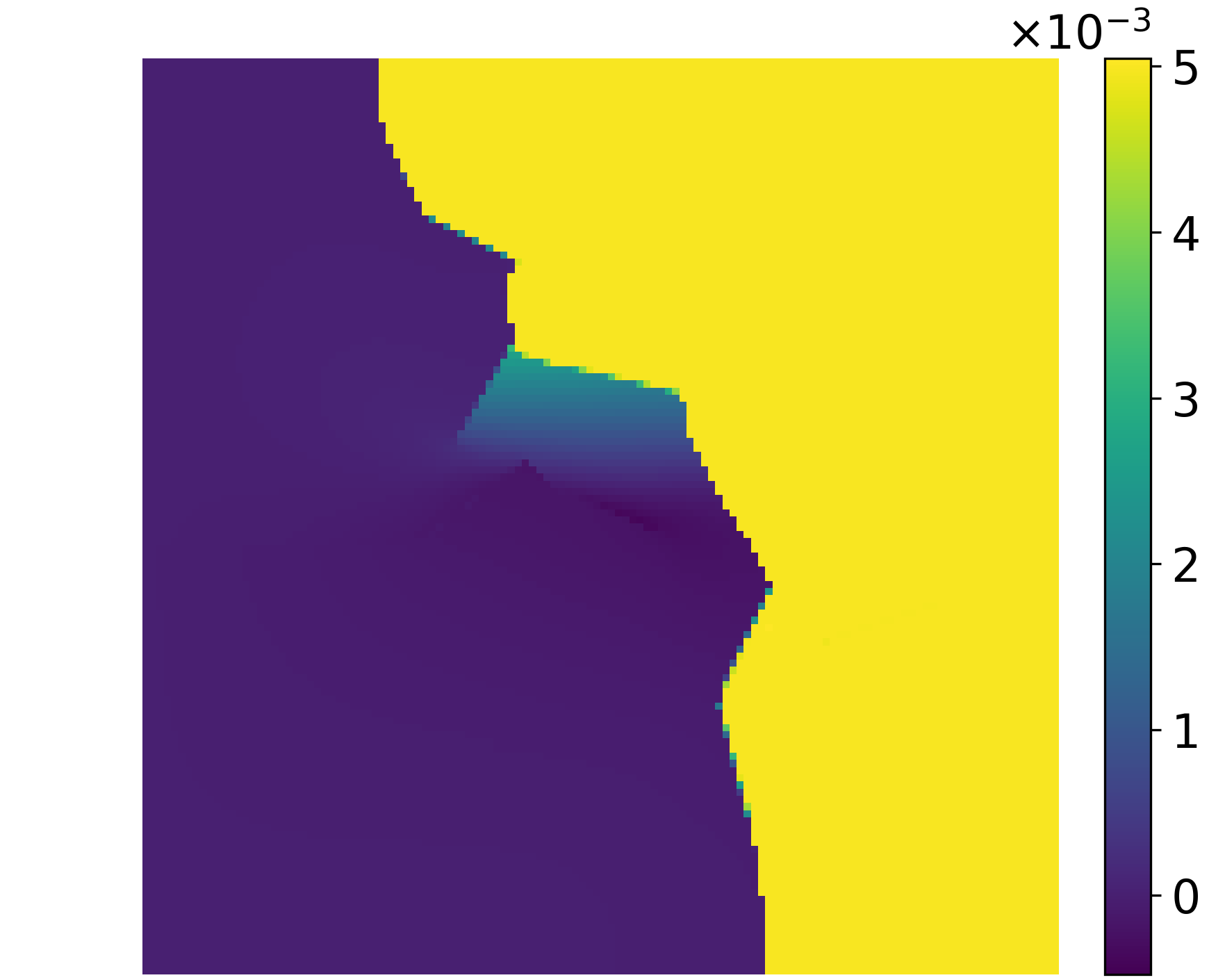}
        \caption{}
    \end{subfigure}%
    \hfill%
    \begin{subfigure}[t]{0.33\linewidth}
        \centering
        \includegraphics[width=\linewidth]{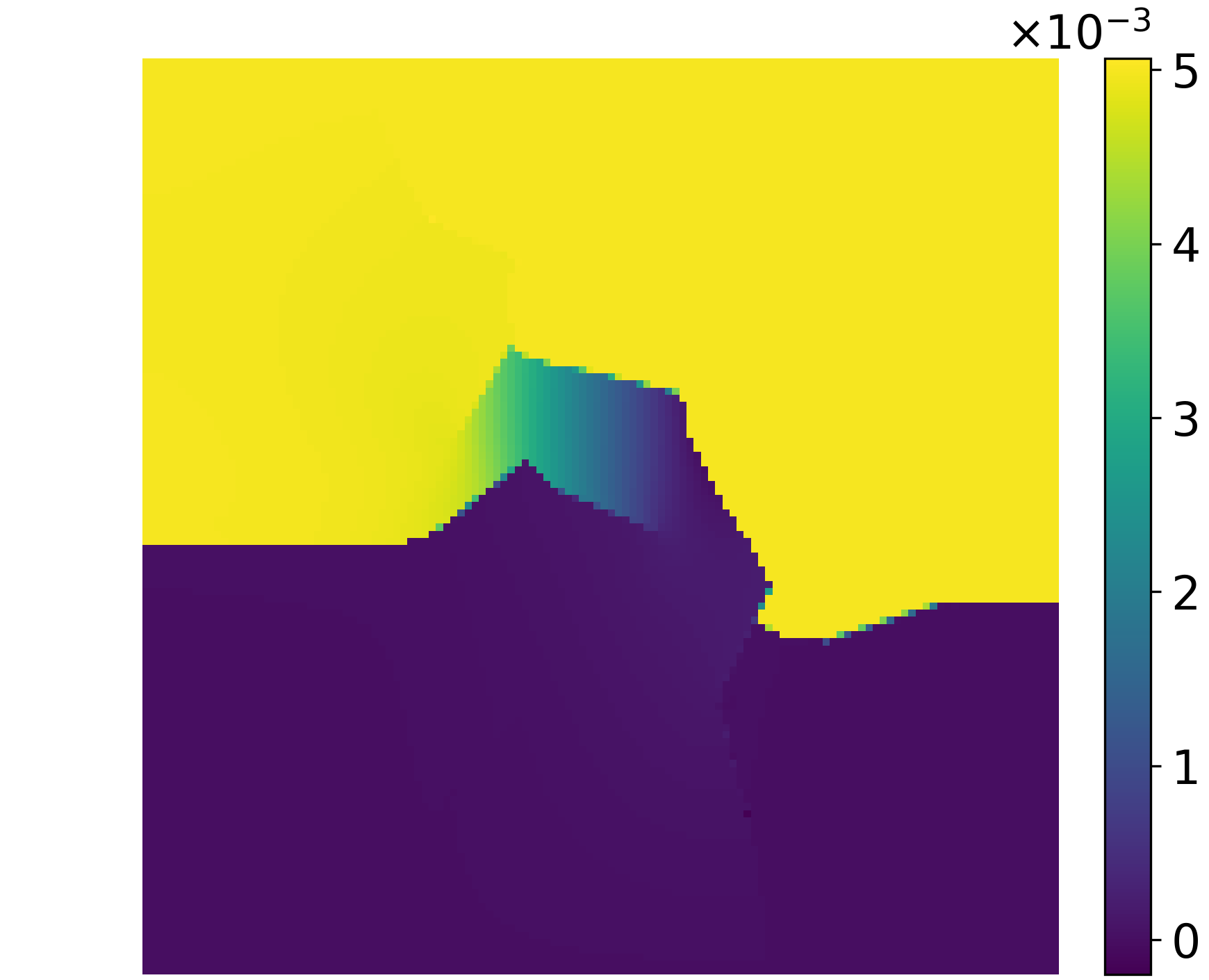}
        \caption{}
    \end{subfigure}%
    \hfill%
    \begin{subfigure}[t]{0.33\linewidth}
        \centering
        \includegraphics[width=\linewidth]{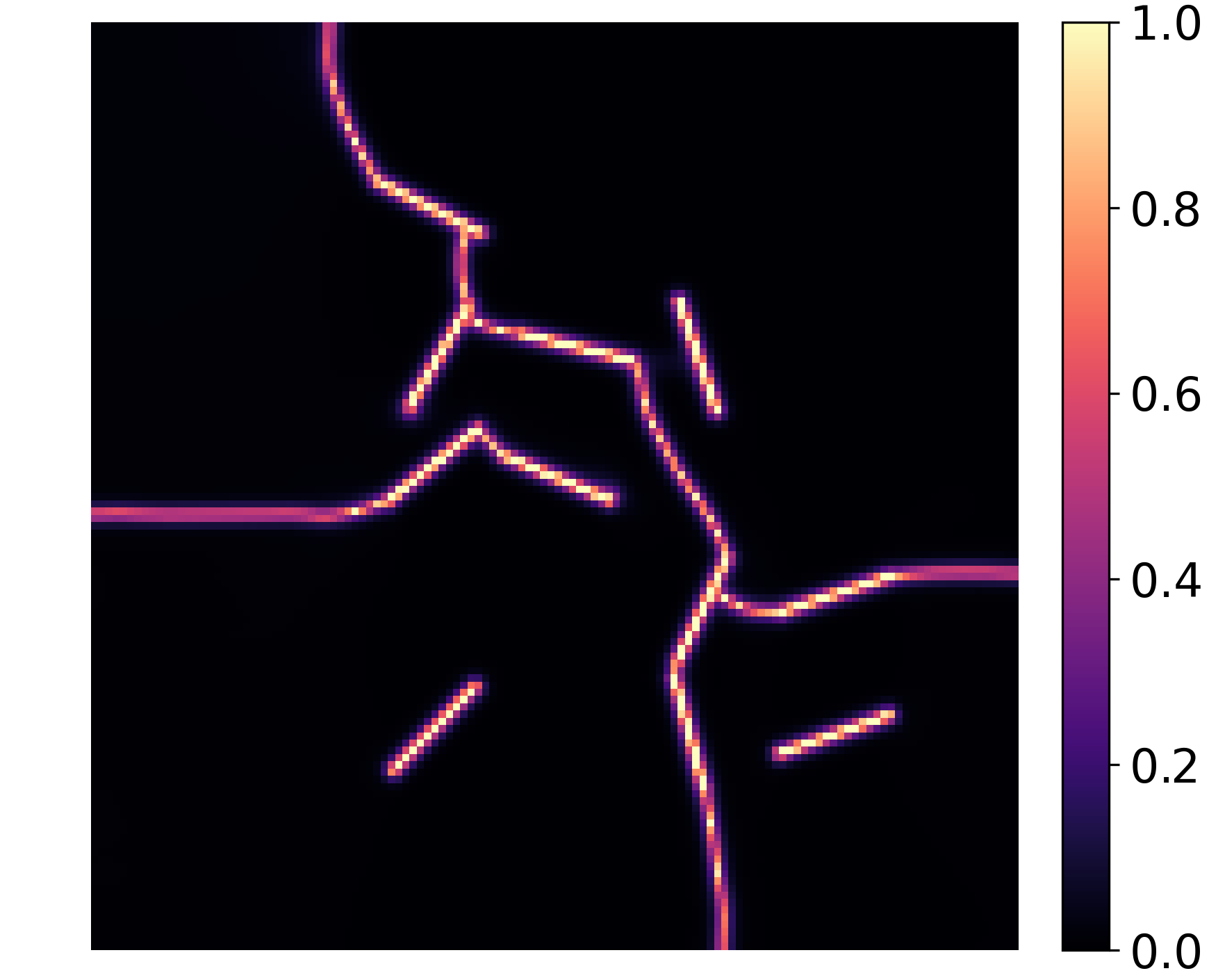}
        \caption{}
    \end{subfigure}

    \vspace{0.6ex}
    \begin{subfigure}[t]{0.33\linewidth}
        \centering
        \includegraphics[width=\linewidth]{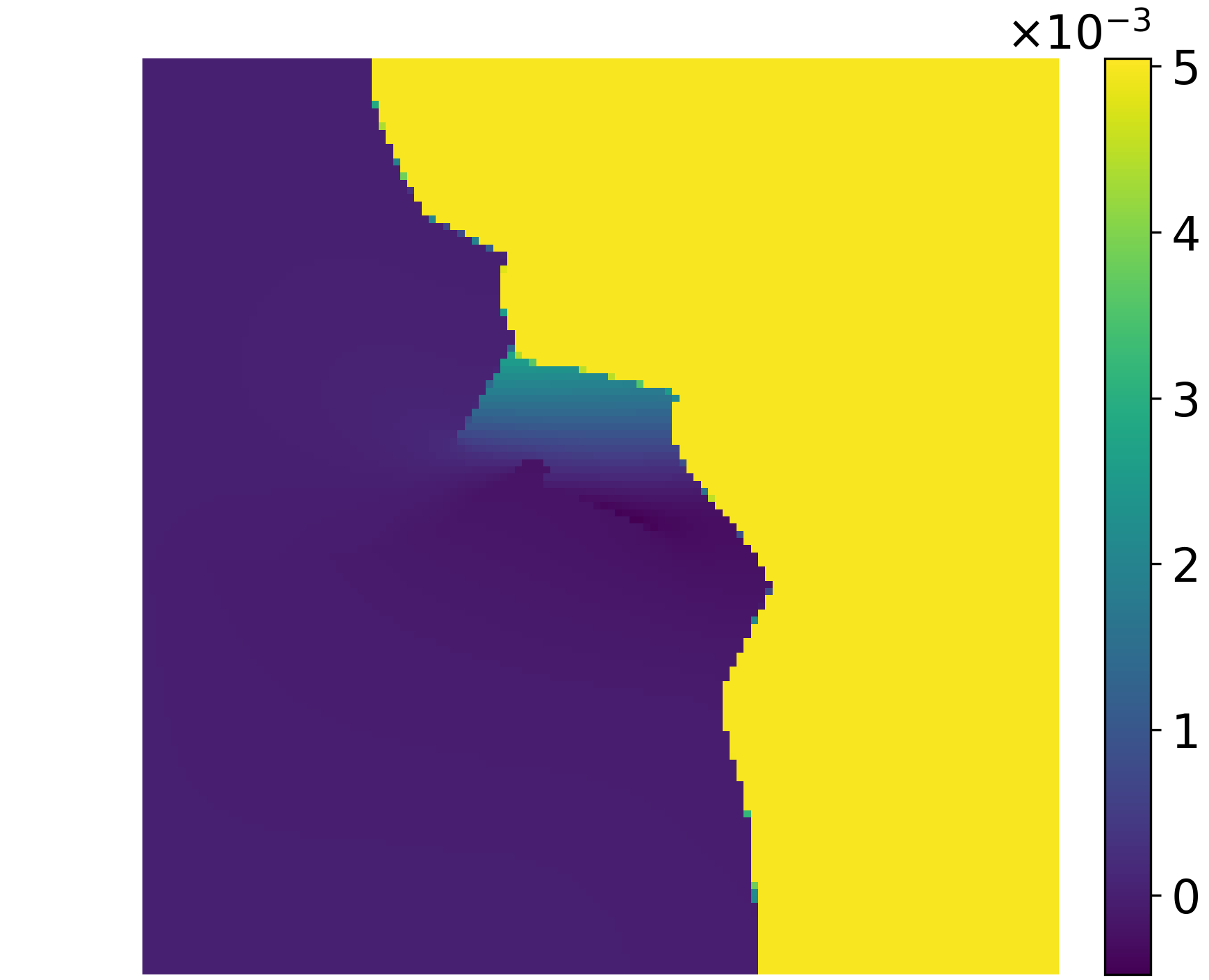}
        \caption{}
    \end{subfigure}%
    \hfill%
    \begin{subfigure}[t]{0.33\linewidth}
        \centering
        \includegraphics[width=\linewidth]{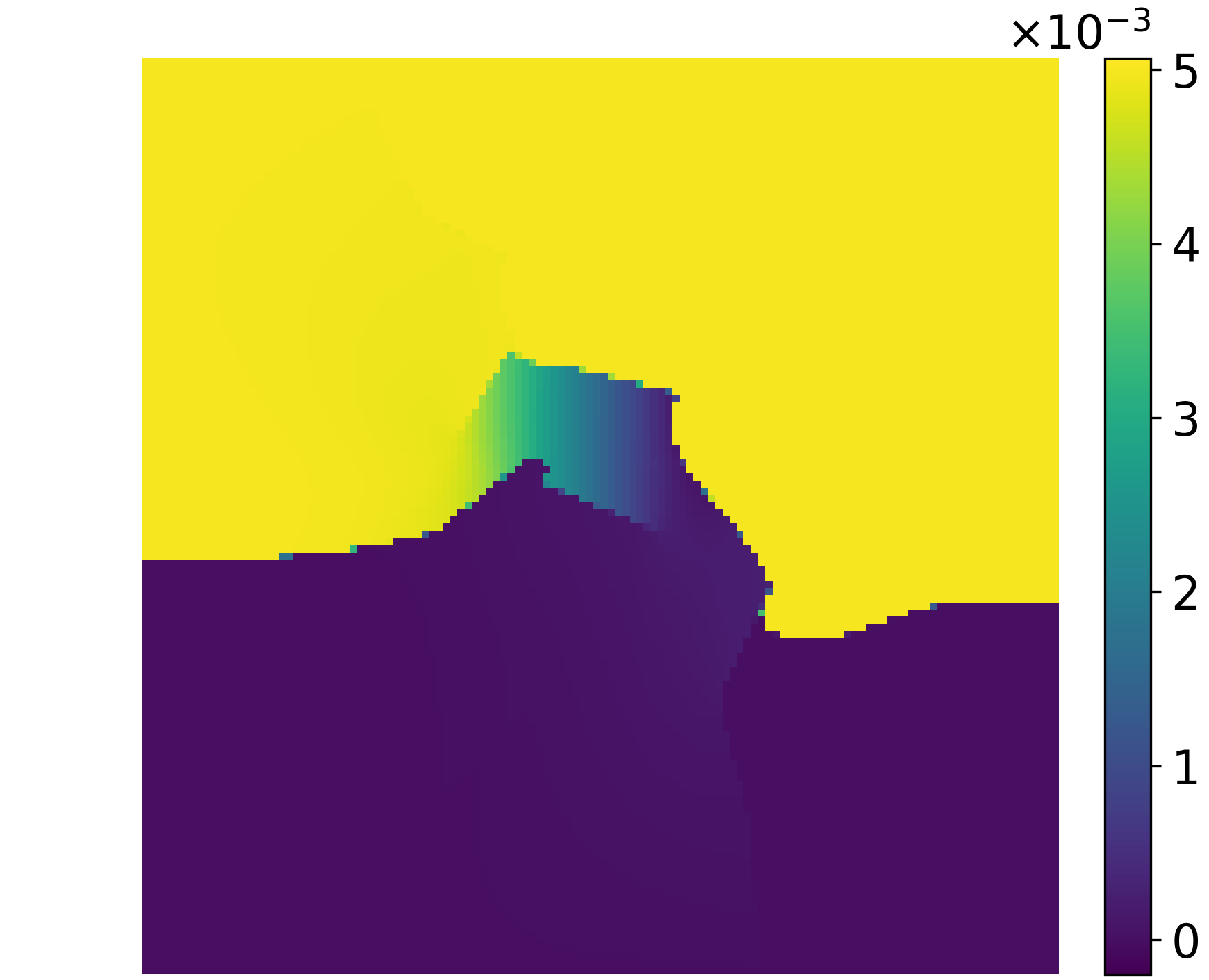}
        \caption{}
    \end{subfigure}%
    \hfill%
    \begin{subfigure}[t]{0.33\linewidth}
        \centering
        \includegraphics[width=\linewidth]{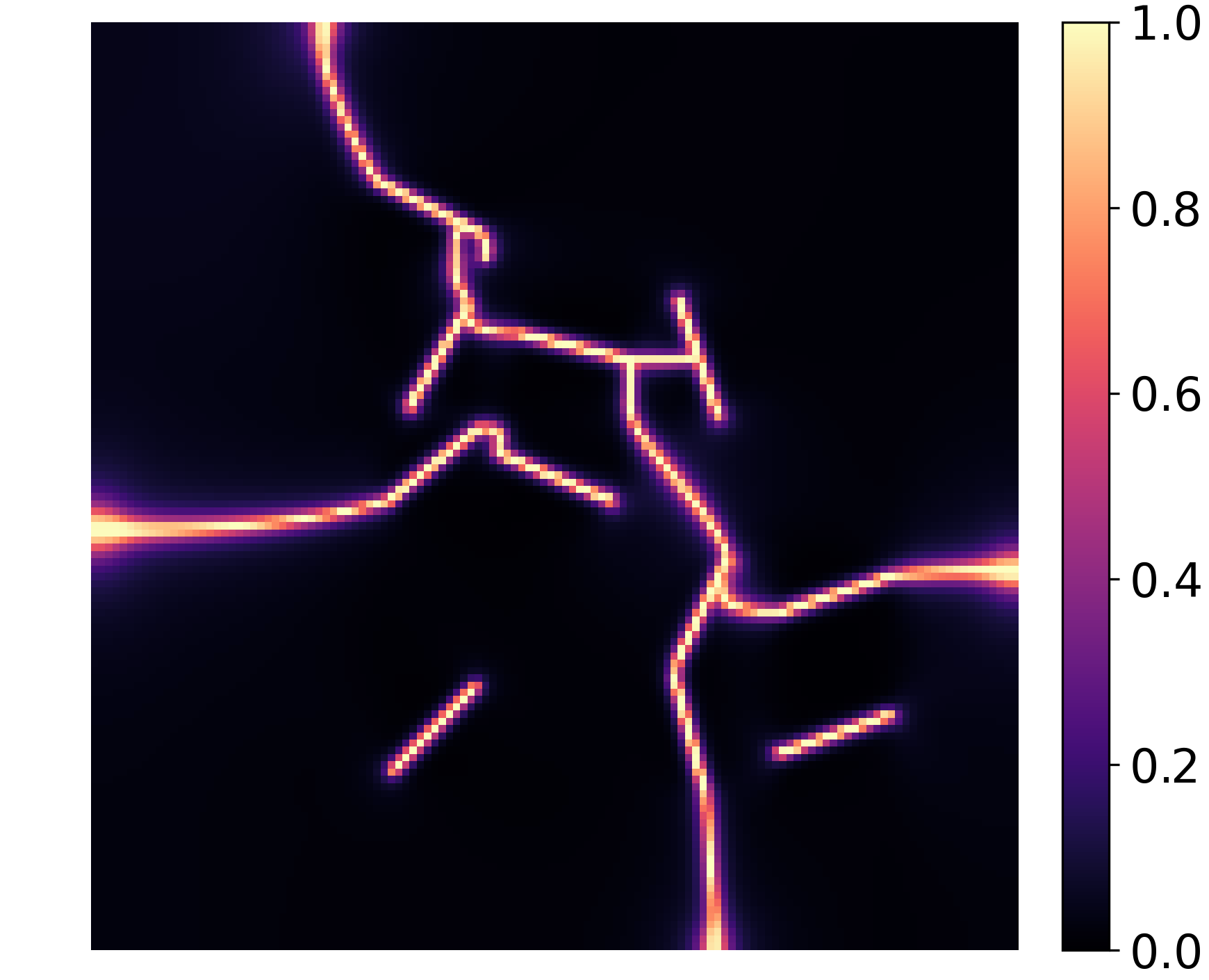}
        \caption{}
    \end{subfigure}
    \caption{Benchmark configuration $106244$ under biaxial tension
    at the end of the loading range,
    $\delta = 5.0 \times 10^{-3}\,\mathrm{mm}$, on the pixel grid of
    the dataset. (a)--(c)~Displacement components $u$, $v$ and phase
    field of the computed solution; (d)--(f)~the reference fields.
    Displacements in mm; each displacement component shares one color
    scale between the two rows. Eleven of eleven seeded cracks are
    classified correctly.}
    \label{fig:bench-fields-t}
\end{figure}

\begin{figure}[!htb]
    \centering
    \begin{subfigure}[t]{0.33\linewidth}
        \centering
        \includegraphics[width=\linewidth]{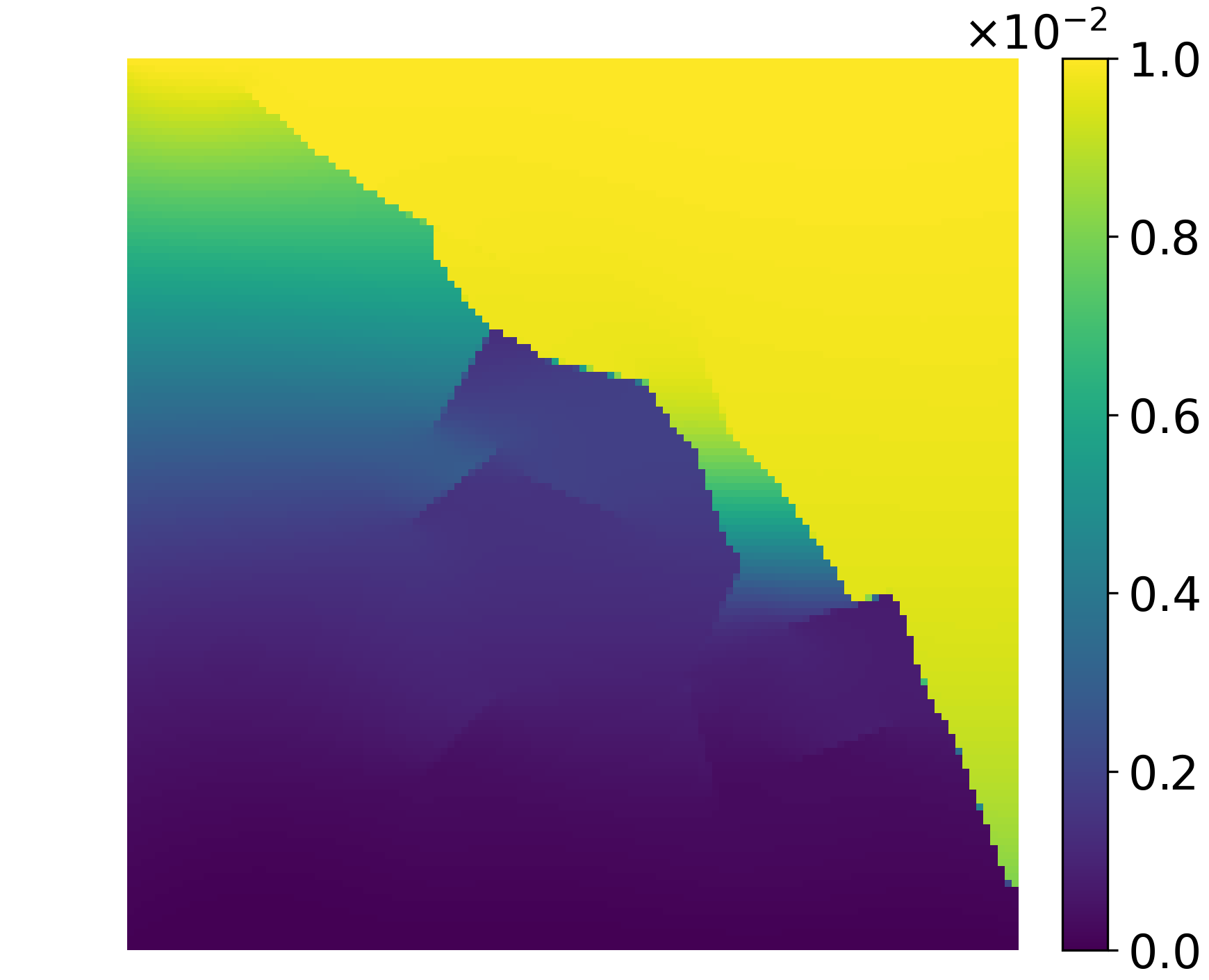}
        \caption{}
    \end{subfigure}%
    \hfill%
    \begin{subfigure}[t]{0.33\linewidth}
        \centering
        \includegraphics[width=\linewidth]{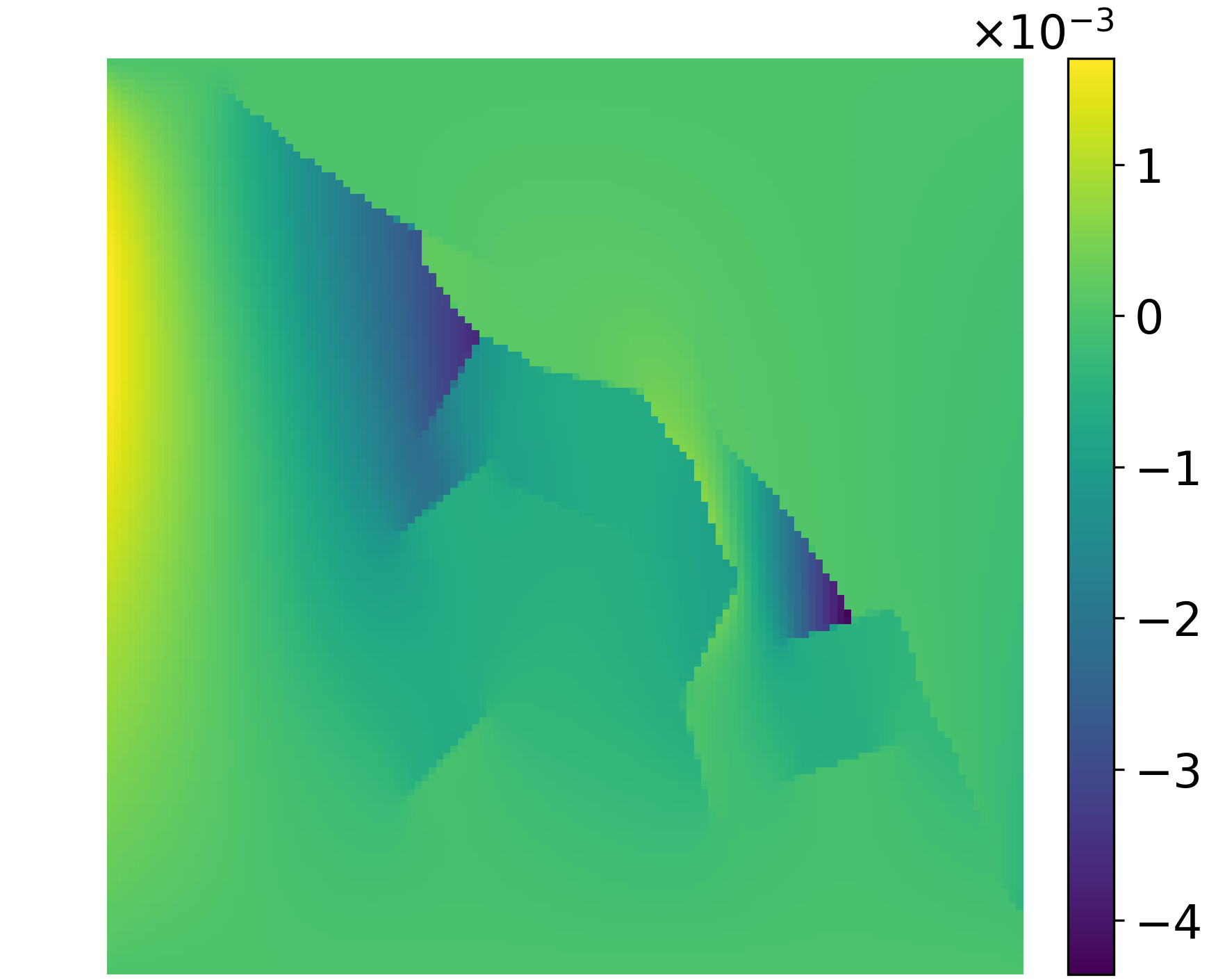}
        \caption{}
    \end{subfigure}%
    \hfill%
    \begin{subfigure}[t]{0.33\linewidth}
        \centering
        \includegraphics[width=\linewidth]{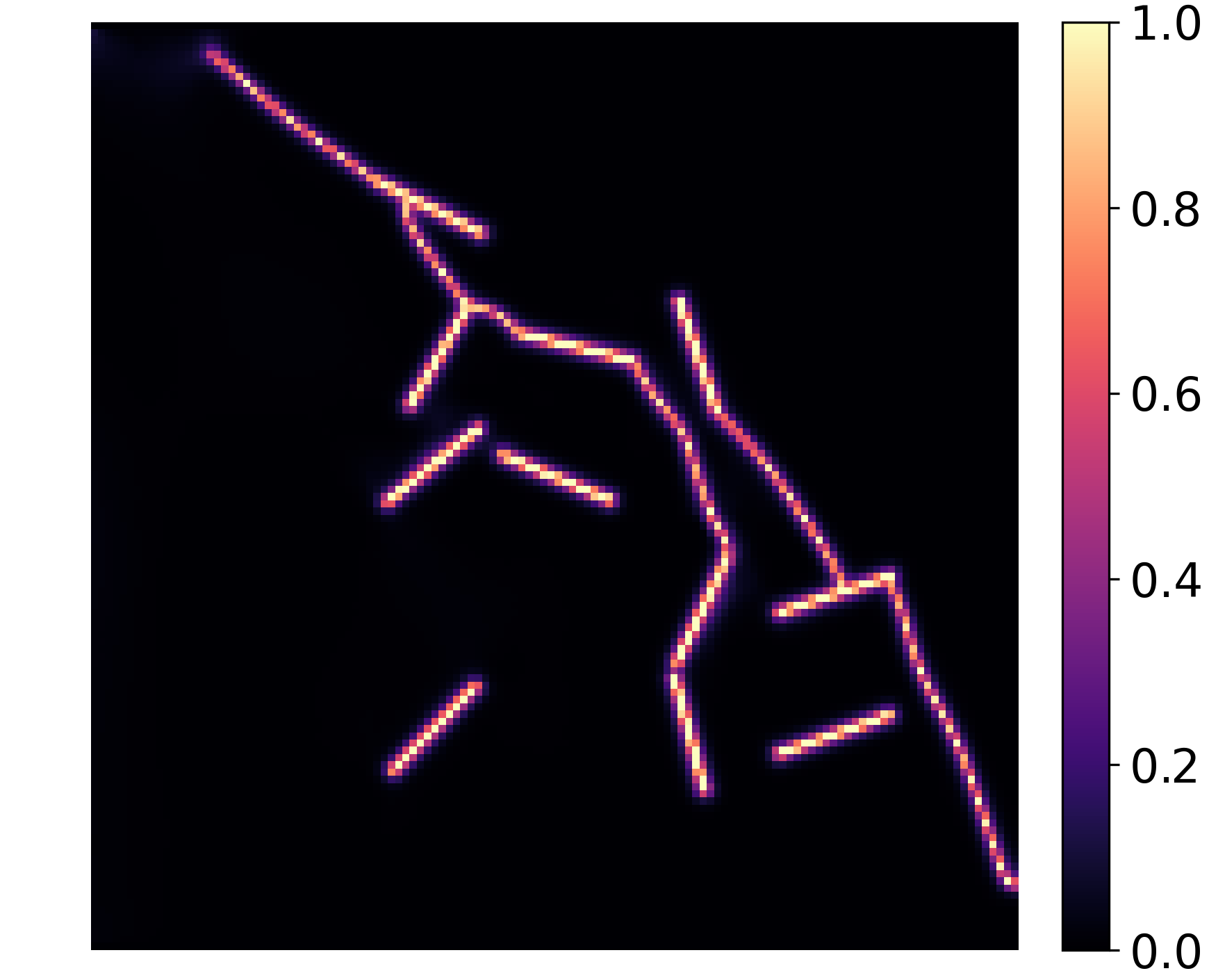}
        \caption{}
    \end{subfigure}

    \vspace{0.6ex}
    \begin{subfigure}[t]{0.33\linewidth}
        \centering
        \includegraphics[width=\linewidth]{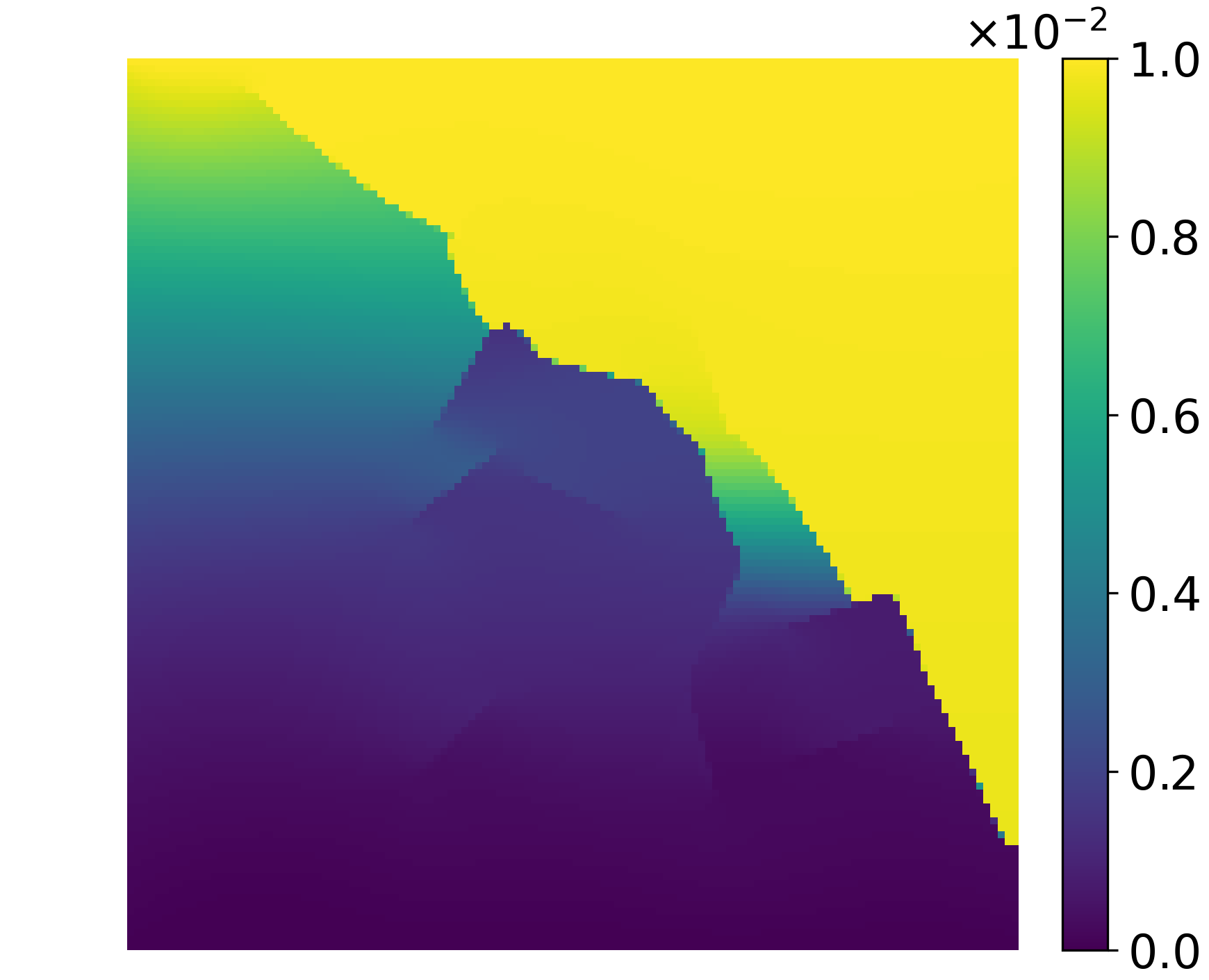}
        \caption{}
    \end{subfigure}%
    \hfill%
    \begin{subfigure}[t]{0.33\linewidth}
        \centering
        \includegraphics[width=\linewidth]{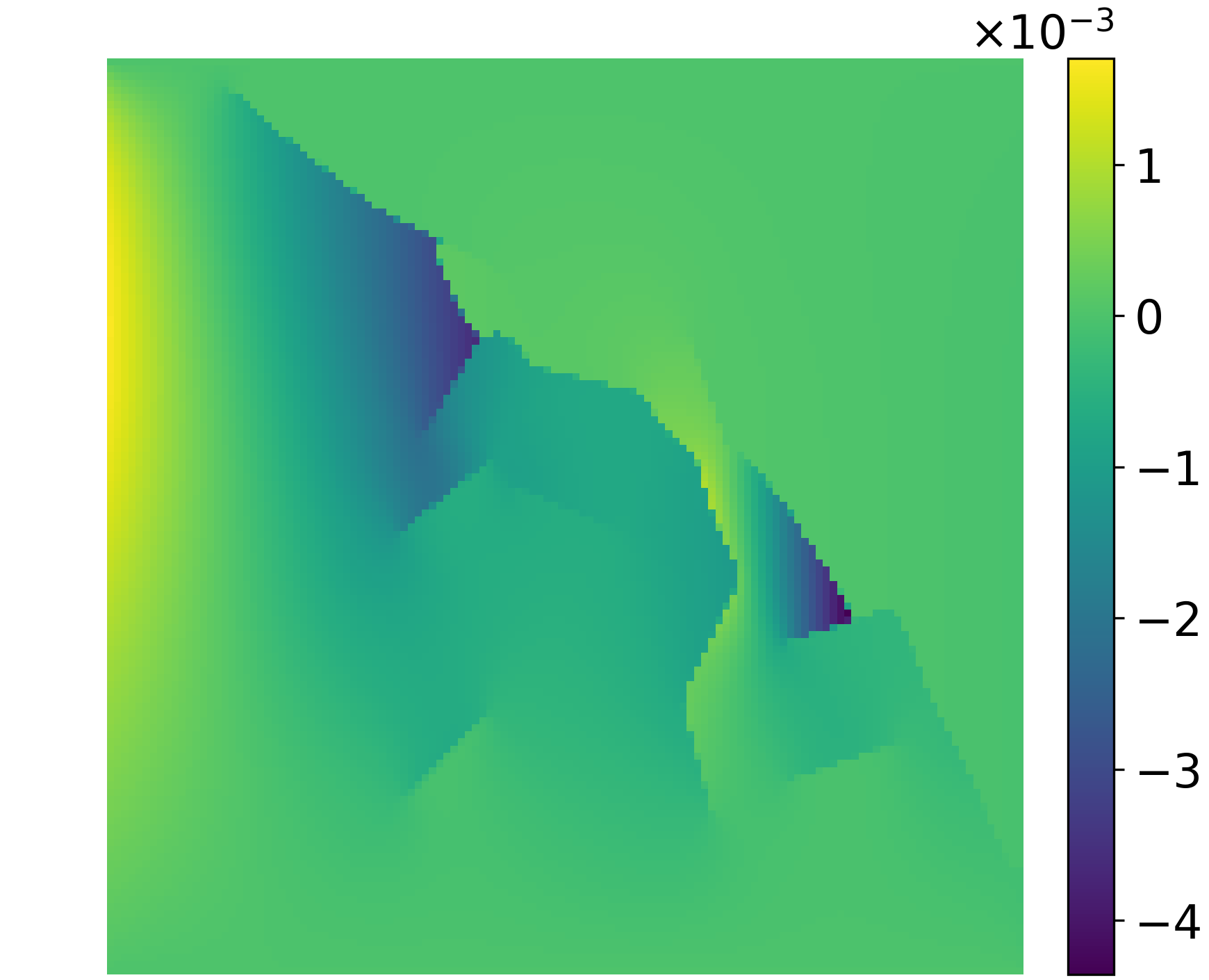}
        \caption{}
    \end{subfigure}%
    \hfill%
    \begin{subfigure}[t]{0.33\linewidth}
        \centering
        \includegraphics[width=\linewidth]{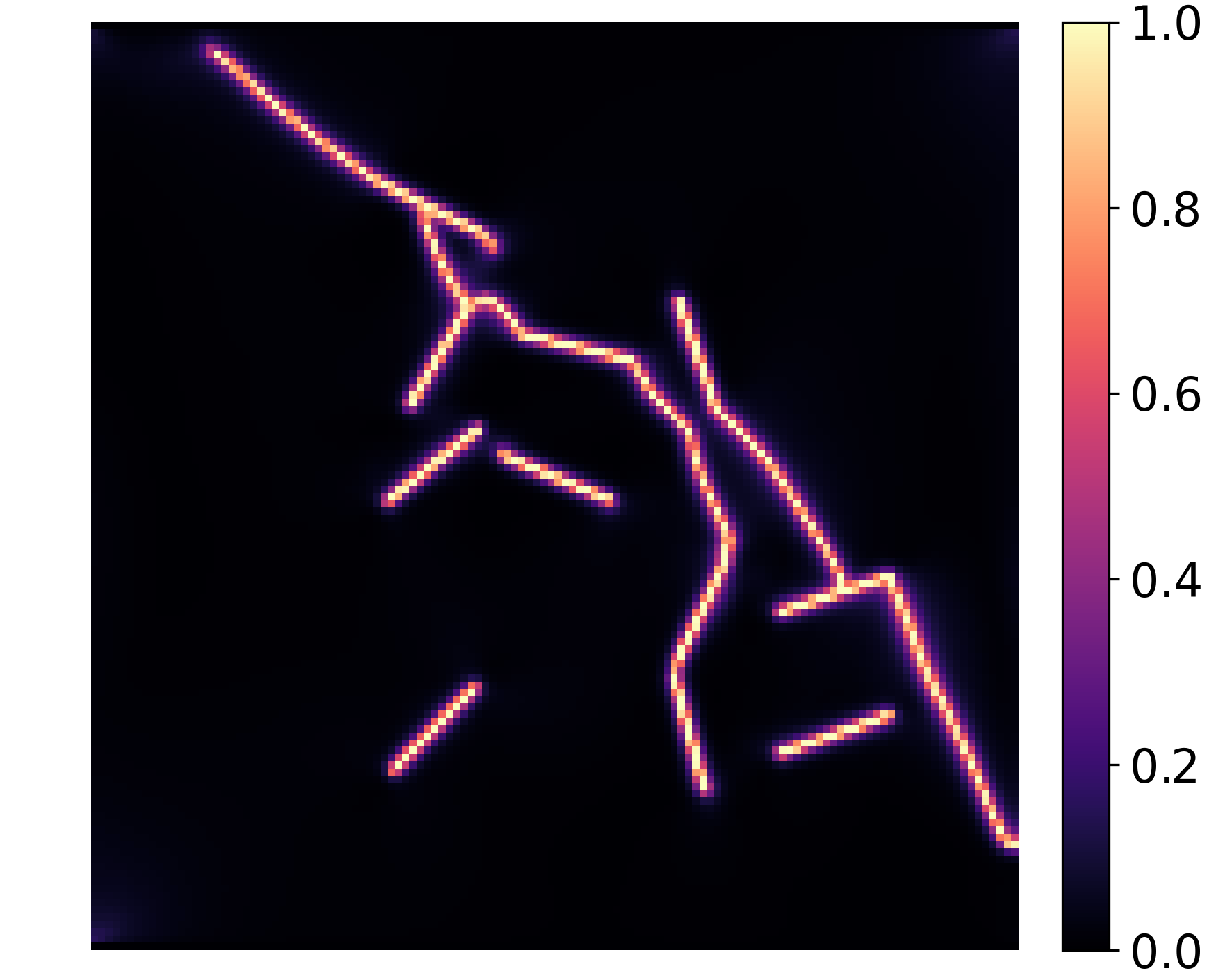}
        \caption{}
    \end{subfigure}
    \caption{Benchmark configuration $106244$ under shear at the end
    of the loading range, $\delta = 1.0 \times 10^{-2}\,\mathrm{mm}$,
    on the pixel grid of the dataset. (a)--(c)~Displacement
    components $u$, $v$ and phase field of the computed solution;
    (d)--(f)~the reference fields. Displacements in mm; each
    displacement component shares one color scale between the two
    rows. Eleven of eleven seeded cracks are classified correctly.}
    \label{fig:bench-fields-s}
\end{figure}

Fig.~\ref{fig:bench-fd} shows the load--displacement response of
configuration $106244$, classified correctly for every crack under
both loadings, and Figs.~\ref{fig:bench-fields-t}
and~\ref{fig:bench-fields-s} the corresponding final fields.
Under tension the computed curve reproduces the sequence of load
drops of the reference. The peak of $4418\,\mathrm{N}$ at
$\delta = 1.75 \times 10^{-3}\,\mathrm{mm}$ sits $4.6\%$ above and
later than the reference peak of $4223\,\mathrm{N}$ at
$1.63 \times 10^{-3}\,\mathrm{mm}$, and both curves collapse to below
$2\%$ of their peaks by the end of the range. Under shear the
computed peak of $1247\,\mathrm{N}$ at
$\delta = 6.5 \times 10^{-3}\,\mathrm{mm}$ lies $2.6\%$ above the
reference peak of $1216\,\mathrm{N}$ at
$6.2 \times 10^{-3}\,\mathrm{mm}$. The failure
that follows is reproduced, the load declining as the crack network
percolates diagonally across the
specimen, and the computed curve ends the loading range at $0.45$ of
its peak against $0.43$ for the reference. Compared at the same displacement
rather than peak to peak, the two solutions agree closely. At the
displacement where the reference attains its peak, the computed force
agrees with the reference to within about $0.5\%$ under both
loadings. The
discrepancy is therefore confined to the onset of failure. The
reference advances the load in much finer increments, and the
incremental minimization follows the metastable branch until that
branch vanishes, so failure occurs later in the computed response. The final phase fields
admit a crack-by-crack comparison, in which the same seeds grow, the
same seeds stay dormant, and the percolation path is shared; the
displacement components decompose into the same near-rigid blocks
separated by jumps across the opened cracks.

The stability of these results under network initialization is
measured on the same configuration. Repeating each loading with four
initializations moves the peak load by at most $2.0\%$ in tension and
$1.1\%$ in shear, and $87$ of the $88$ per-crack decisions are
identical across the eight runs. No two initializations of the deep Ritz baseline produced the same
crack pattern~\cite{hamdi2026}. We attribute the contrast to the
resampled estimator of Section~\ref{subsec:quadrature}, which leaves
no fixed point set for the optimizer to exploit whatever the
initialization, although the two methods differ in more than this one
ingredient.
The curves and fields of all eight runs are collected in
\ref{app:seeds}.

In one shear configuration ($105657$) damage develops along
the constrained boundary, the mode the dataset introduced its
$\pf = 0$ condition to suppress, since it dominated many of the raw
simulations~\cite{hamdi2026}; the reference field of the same
configuration exhibits a matching near-boundary band. The crack
network of this configuration percolates only beyond the loading
range of the dataset, and the classification column of
Table~\ref{tab:benchmark} counts the shortfall against the method.

% =====================================================================
\section{Conclusions}
\label{sec:conclusions}

We have presented a mesh-free discretization of phase-field modeling
of brittle fracture in which one network represents the displacement
and phase fields, a multiresolution $C^{1}$ feature encoding sets the
finest resolvable scale, and the incremental energy is minimized
directly on integration points that are redrawn at every iteration.
The framework treats the second- and the fourth-order fracture energy
densities alike, imposes essential boundary conditions exactly through
lifts, and reaches curved domains through an exact NURBS map and a
domain mask.

On the single-edge-notched benchmarks the computed load--displacement
curves agree with staggered finite element references at the same
regularization length to within about $1\%$ at the peaks, and
they reproduce the softening valley and the second local maximum of
the shear response; repeating the shear test over four random seeds
moves the peak by $0.5\%$. Since only the fracture density is
exchanged, the fourth-order model runs on the identical discretization
at a moderate increase in cost, with no higher-continuity trial space
to construct, and it retains the branching bifurcation. Crack
branching, the coalescence of en-echelon cracks, nucleation at a
circular hole and the thick-walled ring were computed without crack
tracking, remeshing or boundary-fitted meshing. The crack patterns
follow the reference solutions where these exist, with peak loads
within $8\%$ of the finite element references of the two
topology-changing tests. On the public benchmark dataset of Hamdi and
Lejeune~\cite{hamdi2026}, twenty zero-shot runs on ten random
multi-crack configurations of $11$ to $17$ cracks classify the
active or dormant state of $90\%$ of the seeded cracks correctly and
reach mean Dice scores of $0.739$ in tension and $0.824$ in shear,
against the $0.680$ and $0.733$ of the surrogates that the dataset
authors trained on eight hundred solved samples, a clear margin under
shear and a narrow one under tension; their deep Ritz baseline fails
on these configurations.
The evidence of \ref{app:failure} attributes the behavior of the
discretization across these examples to the pairing of the encoding
with the per-iteration resampling. Removing either ingredient stops
the crack from advancing, whether by holding the integration points
fixed or by coarsening the encoding beyond the width of the band,
while the method is insensitive to a resolution cap far finer than
required and to a redraw interval ten times longer than the one used,
so that the requirement on each ingredient is a threshold rather than
a narrow window. The alternative strain energy
decompositions reproduce documented failure modes.

The scope of the study is limited in several respects. All results
are quoted at a fixed
regularization length per example, and no claim about the sharp-crack
limit is made. The hybrid formulation is not variationally
consistent~\cite{ambati2015}, an inconsistency we accept since the
spectral decomposition stalls the mode II test on the shear testbed
of \ref{app:failure}. The
corner toughening is a modeling device at the grips rather
than a material statement, and the small force plateau after complete
severance stems from the same boundary treatment. Essential conditions
hold exactly wherever the prescribed data are constant along a patch
edge; on the loaded arc of the ring, where two conditions meet on one
edge, they are imposed through a boundary term instead. Irreversibility is
enforced approximately by a penalty with a fixed
weight~\cite{gerasimov2019}, which held $\pf \ge \pf_{n-1}$ to within
the dead band in every run reported here. The computational cost per load increment remains
roughly an order of magnitude above that of the
finite element references, as
measured in \ref{app:hyper} and consistent with the experience
of~\cite{manav2024}, so the method is not proposed as a faster route
to a single quasi-static forward solution. The number of trainable
parameters is likewise tied to the resolution rather than to the crack:
the feature grids are dense over the whole domain and carry between
$87\%$ and $98\%$ of the parameters of a run (\ref{app:hyper}), although
the damage band occupies a small fraction of any of these specimens. A
representation that placed its localized capacity only where the phase
field is active would remove that tie.
Finally, all examples are two-dimensional and quasi-static.

The properties that generate the computational cost, a smooth
representation of every field and a quadrature that requires no
fixed point set, are also the ones that carry over to coupled
multi-field problems. The estimator is indifferent to how many fields
the problem has and to whether they are represented by one network or
by several, so the subproblems of a staggered solve integrate on one
and the same freshly drawn point set, with no interpolation between
discretizations of the individual fields. That extension is the
subject of ongoing work.

% ---------------------------------------------------------------------
\section*{Code and data availability}

The solver, the staggered finite element reference implementation and
the scripts reproducing all examples of this paper will be made
publicly available upon publication at
\url{https://github.com/HannnZH/Meshfree-DEM-PhaseField}.

% ---------------------------------------------------------------------
\section*{Acknowledgements}

This research was undertaken with the assistance of resources from
the National Computational Infrastructure (NCI Australia), an NCRIS
enabled capability supported by the Australian Government. All
production runs of this paper were computed on the Gadi
supercomputer.

\appendix
\renewcommand{\thesection}{Appendix~\Alph{section}}
\renewcommand{\thesubsection}{\Alph{section}.\arabic{subsection}}
% ---------------------------------------------------------------------
\section{Convention map and surface-energy check for the fourth-order model}
\label{app:fourth}

Borden et al.~\cite{borden2014} write the phase field as $c$, equal to
one in intact material and zero on the crack, and use a length parameter
$\ell_{0}$. In that notation their second- and fourth-order crack
surface densities read
\begin{equation}
    \gamma_{2}^{\mathrm{B}}
    =
    \frac{(1-c)^{2}}{4\ell_{0}} + \ell_{0}\, |\grad c|^{2} ,
    \qquad
    \gamma_{4}^{\mathrm{B}}
    =
    \frac{(1-c)^{2}}{4\ell_{0}}
    + \frac{\ell_{0}}{2}\, |\grad c|^{2}
    + \frac{\ell_{0}^{3}}{4}\, (\lap c)^{2} .
    \label{eq:borden-densities}
\end{equation}
Substituting $c = 1 - \pf$ and $\ell_{0} = \lreg/2$ maps
$\Gc \gamma_{2}^{\mathrm{B}}$ and $\Gc \gamma_{4}^{\mathrm{B}}$ term by
term onto the fracture energy
densities of Eq.~\eqref{eq:at2} and~\eqref{eq:fourth}, and maps their optimal
profiles, $1 - \exp(-|x|/2\ell_{0})$ and
$1 - \exp(-|x|/\ell_{0})(1 + |x|/\ell_{0})$,
onto Eq.~\eqref{eq:profile2} and~\eqref{eq:profile4}. The identification
$\lreg = 2\ell_{0}$ follows from equating the decay lengths of the
second-order profiles, since our $\lreg$ is defined as that decay
length.

On its own optimal profile, each fracture energy density integrates to
exactly $\Gc$ per unit crack length, which is the design property behind
the second term of~Eq.~\eqref{eq:pi}. For the second-order pair the bulk and
gradient terms contribute one half each. For the fourth-order pair, the
substitution $t = 2|x|/\lreg$ in~Eq.~\eqref{eq:profile4} reduces the three
contributions to elementary integrals of $t^{k} e^{-2t}$, with the
result
\begin{equation}
    \int_{-\infty}^{\infty} \frac{\Gc\, \pf_{4}^{2}}{2\lreg}\,
    \mathrm{d}x
    = \frac{5 \Gc}{8},
    \qquad
    \int_{-\infty}^{\infty} \frac{\Gc \lreg}{4}\, (\pf_{4}')^{2}\,
    \mathrm{d}x
    = \frac{\Gc}{4},
    \qquad
    \int_{-\infty}^{\infty} \frac{\Gc \lreg^{3}}{32}\, (\pf_{4}'')^{2}\,
    \mathrm{d}x
    = \frac{\Gc}{8},
    \label{eq:split4}
\end{equation}
so the bulk, gradient and Laplacian terms carry the surface energy in
the ratio $5 : 2 : 1$ and sum to $\Gc$.

The implementation is checked against these identities directly. The
check solves no boundary-value problem; it feeds the closed-form
profile through the energy routines of the solver and asks whether
the integrals of~Eq.~\eqref{eq:split4} come back. The
seeded profile and the energy density as implemented in the solver are
evaluated on a transverse line through the band, with $\pf'$ and $\pf''$
obtained by the same two automatic-differentiation passes the solver
uses in two dimensions, and integrated by the trapezoidal rule on a
window of half-width $30\lreg$ with $4 \times 10^{5}$ points. With
$\lreg = 0.01$ and the material value $\Gc = 0.04247$ used in the
examples, both models return a surface energy of $0.042467$ per unit
crack length, an error of $-0.007\%$, and the fourth-order term
integrals evaluate to $0.026544$, $0.010617$ and $0.005306$ against the
values $5\Gc/8 = 0.026544$, $\Gc/4 = 0.010618$ and $\Gc/8 = 0.005309$
from~Eq.~\eqref{eq:split4}. The residual comes from quadrature near the band
center, where the transverse profile is only piecewise smooth, and is
orders of magnitude below any effect discussed in the paper. The check
is reproduced by the script \texttt{tests/verify\_profile.py} distributed
with the code.

% ---------------------------------------------------------------------
\section{Hyperparameters}
\label{app:hyper}

Every production run of Section~\ref{sec:results} was computed on a
single NVIDIA V100 GPU with 12 CPU cores on the Gadi system of the
National Computational Infrastructure, Australia. Under the schedules
of Section~\ref{subsec:training} the second-order runs complete in
$3.8$ hours for the tension test, $6.2$ for shear, $6.0$ for
branching, $5.2$ for coalescence, $4.3$ for the plate with a hole and
$6.6$ for the ring. The fourth-order
runs on identical discretizations cost $1.4$ to $1.6$ times as much
as their second-order counterparts, since only the additional
derivative pass of Section~\ref{subsec:autodiff} separates them. That
figure is the ratio of complete runs and is therefore larger than the
$1.3$ quoted per iteration in Section~\ref{subsec:autodiff}, which
times the iteration alone.
Table~\ref{tab:hyper} lists the
discretization and training parameters of every run.

At the width and depth used throughout, the perceptron
of Section~\ref{subsec:fields} carries $51\,075$ trainable parameters.
The three feature grids add a further $329\,728$ on the unit-square
specimens, $1\,257\,984$ on the multi-crack configurations and
$2\,236\,416$ on the ring, so that the totals are $380\,803$,
$1\,309\,059$ and $2\,287\,491$ respectively. The count is therefore
set by the encoding rather than by the network, and it measures the
resolution made available rather than the complexity of the solution:
the two-level variant of \ref{app:failure} reproduces the shear
response on $85\,635$ parameters in all.

The remaining parameters are shared by every run reported here. The
irreversibility penalty of~Eq.~\eqref{eq:penalty} uses
$\gamma_{\mathrm{ir}} = 10^{3}$ with a dead band $\tau = 0$, raised to
$\tau = 0.02$ for the multi-crack shear runs alone, and the residual
stiffness of~Eq.~\eqref{eq:g} is $\kres = 10^{-6}$. The displacement scale
is $U_{\mathrm{ref}} = 10^{-3}\,\mathrm{mm}$ on the unit-square
specimens, $5 \times 10^{-3}$ and $10^{-2}\,\mathrm{mm}$ on the
multi-crack configurations under tension and shear, equal to their
final prescribed displacements,
and $3.3 \times 10^{-2}\,\mathrm{mm}$ on the ring. The sampling
mixture of Eq.~\eqref{eq:mixture} carries the weights
$w_{\mathrm{u}} = 0.4$, $w_{\mathrm{c}} = w_{\mathrm{p}} = 0.3$, with
$\eta_{\mathrm{d}} = 0.3$ and $\beta_{\mathrm{d}} = 0.5$ in the
process-zone stratum and an auxiliary grid of $256^{2}$ cells for the
piecewise-constant densities, $512^{2}$ on the multi-crack
configurations, whose domain has twice the side length; the points
are redrawn at every
iteration, $n_{\mathrm{r}} = 1$. The corner
toughening of Eq.~\eqref{eq:gcgrip} decays over three $\lreg$. Iterations
stop when the relative range of the estimated energy over a trailing
window of $400$ iterations falls below $2 \times 10^{-4}$, or when the
budget of Table~\ref{tab:hyper} is exhausted. No weight
regularization is applied to the network, and the penalty on the
squared feature amplitudes of Section~\ref{subsec:encoding} carries
the coefficient $10^{-8}$ in every run.

The staggered finite element references ran on four CPU cores of the
same system and completed in $5.1$ hours for tension, shear and
branching and $1.4$ hours for coalescence, while resolving the
loading with far finer increments, $500$ to $800$ steps against the
$80$ to $126$ of the deep energy runs. Per load increment the finite
element solver is therefore roughly an order of magnitude cheaper, on
far more modest hardware, and
no claim is made that the method
competes with a mature finite element implementation on the cost of a
single forward solution; a comparable observation is reported for the
deep Ritz approach of~\cite{manav2024}, a systematic comparison
across several linear and nonlinear equations found the finite
element method faster at equal or better accuracy in every case
examined~\cite{grossmann2024}, and a recent survey of the field
reaches the same conclusion~\cite{ani2026}. On the benchmark of
Section~\ref{subsec:benchmark} the production runs of
Table~\ref{tab:hyper} cost about $34$ hours each on the V100, against
$2.9$ to $5.4$ hours for the reference
solutions distributed with the dataset, computed on sixteen CPU
cores, and about $24$ hours reported for the deep Ritz baseline on
the same class of GPU~\cite{hamdi2026}. The two network solvers are
thus of the same order in price, and what separates them is the
solutions they return.

\begin{table}[!htb]
    \centering
    \caption{Discretization and training parameters of the production
    runs. All runs share the network of Section~\ref{subsec:fields},
    four hidden layers of width $128$, the learning rates
    $5 \times 10^{-4}$ for the network and $2 \times 10^{-3}$ for the
    feature grids, and two feature channels per level; the
    fourth-order runs use the discretization of their second-order
    counterparts unchanged. The corner toughening of Eq.~\eqref{eq:gcgrip}
    is active in the plate with a hole and in the single-edge-notched,
    branching and coalescence specimens; it is not needed for the
    ring, and the multi-crack runs use none.}
    \label{tab:hyper}
    \footnotesize
    \begin{tabular}{lcccccc}
        \toprule
        Example & split & levels & $M$ & steps &
        $\Delta\delta$ (mm) & iterations \\
        \midrule
        SEN tension & hybrid & 32/128/384 & 12000 & 80 &
        $1.0 \times 10^{-5}$ & 2000 (4000 cold) \\
        SEN shear & hybrid & 32/128/384 & 12000 & 126 &
        $2.5 \times 10^{-5}$ & 2000 (4000 cold) \\
        branching & isotropic & 32/128/384 & 12000 & 126 &
        $2.5 \times 10^{-5}$ & 2000 (4000 cold) \\
        coalescence & hybrid & 32/128/384 & 12000 & 100 &
        $1.0 \times 10^{-5}$ & 2000 (4000 cold) \\
        plate with a hole & hybrid & 32/128/384 & 12000 & 110 &
        $1.0 \times 10^{-5}$ & 1500 (3000 cold) \\
        ring & hybrid & 64/256/1024 & 18000 & 120 &
        $3.0 \times 10^{-4}$ & 1500 (3000 cold) \\
        multi-crack tension ($\times 10$) & hybrid & 48/192/768 &
        160000 & 100 & $5.0 \times 10^{-5}$ & 3000 \\
        multi-crack shear ($\times 10$) & hybrid & 48/192/768 &
        160000 & 100 & $1.0 \times 10^{-4}$ & 3000 \\
        \bottomrule
    \end{tabular}
\end{table}

% ---------------------------------------------------------------------
\section{Failure modes of alternative configurations}
\label{app:failure}

The main text presents the proposed method of Section~\ref{sec:method}
in its production configuration. During its development a number of
alternatives that appear plausible a priori were examined on the shear
test of Section~\ref{subsubsec:sens}, one ingredient changed at a time
with everything else held fixed. The alternatives that fail reproduce
failure modes that the phase-field or deep-Ritz literature documents,
and the variations around the production setting measure the range
each ingredient tolerates. To keep the cost of the sweep moderate, the runs use a
shortened loading program of $24$ increments of
$5 \times 10^{-5}\,\mathrm{mm}$, which drives the specimen past its
peak; under this program the production configuration peaks at
$68.0\,\mathrm{N}$ and fails across the ligament, consistent with
Section~\ref{subsubsec:sens} once the coarser increments are taken into
account. The load--displacement curves of the sampling-related runs
are compared in Fig.~\ref{fig:ablation} and their final phase fields
are collected in Fig.~\ref{fig:ablation-fields}; the fields of the
rejected alternatives are collected in
Fig.~\ref{fig:failure}.

\subsection{Reuse of the integration points}

\begin{figure}[!htb]
    \centering
    \begin{subfigure}[t]{0.49\linewidth}
        \centering
        \begin{tikzpicture}
            \begin{axis}[
                paperaxis, width=\linewidth, height=5.6cm,
                xlabel={$\delta$ ($10^{-3}$ mm)}, ylabel={$F$ (N)},
                xmin=0, xmax=1.28, ymin=0, ymax=95,
                legend style={at={(0.5, -0.30)}, anchor=north,
                              legend columns=2,
                              /tikz/every even column/.append style=
                                  {column sep=0.35cm}},
            ]
                \addplot[cblue, thick]
                    table[col sep=comma,
                          x expr=\thisrow{delta_mm}*1000, y=F]
                    {figures/data/ablate_base_fd.csv};
                \addlegendentry{production, $h = 0.26\lreg$}
                \addplot[cpurple, thick, dotted]
                    table[col sep=comma,
                          x expr=\thisrow{delta_mm}*1000, y=F]
                    {figures/data/ablate_ladder1536_fd.csv};
                \addlegendentry{$h = 0.065\lreg$}
                \addplot[cgreen, thick, dash dot]
                    table[col sep=comma,
                          x expr=\thisrow{delta_mm}*1000, y=F]
                    {figures/data/ablate_enc2lev_fd.csv};
                \addlegendentry{$h = 0.79\lreg$}
                \addplot[cred, thick, dashed]
                    table[col sep=comma,
                          x expr=\thisrow{delta_mm}*1000, y=F]
                    {figures/data/ablate_enc4_fd.csv};
                \addlegendentry{$h = 50\lreg$}
            \end{axis}
        \end{tikzpicture}
        \caption{}
        \label{fig:ablation-enc}
    \end{subfigure}%
    \hfill%
    \begin{subfigure}[t]{0.49\linewidth}
        \centering
        \begin{tikzpicture}
            \begin{axis}[
                paperaxis, width=\linewidth, height=5.6cm,
                xlabel={$\delta$ ($10^{-3}$ mm)}, ylabel={$F$ (N)},
                xmin=0, xmax=1.28, ymin=0, ymax=178,
                legend style={at={(0.5, -0.30)}, anchor=north,
                              legend columns=2,
                              /tikz/every even column/.append style=
                                  {column sep=0.35cm}},
            ]
                \addplot[cblue, thick]
                    table[col sep=comma,
                          x expr=\thisrow{delta_mm}*1000, y=F]
                    {figures/data/ablate_base_fd.csv};
                \addlegendentry{$n_{\mathrm{r}} = 1$}
                \addplot[cgreen, thick, dash dot]
                    table[col sep=comma,
                          x expr=\thisrow{delta_mm}*1000, y=F]
                    {figures/data/ablate_nr10_fd.csv};
                \addlegendentry{$n_{\mathrm{r}} = 10$}
                \addplot[cpurple, thick, dotted]
                    table[col sep=comma,
                          x expr=\thisrow{delta_mm}*1000, y=F]
                    {figures/data/ablate_nr100_fd.csv};
                \addlegendentry{$n_{\mathrm{r}} = 100$}
                \addplot[corange, thick, densely dashed]
                    table[col sep=comma,
                          x expr=\thisrow{delta_mm}*1000, y=F]
                    {figures/data/ablate_nr500_fd.csv};
                \addlegendentry{$n_{\mathrm{r}} = 500$}
                \addplot[cred, thick, dashed]
                    table[col sep=comma,
                          x expr=\thisrow{delta_mm}*1000, y=F]
                    {figures/data/ablate_frozen_fd.csv};
                \addlegendentry{frozen}
            \end{axis}
        \end{tikzpicture}
        \caption{}
        \label{fig:ablation-nr}
    \end{subfigure}
    \caption{Load--displacement curves of the sampling-related runs on
    the shear testbed, under the shortened loading program, with the
    two ingredients of the pairing varied one at a time.
    (a)~The resolution cap, at a fixed point budget and with the
    points redrawn at every iteration. A cap four times finer than the
    production one and a cap three times coarser both reproduce the
    response, while a single grid too coarse to represent the band
    leaves the reaction climbing with no fracture.
    (b)~The interval $n_{\mathrm{r}}$ between redraws, at the
    production resolution. Redrawing every ten iterations is
    indistinguishable from redrawing at every one, and beyond that the
    response degrades continuously toward the frozen limit, in which
    one point set serves a whole load step.}
    \label{fig:ablation}
\end{figure}
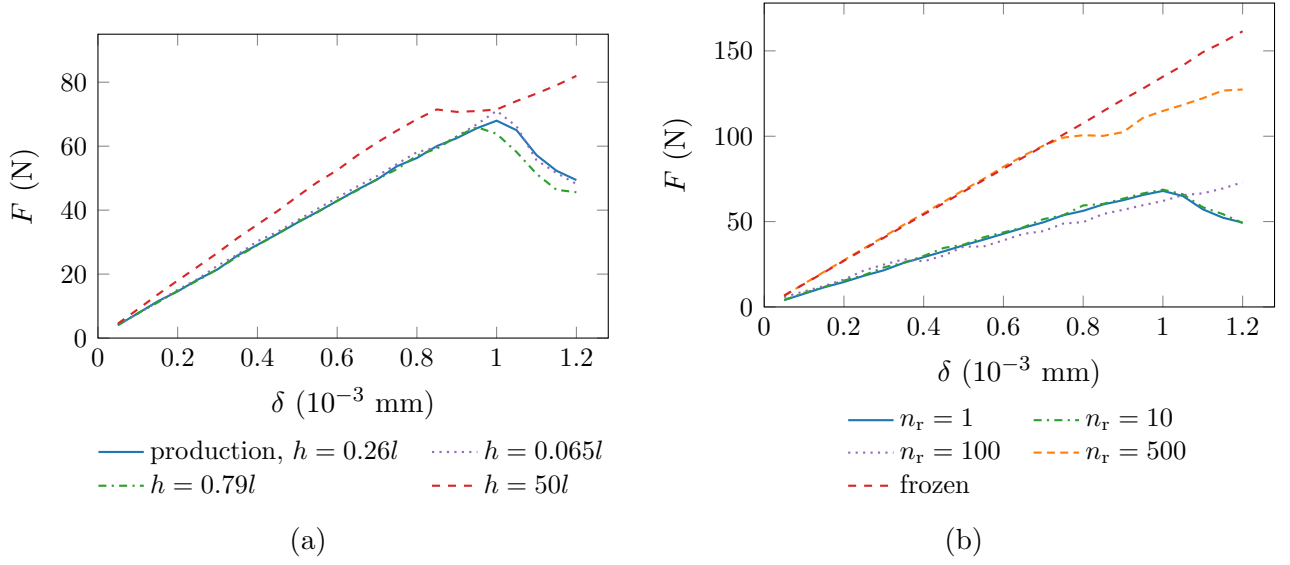

In the first run of this group the integration points of~Eq.~\eqref{eq:mc}
are drawn once at
the beginning of every load step and held fixed through all of its
iterations, instead of being redrawn every iteration from the
mixture of Eq.~\eqref{eq:mixture}. The optimizer still lowers the estimated
energy at every step, but the reaction force climbs monotonically to
$161.5\,\mathrm{N}$ at $\delta = 1.2 \times 10^{-3}\,\mathrm{mm}$,
more than twice the production peak, and no failure event occurs
(Fig.~\ref{fig:ablation-nr}). The final phase field
(Fig.~\ref{fig:failure}a) shows the state behind this curve. The
seeded band has degenerated into a ragged patch of speckled damage,
values near one alternate with intact material at the scale of the
finest encoding level, and no crack has propagated. Since the
force of Eq.~\eqref{eq:force} is evaluated on freshly drawn points, the climb
is a genuine property of the converged fields and not an artifact of
the estimator being read at its own sample.

The cause is the pairing argument of Section~\ref{subsec:pairing}.
With the point set fixed, the finest encoding levels can lower the
estimated energy by deteriorating between the points, where the
estimator cannot see the fields, and the descent of the estimate
decouples from the descent of the energy. The remedy is the
per-iteration resampling of the production configuration; restoring it
and changing nothing else returns the response to its
$68.0\,\mathrm{N}$ peak and complete failure.

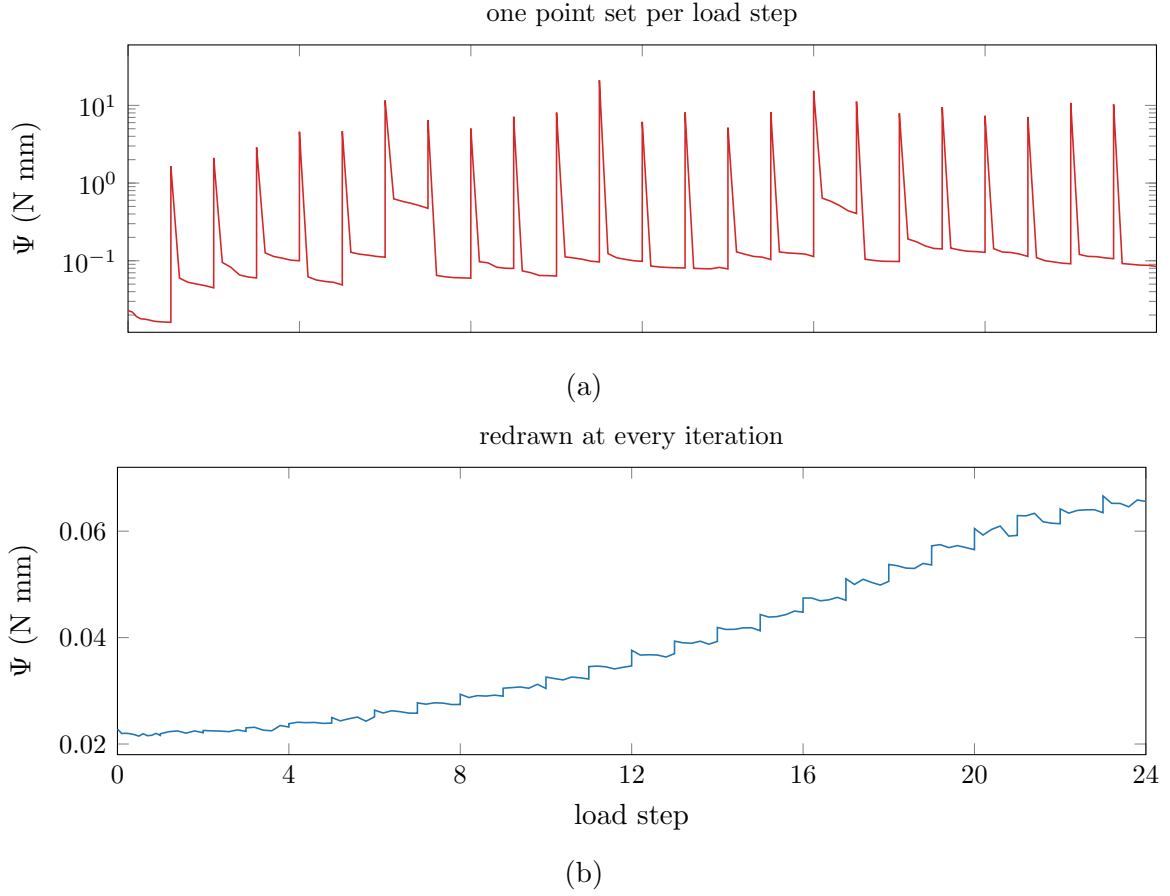
\begin{figure}[!htb]
    \centering
    \begin{subfigure}[t]{\linewidth}
        \centering
        \begin{tikzpicture}
            \begin{axis}[
                paperaxis, scale only axis,
                width=0.80\linewidth, height=3.8cm,
                title={\footnotesize one point set per load step},
                title style={yshift=-0.6ex},
                ylabel={$\Ene$ (N mm)},
                ymode=log,
                xmin=0, xmax=24, ymin=1.2e-2, ymax=6e1,
                xtick={0,4,8,12,16,20,24}, xticklabels={},
                ytick={1e-2,1e-1,1e0,1e1},
            ]
                \addplot[cred, semithick]
                    table[col sep=comma, x=s, y=Pi]
                    {figures/data/ablate_energy_frozen.csv};
            \end{axis}
        \end{tikzpicture}
        \caption{}
        \label{fig:ablenergy-frozen}
    \end{subfigure}

    \vspace{0.2em}
    \begin{subfigure}[t]{\linewidth}
        \centering
        \begin{tikzpicture}
            \begin{axis}[
                paperaxis, scale only axis,
                width=0.80\linewidth, height=3.8cm,
                title={\footnotesize redrawn at every iteration},
                title style={yshift=-0.6ex},
                xlabel={load step},
                ylabel={$\Ene$ (N mm)},
                xmin=0, xmax=24, ymin=0.018, ymax=0.072,
                xtick={0,4,8,12,16,20,24},
                ytick={0.02,0.04,0.06},
                scaled y ticks=false,
                yticklabel style={/pgf/number format/fixed,
                                  /pgf/number format/precision=2},
            ]
                \addplot[cblue, semithick]
                    table[col sep=comma, x=s, y=Pi]
                    {figures/data/ablate_energy_base.csv};
            \end{axis}
        \end{tikzpicture}
        \caption{}
        \label{fig:ablenergy-base}
    \end{subfigure}
    \caption{Estimated total energy during training, over the whole
    loading program of the shear testbed. The abscissa is the load
    step, the iterations of a step filling the unit interval that
    follows it, so that one tooth of a trace is one load step. The two
    runs differ in the interval between redraws of the integration
    points and in nothing else. (a)~One point set serves a whole load
    step. Within the step the estimate falls by about two orders of
    magnitude, and the redraw at the next integer returns it above its
    starting value, so that the descent measures the agreement between
    the fields and that one point set rather than the energy.
    (b)~The production configuration, in which the points are redrawn
    at every iteration. The estimate follows the load and carries no
    such structure, its rise across a step boundary being the load
    increment. The vertical scales differ: (a) is logarithmic over
    three decades, (b) is linear over a factor of three.}
    \label{fig:ablation-energy}
\end{figure}

The energy recorded during training makes the decoupling directly
visible (Fig.~\ref{fig:ablation-energy}). In the frozen run the
estimate falls by about two orders of magnitude over the iterations of
a load step, a median factor of $74$, and the redraw that opens the
next step undoes that descent completely, a median factor of $81$ and
as much as $218$. What falls within a step is therefore not the energy
but the discrepancy between the fields and one particular point set,
and a fresh set recovers it in full. Across the same step boundaries
the production run rises by a median factor of $1.06$, which is the
load increment itself, since a set that is never reused leaves nothing
for the next draw to expose. The first step shows the mechanism in its
purest form. The frozen run closes it a quarter below the production
estimate, on fields that are the worse of the two, and the first
redraw raises that estimate by a factor of $102$.

The failure is therefore not a statement about the accuracy of one
integration rule against another. What the fixed set offers the
optimizer is the opportunity to place error where the estimate cannot
register it, and a fixed rule of any construction, Gauss included,
offers the same opportunity to a representation fine enough to exploit
it. This is visible in the interval between redraws.
Repeating the run with the points held for
$n_{\mathrm{r}} = 10$ iterations rather than one leaves the response
unchanged, at a peak of $68.7\,\mathrm{N}$ and a complete failure,
although the estimate then sees ten times fewer distinct points over
the load step. Holding them for $100$ iterations already removes the
failure event, at $73.0\,\mathrm{N}$, and for $500$ iterations the
reaction reaches $127.4\,\mathrm{N}$ with the seeded band visibly
speckled, so that the frozen run above is the end of a continuous
progression rather than an isolated case
(Fig.~\ref{fig:ablation-nr}; the fields are collected in
Fig.~\ref{fig:ablation-fields}c--e). What the estimator requires is thus not
a large number of points but a point set that no representation can
learn, and ten iterations of reuse are short enough to keep it so.

\subsection{Resolution of the encoding}

\begin{figure}[!htb]
    \centering
    \begin{subfigure}[t]{0.33\linewidth}
        \centering
        \includegraphics[width=\linewidth]{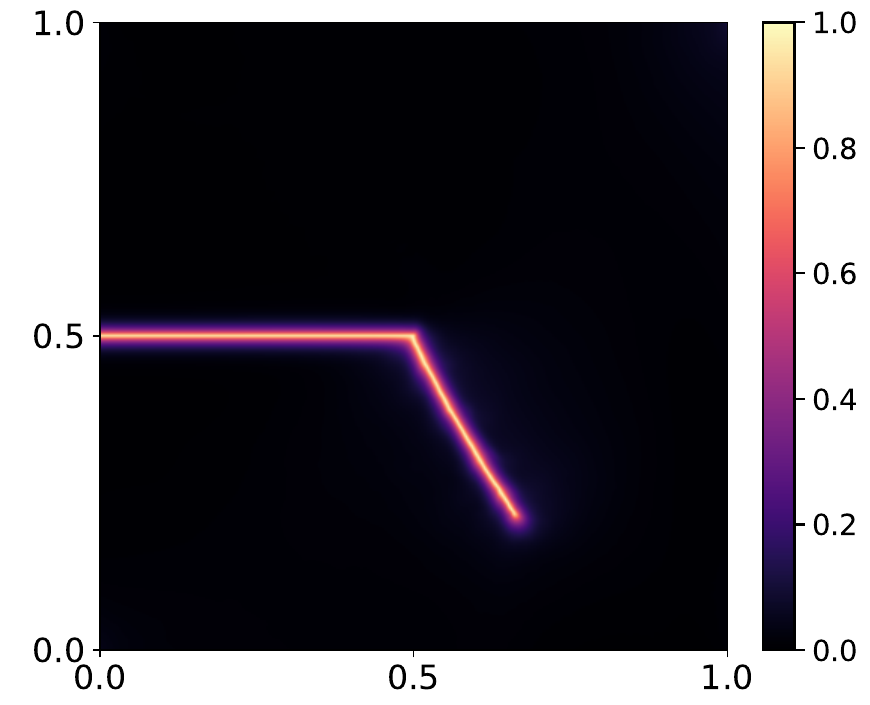}
        \caption{}
        \label{fig:ablfield-enc2lev}
    \end{subfigure}%
    \hspace{0.04\linewidth}%
    \begin{subfigure}[t]{0.33\linewidth}
        \centering
        \includegraphics[width=\linewidth]{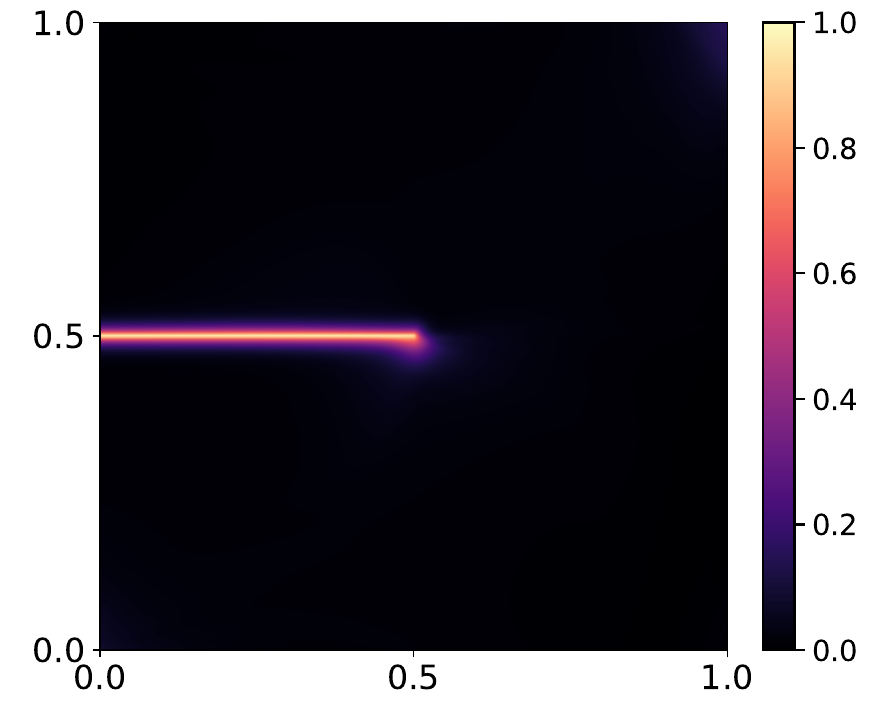}
        \caption{}
        \label{fig:ablfield-enc4}
    \end{subfigure}

    \vspace{0.4em}
    \begin{subfigure}[t]{0.31\linewidth}
        \centering
        \includegraphics[width=\linewidth]{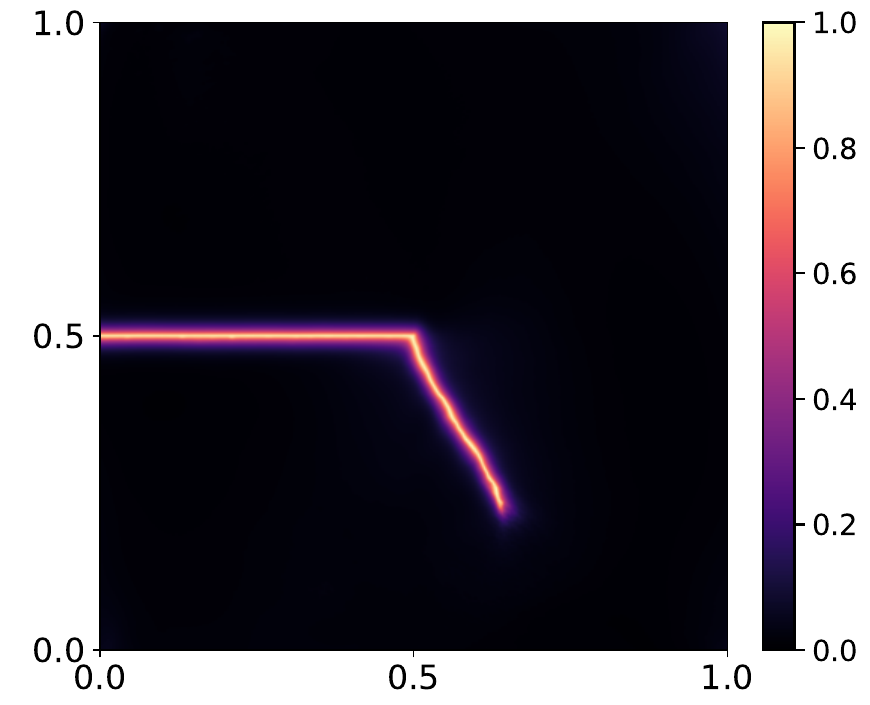}
        \caption{}
        \label{fig:ablfield-nr10}
    \end{subfigure}%
    \hspace{0.02\linewidth}%
    \begin{subfigure}[t]{0.31\linewidth}
        \centering
        \includegraphics[width=\linewidth]{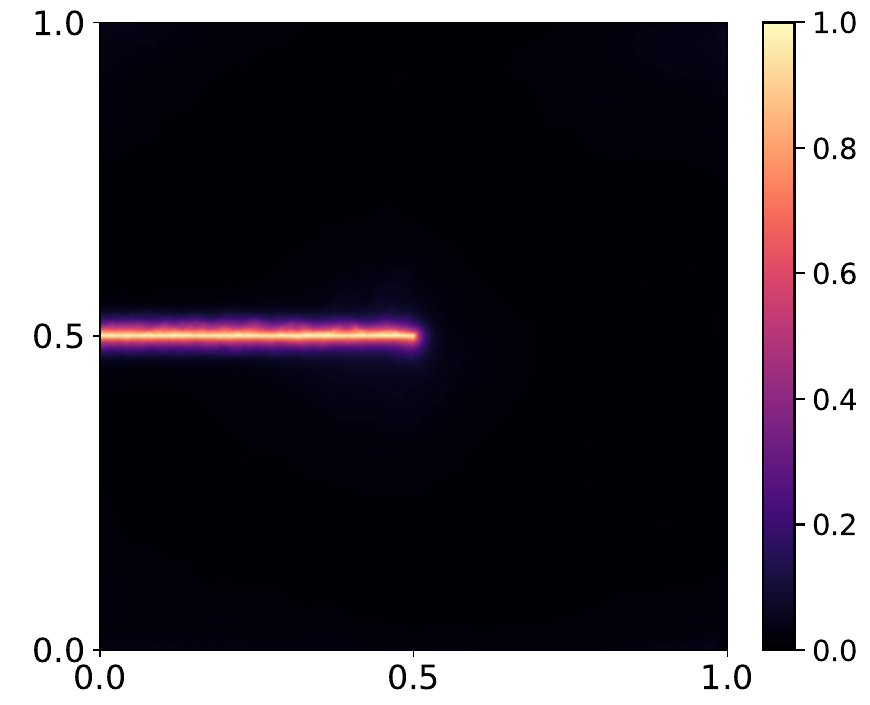}
        \caption{}
        \label{fig:ablfield-nr100}
    \end{subfigure}%
    \hspace{0.02\linewidth}%
    \begin{subfigure}[t]{0.31\linewidth}
        \centering
        \includegraphics[width=\linewidth]{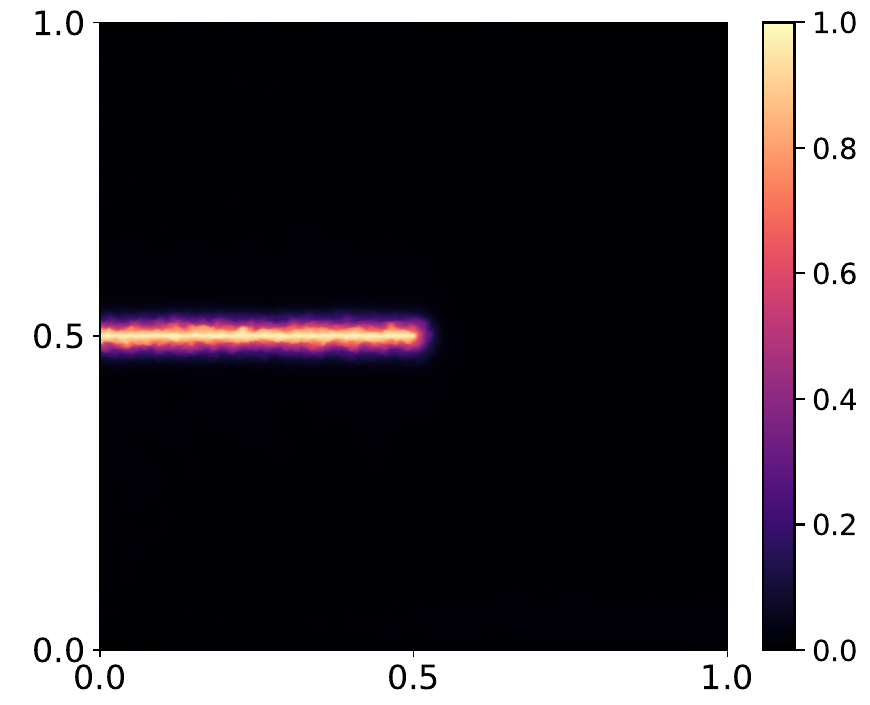}
        \caption{}
        \label{fig:ablfield-nr500}
    \end{subfigure}
    \caption{Final phase fields behind the curves of
    Fig.~\ref{fig:ablation}, at $\delta = 1.2 \times 10^{-3}$ mm.
    Top row, the resolution of the encoding:
    (a)~with the finest level removed, $h = 0.79\lreg$, the crack still
    leaves the notch, kinks and crosses the ligament;
    (b)~on a single coarse grid, $h = 50\lreg$, the seeded profile has
    not advanced and only a faint diffuse damage sits past its tip.
    Bottom row, the interval between redraws:
    (c)~at $n_{\mathrm{r}} = 10$ the crack is indistinguishable from
    the production one;
    (d)~at $n_{\mathrm{r}} = 100$ the notch has not propagated;
    (e)~at $n_{\mathrm{r}} = 500$ the seeded band itself has broken
    into speckle at the scale of the finest encoding level, the state
    the frozen run of Fig.~\ref{fig:failure}a reaches in full.}
    \label{fig:ablation-fields}
\end{figure}

The second ingredient of the pairing was varied in the same way, with
the resampling left active throughout and the point budget unchanged.
Raising the finest level from $384$ to $1536$, a cap four times finer
than the regularization length requires, leaves the response
essentially unaffected, with a peak of $71.0\,\mathrm{N}$, the same
crack path and the same failure event. Lowering it in the other
direction, to the two levels $32/128$ and a spacing of
$h = 0.79\lreg$, still reproduces the response, at
$65.9\,\mathrm{N}$, within $3.1\%$ of the production peak, and
still fails across the ligament (Fig.~\ref{fig:ablation-fields}a).

A single level of resolution $4$, which places $h$ at $50\lreg$ and
leaves the representation with no capacity to localize anything on the
scale of the band, behaves differently. The crack does not advance at
all. The seeded profile remains where it was placed, the region beyond
its tip carries only a faint diffuse damage, and the reaction climbs
to $82.0\,\mathrm{N}$ at the final increment without a failure event
(Figs.~\ref{fig:ablation-enc} and~\ref{fig:ablation-fields}b). The
resampling alone therefore does not
carry the method either.

Taken together the two subsections bracket the pairing of
Section~\ref{subsec:pairing} from both sides. Neither ingredient
suffices alone, since removing either one of them stops the crack from
propagating, and the requirement on each is a threshold rather than a
narrow window, since a cap four times finer and a cap three times
coarser than the production value both reproduce the response, and a
redraw interval ten times longer does the same. The choice of the level
resolutions and of $n_{\mathrm{r}}$ is accordingly not a delicate one,
provided the band is representable and the point set is not reused for
long.

\subsection{Spectral decomposition under shear}

Selecting the driving energy by the spectral
decomposition of Eq.~\eqref{eq:spectral} instead of the hybrid formulation
stalls the mode II test. The crack never leaves the seeded tip
(Fig.~\ref{fig:failure}b), and
the reaction climbs monotonically to $104.8\,\mathrm{N}$ at
$\delta = 1.2 \times 10^{-3}\,\mathrm{mm}$ with no failure event.

The origin of the stall is kinematic. In pure shear the principal strains come in
opposite pairs and only the tensile one enters the driving part, so
the band receives about half of the available driving force, too
little to advance at the loads the test reaches. The difficulty of
the mode II test under spectral-type driving is a known
observation~\cite{ambati2015}, and the discretization reproduces it
faithfully; the formulation, not the solver, is what stalls.

\subsection{Corner damage under the volumetric--deviatoric
decomposition}

\begin{figure}[!htb]
    \centering
    \begin{subfigure}[t]{0.33\linewidth}
        \centering
        \includegraphics[width=\linewidth]{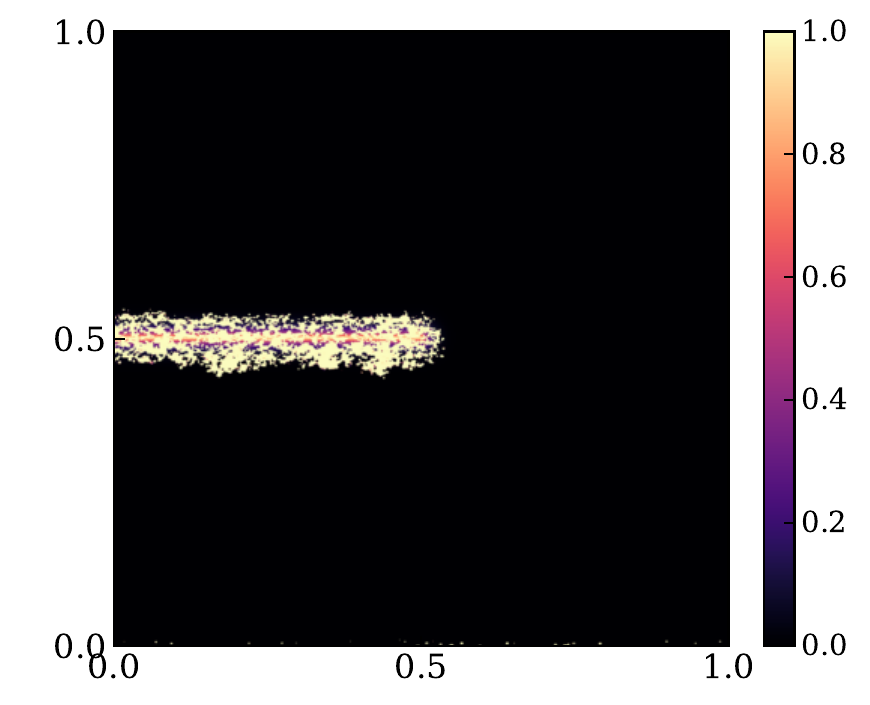}
        \caption{}
    \end{subfigure}%
    \hspace{0.04\linewidth}%
    \begin{subfigure}[t]{0.33\linewidth}
        \centering
        \includegraphics[width=\linewidth]{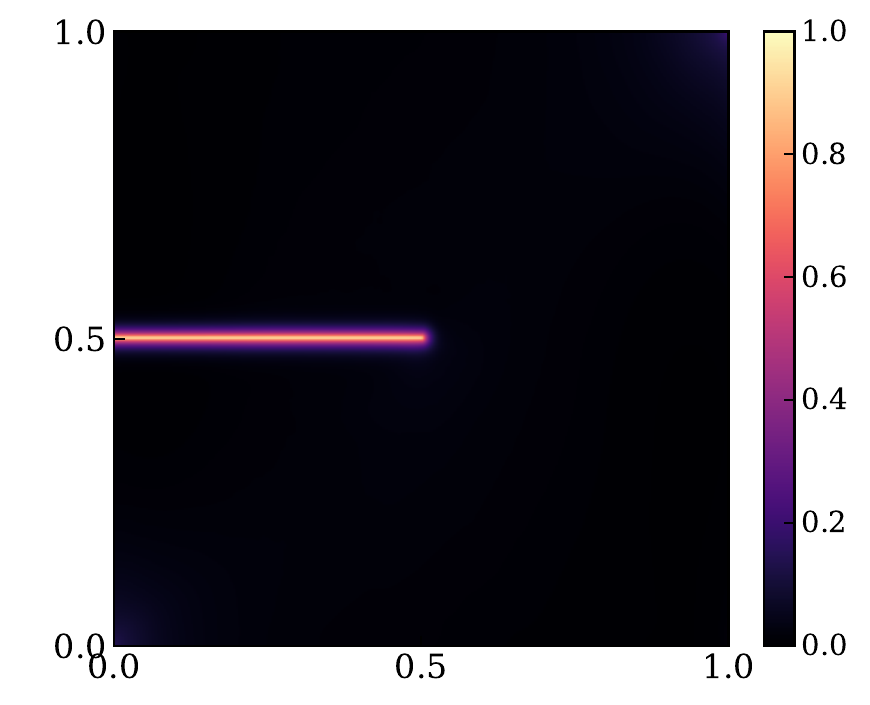}
        \caption{}
    \end{subfigure}

    \vspace{0.6ex}
    \begin{subfigure}[t]{0.33\linewidth}
        \centering
        \includegraphics[width=\linewidth]{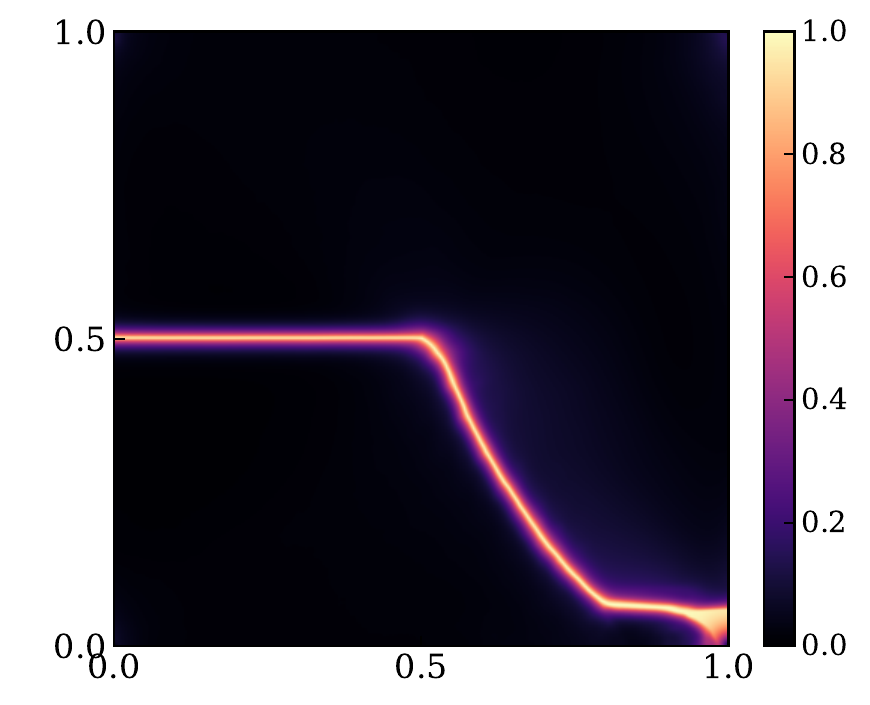}
        \caption{}
    \end{subfigure}%
    \hspace{0.04\linewidth}%
    \begin{subfigure}[t]{0.33\linewidth}
        \centering
        \includegraphics[width=\linewidth]{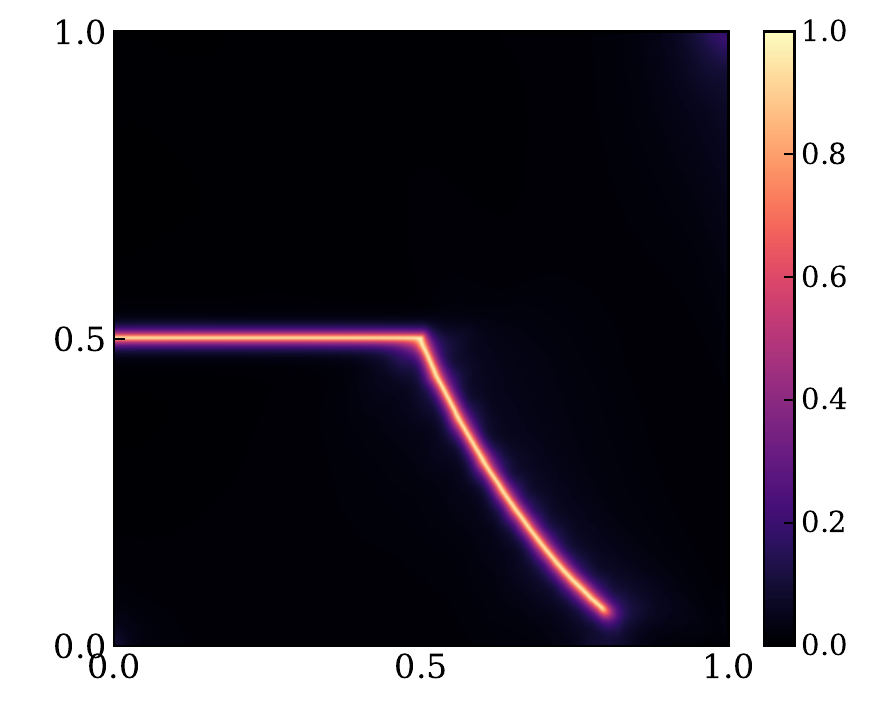}
        \caption{}
    \end{subfigure}
    \caption{Final phase fields of the rejected configurations on the
    shear testbed. (a)~Frozen integration points at
    $\delta = 1.2 \times 10^{-3}\,\mathrm{mm}$, speckled damage and no
    propagation; (b)~spectral decomposition at the same displacement,
    band stalled at the seeded tip; (c)~volumetric--deviatoric
    decomposition without corner toughening at
    $\delta = 1.8 \times 10^{-3}\,\mathrm{mm}$, spurious damage at the
    fixed-edge corner; (d)~hybrid formulation on the identical program,
    clean band and intact corners.}
    \label{fig:failure}
\end{figure}

The volumetric--deviatoric decomposition of Eq.~\eqref{eq:amor} was probed on
a longer program of $36$ increments of $5 \times 10^{-5}\,\mathrm{mm}$
with the corner toughening of Eq.~\eqref{eq:gcgrip} switched off. The crack
itself behaves correctly, with a peak of $64.4\,\mathrm{N}$ and full
severance of the ligament by
$\delta = 1.8 \times 10^{-3}\,\mathrm{mm}$, but the fixed-edge corner
accumulates damage to values near one (Fig.~\ref{fig:failure}c), a
spurious zone with no counterpart in the reference solution.

The damage traces back to the shear-rich stress concentration at the
corner of the fixed edge. Under this decomposition all deviatoric
deformation
drives damage, the corner concentration supplies it in abundance, and
a discretization that resolves $\lreg$ everywhere cannot ignore it.
As a control, the hybrid formulation of Eq.~\eqref{eq:hybrid-folded} was run
on the identical program, also without the toughening, and shows a
clean single band with no corner damage (Fig.~\ref{fig:failure}d), a
peak of $68.5\,\mathrm{N}$, and a reaction of $47.4\,\mathrm{N}$ at
the final displacement, in line with the production run of
Section~\ref{subsubsec:sens} at the same load level. The corner
toughening is therefore required under the volumetric--deviatoric
decomposition and acts as a safeguard under the hybrid one; the
production configuration applies it in both cases.

% ---------------------------------------------------------------------
\section{Seed ensembles of the random
multi-crack configurations}
\label{app:seeds}

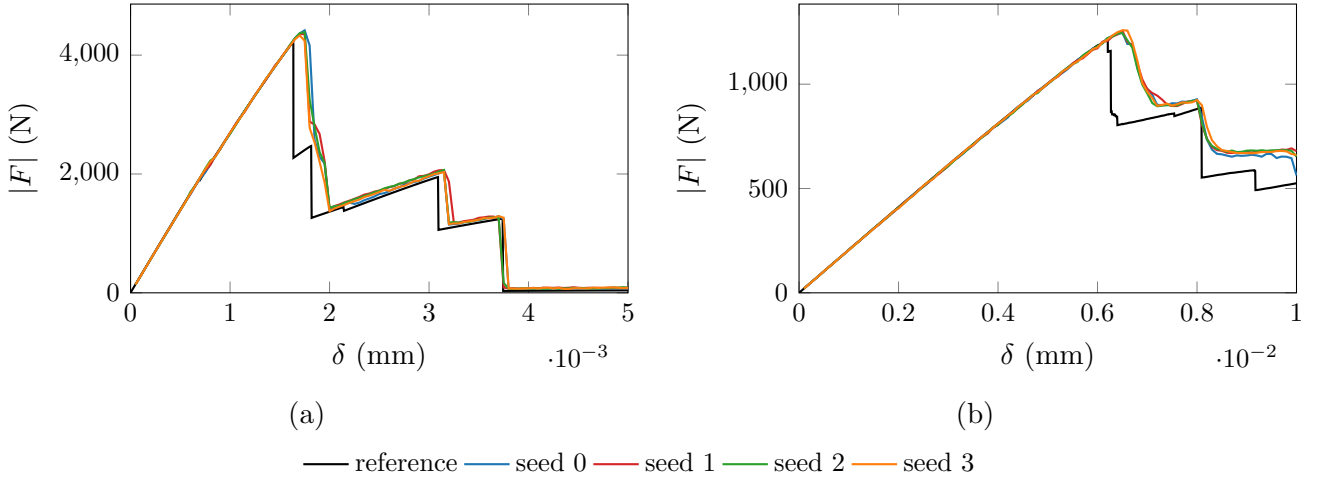
\begin{figure}[!htb]
    \centering
    \begin{subfigure}[t]{0.48\linewidth}
        \centering
        \begin{tikzpicture}
            \begin{axis}[
                paperaxis,
                width=\linewidth, height=5.4cm,
                xlabel={$\delta$ (mm)},
                ylabel={$|F|$ (N)},
                xmin=0, xmax=5e-3, ymin=0,
                legend to name=leg:seedsfd,
                legend columns=5,
                /tikz/every even column/.append style={column sep=0.4cm},
            ]
                \addplot[black, thick]
                    table[col sep=comma, x=delta_mm, y=F]
                    {figures/data/bench_ref_tension.csv};
                \addlegendentry{reference}
                \addplot[cblue, thick]
                    table[col sep=comma, x=delta_mm, y=F]
                    {figures/data/bench_seed_n0_tension.csv};
                \addlegendentry{seed 0}
                \addplot[cred, thick]
                    table[col sep=comma, x=delta_mm, y=F]
                    {figures/data/bench_seed_n1_tension.csv};
                \addlegendentry{seed 1}
                \addplot[cgreen, thick]
                    table[col sep=comma, x=delta_mm, y=F]
                    {figures/data/bench_seed_n2_tension.csv};
                \addlegendentry{seed 2}
                \addplot[corange, thick]
                    table[col sep=comma, x=delta_mm, y=F]
                    {figures/data/bench_seed_n3_tension.csv};
                \addlegendentry{seed 3}
            \end{axis}
        \end{tikzpicture}
        \caption{}
    \end{subfigure}%
    \hfill%
    \begin{subfigure}[t]{0.48\linewidth}
        \centering
        \begin{tikzpicture}
            \begin{axis}[
                paperaxis,
                width=\linewidth, height=5.4cm,
                xlabel={$\delta$ (mm)},
                ylabel={$|F|$ (N)},
                xmin=0, xmax=1e-2, ymin=0,
            ]
                \addplot[black, thick]
                    table[col sep=comma, x=delta_mm, y=F]
                    {figures/data/bench_ref_shear.csv};
                \addplot[cblue, thick]
                    table[col sep=comma, x=delta_mm, y=F]
                    {figures/data/bench_seed_n0_shear.csv};
                \addplot[cred, thick]
                    table[col sep=comma, x=delta_mm, y=F]
                    {figures/data/bench_seed_n1_shear.csv};
                \addplot[cgreen, thick]
                    table[col sep=comma, x=delta_mm, y=F]
                    {figures/data/bench_seed_n2_shear.csv};
                \addplot[corange, thick]
                    table[col sep=comma, x=delta_mm, y=F]
                    {figures/data/bench_seed_n3_shear.csv};
            \end{axis}
        \end{tikzpicture}
        \caption{}
    \end{subfigure}

    \vspace{0.4ex}
    \ref*{leg:seedsfd}
    \caption{Load--displacement curves of the four random seeds of
    configuration $106244$ under (a)~biaxial tension and (b)~shear,
    overlaid on the finite element reference of the dataset.}
    \label{fig:seeds-fd}
\end{figure}

\begin{figure}[!htb]
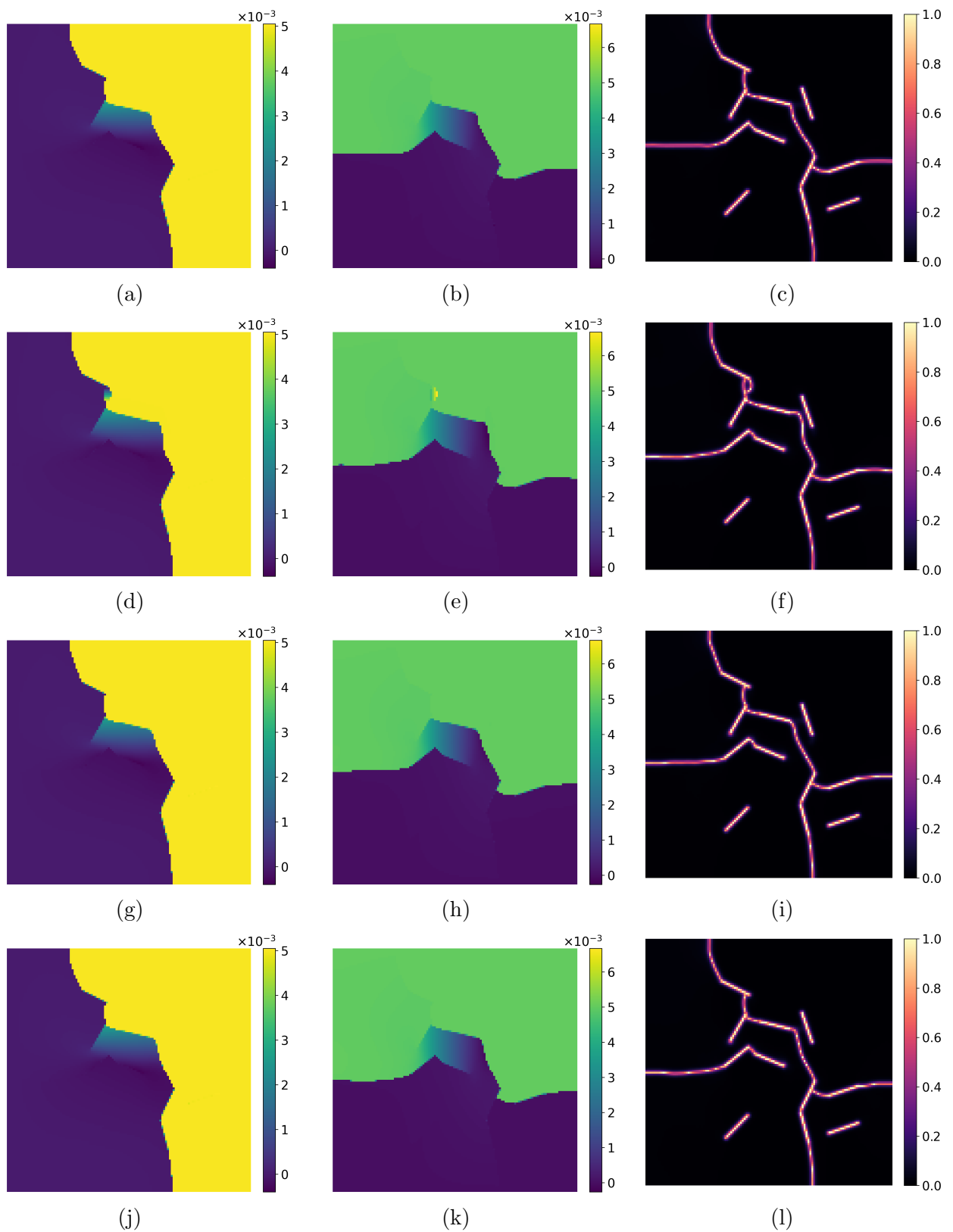

    \centering
\foreach \s in {n0, n1, n2, n3} {%
    \begin{subfigure}[t]{0.33\linewidth}
        \centering
        \includegraphics[width=\linewidth]{bench_u_seed_\s_tension}
        \caption{}
    \end{subfigure}%
    \hfill%
    \begin{subfigure}[t]{0.33\linewidth}
        \centering
        \includegraphics[width=\linewidth]{bench_v_seed_\s_tension}
        \caption{}
    \end{subfigure}%
    \hfill%
    \begin{subfigure}[t]{0.33\linewidth}
        \centering
        \includegraphics[width=\linewidth]{bench_phi_seed_\s_tension}
        \caption{}
    \end{subfigure}
    \par\vspace{0.6ex}
}%
    \caption{Seed ensemble of configuration $106244$ under biaxial
    tension at the end of the loading range. Rows show the four
    random seeds; columns show the displacement components $u$ and
    $v$ in mm and the phase field, each column on one color scale
    shared by the four runs.}
    \label{fig:seeds-tension}
\end{figure}

\begin{figure}[!htb]
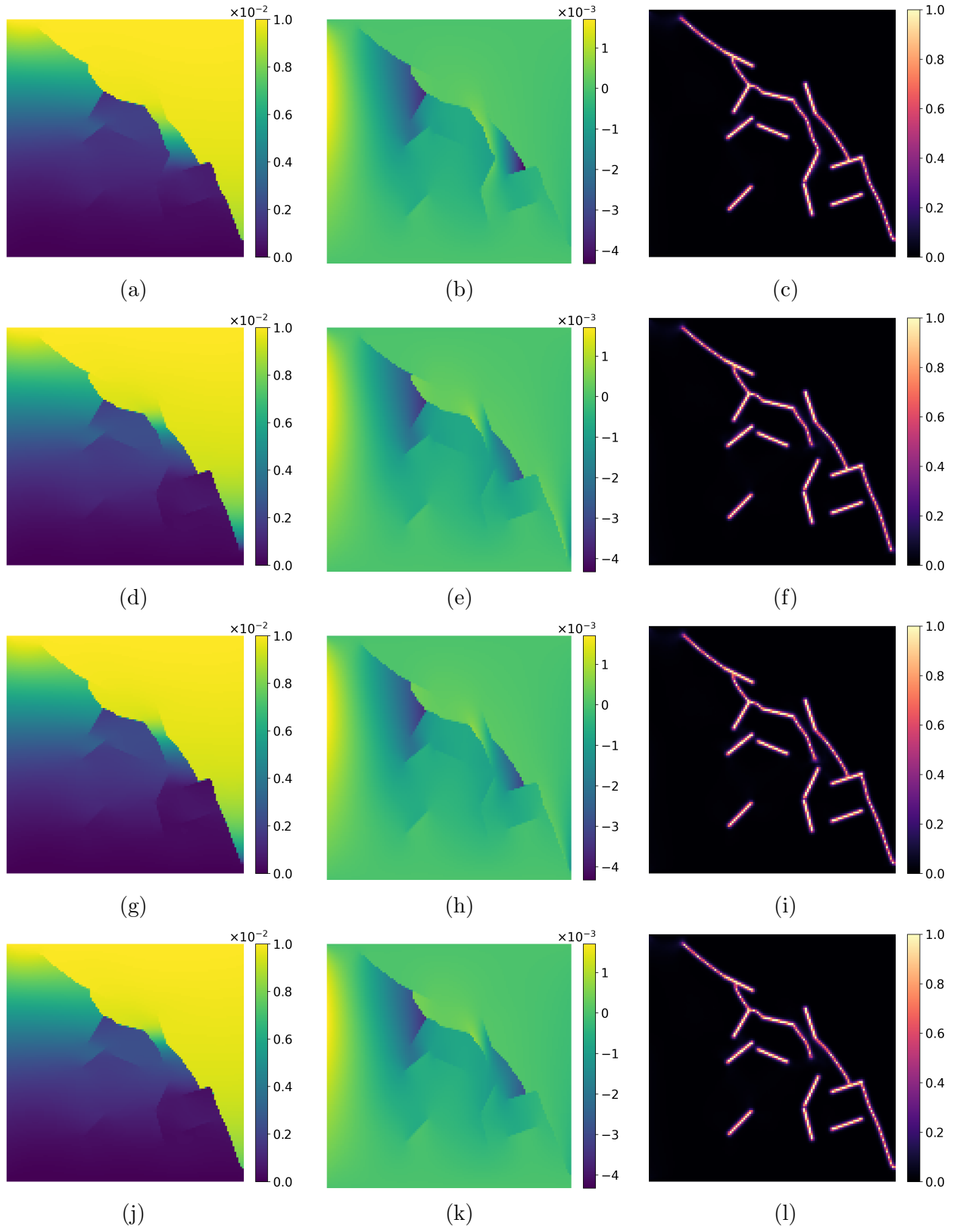

    \centering
\foreach \s in {n0, n1, n2, n3} {%
    \begin{subfigure}[t]{0.33\linewidth}
        \centering
        \includegraphics[width=\linewidth]{bench_u_seed_\s_shear}
        \caption{}
    \end{subfigure}%
    \hfill%
    \begin{subfigure}[t]{0.33\linewidth}
        \centering
        \includegraphics[width=\linewidth]{bench_v_seed_\s_shear}
        \caption{}
    \end{subfigure}%
    \hfill%
    \begin{subfigure}[t]{0.33\linewidth}
        \centering
        \includegraphics[width=\linewidth]{bench_phi_seed_\s_shear}
        \caption{}
    \end{subfigure}
    \par\vspace{0.6ex}
}%
    \caption{Seed ensemble of configuration $106244$ under shear at
    the end of the loading range, in the arrangement of
    Fig.~\ref{fig:seeds-tension}.}
    \label{fig:seeds-shear}
\end{figure}

The deep Ritz baseline of the benchmark dataset produced a different
crack pattern from every network initialization~\cite{hamdi2026}.
Section~\ref{subsec:benchmark} states the summary statistics of the
corresponding test of the proposed method, four random seeds under
each loading of configuration $106244$; this appendix collects the
curves and fields behind those numbers. Fig.~\ref{fig:seeds-fd}
overlays the load--displacement curves of the eight runs on the
reference solutions, with the peaks spread over $2.0\%$ in tension
and $1.1\%$ in shear. Figs.~\ref{fig:seeds-tension}
and~\ref{fig:seeds-shear} show the displacement components and the
phase field of every run at the end of the loading range, each
column on one color scale shared by the four runs. The four
solutions of each case are indistinguishable at the scale of the
figures apart from one short band under tension, which carries the
single differing per-crack decision of the ensemble. That seeded
crack is activated in one of the four runs, as it is in the
reference, and stays dormant in the other three, while the crack
pattern is otherwise unchanged.

\FloatBarrier
\bibliographystyle{unsrtnat}
\bibliography{refs}

\end{document}